\pdfoutput=1
\documentclass{article}

\PassOptionsToPackage{numbers, sort&compress}{natbib}

\usepackage[preprint]{neurips_2026}
\makeatletter
\renewcommand{\@notice}{}
\makeatother

\usepackage[utf8]{inputenc}
\usepackage[T1]{fontenc}
\usepackage{hyperref}
\usepackage{url}
\usepackage{booktabs}
\usepackage{amsfonts}
\usepackage{amsmath}
\usepackage{amssymb}
\usepackage{amsthm}
\usepackage{bm}
\usepackage{nicefrac}
\usepackage{microtype}
\usepackage{xcolor}
\usepackage{graphicx}
\usepackage{multirow}
\usepackage{tabularx}
\usepackage{caption}
\usepackage{float}
\newfloat{algorithm}{tbp}{loa}
\floatname{algorithm}{Algorithm}
\usepackage{tikz}
\usetikzlibrary{arrows.meta, positioning, fit, calc}

\newtheorem{proposition}{Proposition}
\newtheorem{corollary}{Corollary}
\theoremstyle{remark}

\theoremstyle{plain}



\newcommand{\z}{\mathbf{z}}
\newcommand{\uu}{\mathbf{u}}
\newcommand{\vv}{\mathbf{v}}
\newcommand{\A}{\mathbf{A}}
\newcommand{\Dop}{\mathcal{D}}
\newcommand{\Mop}{\mathcal{M}}
\newcommand{\R}{\mathbb{R}}
\newcommand{\Hone}{H^1}
\newcommand{\ip}[2]{\langle #1, #2 \rangle}
\newcommand{\fjepa}{f-JEPA}
\newcommand{\Epi}{E_{[-\pi,\pi]}}
\newcommand{\Eunit}{E_{[0,5]}}

\title{Func-JEPA: Learning a Sobolev Space of Functions\\as Latent Vectors}

\newcommand{\repourl}{https://github.com/amorarc/Func-JEPA}
\newcommand{\hfurl}{https://huggingface.co/amorarc/func-jepa}
\newcommand{\githubicon}{\raisebox{-0.14em}{\tikz[x=0.0625em,y=0.0625em]\fill (6.7660,-11.3280) .. controls (4.7030,-11.0780) and (3.2500,-9.5940) .. (3.2500,-7.6720) .. controls (3.2500,-6.8910) and (3.5310,-6.0470) .. (4.0000,-5.4840) .. controls (3.7970,-4.9690) and (3.8280,-3.8750) .. (4.0630,-3.4220) .. controls (4.6880,-3.3440) and (5.5310,-3.6720) .. (6.0310,-4.1250) .. controls (6.6250,-3.9380) and (7.2500,-3.8440) .. (8.0160,-3.8440) .. controls (8.7810,-3.8440) and (9.4060,-3.9380) .. (9.9690,-4.1090) .. controls (10.4530,-3.6720) and (11.3130,-3.3440) .. (11.9380,-3.4220) .. controls (12.1560,-3.8440) and (12.1880,-4.9370) .. (11.9840,-5.4690) .. controls (12.4840,-6.0620) and (12.7500,-6.8590) .. (12.7500,-7.6720) .. controls (12.7500,-9.5940) and (11.2970,-11.0470) .. (9.2030,-11.3120) .. controls (9.7340,-11.6560) and (10.0930,-12.4060) .. (10.0930,-13.2660) -- (10.0930,-14.8910) .. controls (10.0930,-15.3590) and (10.4840,-15.6250) .. (10.9530,-15.4380) .. controls (13.7810,-14.3590) and (16.0000,-11.5300) .. (16.0000,-8.0300) .. controls (16.0000,-3.6100) and (12.4060,-0.0000) .. (7.9840,-0.0000) .. controls (3.5630,-0.0000) and (0.0000,-3.6100) .. (0.0000,-8.0310) .. controls (-0.0092,-11.3467) and (2.0582,-14.3135) .. (5.1720,-15.4530) .. controls (5.5940,-15.6090) and (6.0000,-15.3280) .. (6.0000,-14.9060) -- (6.0000,-13.6560) .. controls (5.7810,-13.7500) and (5.5000,-13.8120) .. (5.2500,-13.8120) .. controls (4.2190,-13.8120) and (3.6100,-13.2500) .. (3.1720,-12.2030) .. controls (3.0000,-11.7810) and (2.8120,-11.5310) .. (2.4530,-11.4840) .. controls (2.2660,-11.4690) and (2.2030,-11.3910) .. (2.2030,-11.2970) .. controls (2.2030,-11.1090) and (2.5160,-10.9690) .. (2.8280,-10.9690) .. controls (3.2810,-10.9690) and (3.6720,-11.2500) .. (4.0780,-11.8290) .. controls (4.3910,-12.2810) and (4.7180,-12.4840) .. (5.1090,-12.4840) .. controls (5.5000,-12.4840) and (5.7500,-12.3440) .. (6.1090,-11.9840) .. controls (6.3750,-11.7190) and (6.5790,-11.4840) .. (6.7660,-11.3280);}}
\newcommand{\hficon}{\raisebox{-0.17em}{\includegraphics[height=1.05em]{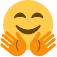}}}

\author{%
  Alejandro Moreno Arcas \\
  Independent Researcher \\[6pt]
  \githubicon~Code: \url{\repourl} \\
  \hficon~Model: \url{\hfurl} \\
}

\begin{document}

\maketitle

\begin{abstract}
Pretraining an encoder once and reusing it across tasks is a widespread approach in vision and language.
However, neural symbolic mathematics still trains a dedicated sequence-to-sequence model for each task, so that no representation of a function is shared.
In this work, we propose the Function Joint-Embedding Predictive Architecture (Func-JEPA, abbreviated \fjepa{}): an encoder of mathematical functions, pretrained on their symbolic form, whose latent space imitates the Sobolev space $\Hone$ and in which differentiation is a learned latent operator.
The norm of a latent is computed exactly from the function, whereas its direction is learned to match Sobolev angles.
The frozen encoder is then reused by lightweight modules for decoding, composition, integration, and regression from samples.
The representation and four downstream tasks are evaluated within and outside the pretraining distribution.
\fjepa{} recovers exact formulas from noisy samples more often than the compared symbolic-regression methods, whereas computer algebra and task-specific models remain stronger on standard integration and differential-equation benchmarks.
\end{abstract}

\section{Introduction}
\label{sec:introduction}

Nowadays, encoders pretrained once on large corpora and reused for many downstream tasks are the standard approach in language and vision \cite{devlin2019bert, assran2023ijepa}.
Neural symbolic mathematics has not followed this path: integration, the solution of ordinary differential equations (ODEs), and symbolic regression are posed as translation problems, each solved by a dedicated sequence-to-sequence transformer trained on its own synthetic corpus \cite{lample2020deep, charton2022linear, biggio2021nesymres, kamienny2022e2e, dascoli2024odeformer}.
Whilst these models achieve remarkable accuracy, the representation of a function learned for one task is not reused by the others.

A reusable encoder of functions should reflect what a function is in analysis: an element of a vector space with an inner product, on which differentiation, integration, and composition act as operators.
Joint-embedding predictive architectures (JEPA) \cite{lecun2022path, assran2023ijepa, bardes2024vjepa, chen2026vljepa} are well-suited to this purpose, since they predict the representation of a target from that of a context: the context is a function $f$, the target is its derivative $f'$, and the predictor is the derivative operator.
However, a predictive loss alone is minimised by collapsed representations, which current methods avoid with architectural asymmetries \cite{grill2020byol, assran2023ijepa} or with regularisers on the distribution of the embeddings \cite{bardes2022vicreg, balestriero2025lejepa}.
For functions, a stronger anchor is available: the inner product of two functions can be computed exactly from their values, so the target geometry of the latent space is defined by the data, a property that existing embeddings of mathematical expressions do not exploit \cite{allamanis2017eqnet, meidani2024snip}.

This work demonstrates that a transformer encoder can be pretrained, from symbolic text alone, into a latent space that is close to an isometric embedding of the Sobolev space $\Hone$ and in which the derivative is a learned operator.
This leads us to propose the Function Joint-Embedding Predictive Architecture (Func-JEPA, abbreviated \fjepa{} throughout this paper): a novel pretrained encoder whose latent vectors factorise into an exactly computed $\Hone$ norm and a learned unit direction, trained with an angle-only Sobolev geometry, a linearity objective, and a JEPA derivative objective.
The derivative predictor is factored into a gain and a direction, which makes it exactly homogeneous.
From this, a set of lightweight modules that reuse the frozen encoder is proposed: a grammar-constrained decoder from latents to formulas, a bank of composition operators, an integrator, and an encoder of sampled values.

In order to evaluate \fjepa{}, a series of experiments are conducted at three levels: the pretrained representation, compared with released encoders of symbolic expressions; four downstream tasks on the frozen encoder, namely symbolic integration, symbolic regression, and the solution and discovery of ODEs; and the transfer to out-of-distribution data and standard benchmarks.
In every task the learned components are compared with a network-free control that differs only in the learned part and with released models and established tools \cite{lample2020deep, kamienny2022e2e, cranmer2023pysr, dascoli2024odeformer, desilva2020pysindy}, and every reported solution is verified with SymPy \cite{meurer2017sympy} or numerically.
As a concrete example, the encoder maps the text of $f = 4\sin 7x - 3\cos 7x$ to a latent whose norm is exactly $\|f\|_{\Hone}$, the operator $\Dop$ maps it to a prediction of the latent of $f'$, and the decoder turns this prediction into $21\sin 7x + 28\cos 7x$, the exact derivative, without any symbolic differentiation (Figure~\ref{fig:predops}).

\section{Related Work}
\label{sec:related}

\paragraph{Representations of Mathematical Expressions}
EqNet \cite{allamanis2017eqnet} and later sequence-to-sequence and contrastive encoders \cite{gangwar2023semantic, zheng2025egen} learn embeddings in which equivalent expressions are close, using pairs of equivalent rewritings as supervision.
SNIP \cite{meidani2024snip} aligns a symbolic and a numeric encoder with a contrastive objective, and Valentino et al. \cite{valentino2024latent} learn mathematical operations, including differentiation, as transformations between expression embeddings evaluated by retrieval.
In \fjepa{}, the target of the geometry is instead a Sobolev inner product computed from function values, so that norms, angles, and distances have an analytical meaning.

\paragraph{Joint-Embedding Predictive Architectures}
Methods in this family pretrain encoders by predicting representations rather than inputs \cite{lecun2022path, assran2023ijepa, bardes2024vjepa}, and VL-JEPA \cite{chen2026vljepa} employs a lightweight text decoder only at inference, which is the template followed by our decoder.
Collapse is avoided by stop-gradient and teacher-student asymmetries \cite{grill2020byol}, by variance and covariance penalties \cite{bardes2022vicreg}, or by matching an isotropic Gaussian distribution \cite{balestriero2025lejepa}, and BRo-JEPA \cite{jha2026brojepa} imposes the circular structure of modular arithmetic on the predictor.
In \fjepa{}, the context-target pair is a function and its derivative, and collapse is prevented mainly by a data-defined Gram target, with only a weak variance floor.

\paragraph{Transformers for Symbolic Mathematics}
Lample and Charton \cite{lample2020deep} showed that sequence-to-sequence transformers trained on synthetic prefix expressions can integrate functions and solve ODEs, and later work studied linear algebra \cite{charton2022linear}.
Such models are brittle outside their training distribution \cite{welleck2022brittleness}, which has motivated search-based alternatives in which a transformer selects rules applied by a symbolic engine \cite{unsal2024alphaintegrator}.
We adopt their prefix notation, but every task shares a single pretrained encoder and is solved by operators and searches in its latent space, followed by exact verification.

\paragraph{Neural Symbolic Regression}
Transformer-based methods encode a set of points and decode a prefix expression \cite{biggio2021nesymres, kamienny2022e2e, vastl2024symformer, saegert2026flashansr}, usually followed by constant refinement.
Search-based methods include genetic programming \cite{cranmer2023pysr}, recursive decomposition \cite{udrescu2020aifeynman}, reinforcement learning \cite{petersen2021dsr, mundhenk2021seeding}, and large language models proposing equation skeletons \cite{shojaee2025llmsr}, many of which are compared in SRBench \cite{lacava2021srbench}.
In our case, the value encoder is trained to reach the latent of the frozen symbolic encoder, so that symbolic regression shares its decoder with the other tasks.

\paragraph{Operator Learning and Equation Discovery}
Neural operators learn maps between function spaces from sampled values \cite{lu2021deeponet, li2021fno, kovachki2023neuraloperator}, function encoders represent functions as coefficients on a learned basis \cite{ingebrand2024zeroshot}, and Koopman-based methods search for coordinates in which a nonlinear operator acts linearly \cite{koopman1931, schmid2010dmd, lusch2018koopman}.
Equation discovery is commonly addressed by sparse regression on libraries of candidate terms \cite{brunton2016sindy, messenger2021weaksindy}, and more recently with transformers that output symbolic ODEs \cite{dascoli2024odeformer, chang2025ddot}.
\fjepa{} differs in that its input is symbolic and its outputs can be decoded back to exact formulas, and, unlike Sobolev training \cite{czarnecki2017sobolev}, it employs the Sobolev inner product as the geometry of the latent space rather than as a regression target.

\section{Proposed Method}
\label{sec:method}

\subsection{Function-space geometry}
\label{sec:geometry}
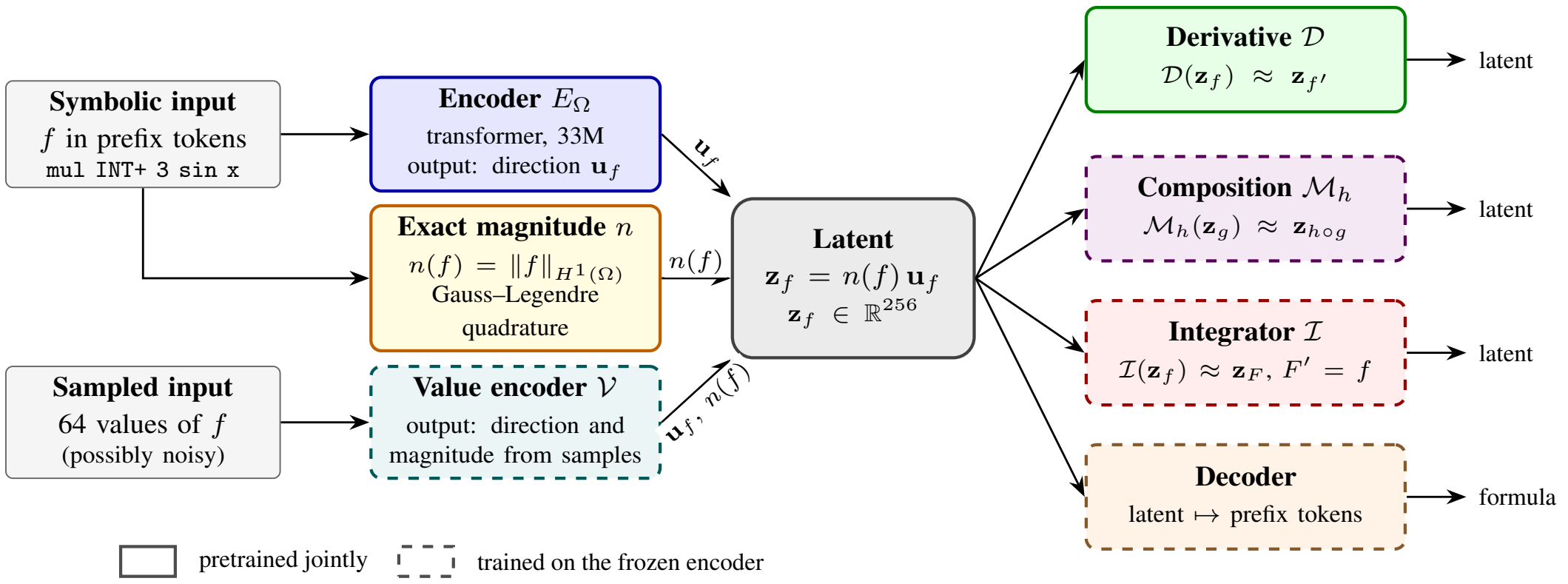
\begin{figure}[t!]
\centering
\resizebox{\linewidth}{!}{%
\begin{tikzpicture}[
  font=\footnotesize,
  io/.style={draw=black!60, fill=gray!8, rounded corners=2pt, align=center, text width=2.55cm, minimum height=1.05cm, inner sep=3pt},
  mod/.style={rounded corners=3pt, align=center, text width=2.7cm, minimum height=1.05cm, inner sep=3pt, line width=0.9pt},
  pre/.style={mod, solid},
  post/.style={mod, dashed},
  arr/.style={-{Stealth[length=2mm]}, semithick},
  lab/.style={font=\scriptsize, fill=white, inner sep=1pt}]
\node[io] (tok) at (0, 1.35) {\textbf{Symbolic input}\\[1pt]$f$ in prefix tokens\\[-1pt]\texttt{\scriptsize mul INT+ 3 sin x}};
\node[io] (smp) at (0, -1.55) {\textbf{Sampled input}\\[1pt]64 values of $f$\\[-1pt]{\scriptsize (possibly noisy)}};
\node[pre, draw=blue!65!black, fill=blue!10] (enc) at (3.75, 1.35) {\textbf{Encoder} $E_\Omega$\\[1pt]{\scriptsize transformer, 33M}\\[-1pt]{\scriptsize output: direction $\uu_f$}};
\node[pre, draw=orange!75!black, fill=yellow!14] (mag) at (3.75, -0.1) {\textbf{Exact magnitude} $n$\\[1pt]{\scriptsize $n(f) = \|f\|_{\Hone(\Omega)}$}\\[-1pt]{\scriptsize Gauss--Legendre quadrature}};
\node[post, draw=teal!70!black, fill=teal!8] (V) at (3.75, -1.55) {\textbf{Value encoder} $\mathcal{V}$\\[1pt]{\scriptsize output: direction and}\\[-1pt]{\scriptsize magnitude from samples}};
\node[draw=black!75, fill=gray!16, rounded corners=6pt, align=center, text width=2.2cm, minimum height=1.6cm, line width=0.9pt] (z) at (7.15, -0.1)
  {\textbf{Latent}\\[1pt]$\z_f = n(f)\,\uu_f$\\[1pt]{\scriptsize $\z_f \in \R^{256}$}};
\node[pre, draw=green!50!black, fill=green!10, text width=3.0cm] (D) at (11.1, 2.1) {\textbf{Derivative} $\Dop$\\[1pt]{\scriptsize $\Dop(\z_f) \approx \z_{f'}$}};
\node[post, draw=violet!70!black, fill=violet!9, text width=3.0cm] (M) at (11.1, 0.6) {\textbf{Composition} $\Mop_h$\\[1pt]{\scriptsize $\Mop_h(\z_g) \approx \z_{h\circ g}$}};
\node[post, draw=red!60!black, fill=red!7, text width=3.0cm] (I) at (11.1, -0.85) {\textbf{Integrator} $\mathcal{I}$\\[1pt]{\scriptsize $\mathcal{I}(\z_f) \approx \z_{F}$, $F' = f$}};
\node[post, draw=brown!70!black, fill=orange!10, text width=3.0cm] (dec) at (11.1, -2.3) {\textbf{Decoder}\\[1pt]{\scriptsize latent $\mapsto$ prefix tokens}};
\draw[arr] (tok) -- (enc);
\draw[arr] (tok.south) |- (mag.west);
\draw[arr] (smp) -- (V);
\draw[arr] (enc.east) -- node[lab, above, sloped] {$\uu_f$} (z.north west);
\draw[arr] (mag.east) -- node[lab, above] {$n(f)$} (z.west);
\draw[arr] (V.east) -- node[lab, below, sloped] {$\uu_f,\ n(f)$} (z.south west);
\draw[arr] (z.east) -- (D.west);
\draw[arr] (z.east) -- (M.west);
\draw[arr] (z.east) -- (I.west);
\draw[arr] (z.east) -- (dec.west);
\draw[arr] (dec.east) -- ++(0.6, 0) node[right, align=left, font=\scriptsize] {formula};
\draw[arr] (D.east) -- ++(0.6, 0) node[right, align=left, font=\scriptsize] {latent};
\draw[arr] (M.east) -- ++(0.6, 0) node[right, align=left, font=\scriptsize] {latent};
\draw[arr] (I.east) -- ++(0.6, 0) node[right, align=left, font=\scriptsize] {latent};
\node[draw=black!70, line width=0.9pt, solid, minimum width=0.55cm, minimum height=0.3cm] (l1) at (0.05, -2.95) {};
\node[right=0.1cm of l1, font=\scriptsize] {pretrained jointly};
\node[draw=black!70, line width=0.9pt, dashed, minimum width=0.55cm, minimum height=0.3cm] (l2) at (2.85, -2.95) {};
\node[right=0.1cm of l2, font=\scriptsize] {trained on the frozen encoder};
\end{tikzpicture}}
\caption{Modules of \fjepa{} and the data flow between them. Solid borders mark the modules pretrained jointly and dashed borders the modules trained afterwards on the frozen encoder}
\label{fig:overview}
\end{figure}

\fjepa{} is built in two stages: an encoder of functions is pretrained once from symbolic text (Sections~\ref{sec:geometry} to~\ref{sec:objective}), and is then frozen and reused by lightweight modules (Section~\ref{sec:modules}), as summarised in Figure~\ref{fig:overview}.
The objective of the pretraining is that the latent space imitates a Sobolev space of functions, so that the operations of analysis become operations on vectors.
For this reason, every structural property of the function space is enforced on the latent space, either by construction or by a training objective:
\begin{itemize}
\setlength\itemsep{0pt}
  \item Norm: $\|\z_f\| = \|f\|_{\Hone}$, computed exactly by quadrature (Section~\ref{sec:encoder}).
  \item Inner product: $\z_f^\top\z_g \approx \ip{f}{g}_{\Hone}$, since latent angles are fitted to Sobolev angles (geometry objective).
  \item Linearity: $\z_{af+bg} \approx a\,\z_f + b\,\z_g$, including $\z_{cf} \approx c\,\z_f$ (linearity objective).
  \item Differentiation: $\Dop(\z_f) \approx \z_{f'}$, with an operator $\Dop$ that is exactly homogeneous by construction (Section~\ref{sec:derivative}) and learned with the JEPA objective.
\end{itemize}

For univariate real functions on an interval $[a, b]$, we consider the inner product and its induced norm
\begin{equation}
\ip{f}{g}_{[a,b]} = \int_a^b f(x)\,g(x)\,dx, \qquad \|f\|_{[a,b]} = \sqrt{\ip{f}{f}_{[a,b]}}.
\label{eq:ip}
\end{equation}
The encoder employs the normalised version of Equation~\ref{eq:ip} on $\Omega = [a, b]$ and its Sobolev extension:
\begin{equation}
\ip{f}{g}_{\Omega} = \frac{1}{b-a}\,\ip{f}{g}_{[a,b]}, \qquad
\ip{f}{g}_{\Hone(\Omega)} = \ip{f}{g}_{\Omega} + \lambda^2 \ip{f'}{g'}_{\Omega},
\label{eq:h1}
\end{equation}
with induced norm $\|f\|_{\Hone(\Omega)} = \ip{f}{f}_{\Hone(\Omega)}^{1/2}$, where $\lambda = 0.5$ and the integrals are evaluated with the $N = 64$-node Gauss-Legendre rule, using the values of $f$ and of its symbolic derivative $f'$ at the nodes.
The $\Hone$ geometry is preferred over $L^2$ because differentiation is unbounded on $L^2$, whereas $\|f'\|_{\Omega} \leq \lambda^{-1}\|f\|_{\Hone(\Omega)}$, which keeps the ratio between the norms of a function and of its derivative well conditioned.

A function encoder on $\Omega$ is a map $E_\Omega$ from the symbolic form of $f$ to a latent vector $\z_f \in \R^d$ whose Euclidean inner product reproduces Equation~\ref{eq:h1}, $\z_f^\top\z_g \approx \ip{f}{g}_{\Hone(\Omega)}$.
Such a representation exists: the coefficients of $f$ on an $\Hone$-orthonormal polynomial basis define a linear map whose inner products reproduce Equation~\ref{eq:h1} up to an explicit quadrature and truncation error (Proposition~\ref{prop:oracle} in Appendix~\ref{app:ip}).
The encoder therefore has to learn an approximately isometric and linear embedding from symbols alone.
Two encoders are trained with the same architecture, objective, and schedule: $\Epi$ on $[-\pi, \pi]$, with which all results of Section~\ref{sec:experiments} are obtained, and $\Eunit$ on $[0, 5]$, evaluated in Appendix~\ref{app:eunit}.

The pretraining corpus of $\Epi$ consists of 2.0M synthetic functions drawn from 17 parameterised families, such as polynomials, Fourier sums, Gaussians, rational functions, and solutions of linear ODEs with integer coefficients, complemented with short random expression trees.
Every row stores the prefix tokens of $f$ and of its derivative $f'$ together with their values at the quadrature nodes, and rewritings of the same function and coefficient sweeps are never split across partitions.
Further details can be found in Appendix~\ref{app:data}.

\subsection{Encoder with computed magnitude}
\label{sec:encoder}
The encoder $E_\Omega$ is a BERT-style transformer \cite{devlin2019bert} with 10 layers, hidden size 512 and about 33M parameters, which reads the prefix tokens of $f$.
Its \texttt{[CLS]} state is projected to $\vv_f \in \R^{d}$ with $d = 256$, and the latent vector is defined as
\begin{equation}
\z_f = n(f)\,\uu_f, \qquad \uu_f = \frac{\vv_f}{\|\vv_f\|}, \qquad n(f) = \|f\|_{\Hone(\Omega)},
\label{eq:latent}
\end{equation}
where $n(f)$ is computed from the values of $f$ and $f'$ at the quadrature nodes rather than learned.
With Equation~\ref{eq:latent}, $\z_f^\top\z_g = n(f)\,n(g)\,\uu_f^\top\uu_g$, so the norms are exact by construction and the only learned quantity is the angle between two latents.
Proposition~\ref{prop:ip} in Appendix~\ref{app:ip} makes this precise: the error $|\z_f^\top\z_g - \ip{f}{g}_{\Hone(\Omega)}|$ is at most $n(f)\,n(g)$ times the angular error of the encoder, plus a quadrature error that decays geometrically with $N$ for analytic functions.
Moreover, the $\Hone$ norms of the training functions span more than four orders of magnitude, and with a computed norm homogeneity reduces to the invariance of the direction, $\uu_{cf} = \mathrm{sign}(c)\,\uu_f$.

\paragraph{Closeness in latent space}
Since latent distances approximate Sobolev distances (Corollary~\ref{cor:dist} in Appendix~\ref{app:ip}), a prediction $\hat{\z}$ is compared with its target $\z$ through the relative error $\delta(\hat{\z}, \z) = \|\hat{\z} - \z\|/\|\z\|$, which, unlike the cosine, penalises a wrong norm.
A prediction is considered correct under the $\varepsilon$-criterion when $\delta(\hat{\z}, \z) \leq \varepsilon$.
The threshold is calibrated on the validation split: the relative distance from a function to its nearest distinct function has a median of 0.128 and a 5th percentile of 0.0102, and we set $\varepsilon = 0.0051$, half of this percentile, so that the $\varepsilon$-balls of two functions do not overlap for 95\% of the functions (Appendix~\ref{app:epscal}).

\subsection{Factored derivative operator}
\label{sec:derivative}
The derivative operator $\Dop : \R^d \to \R^d$ predicts $\z_{f'}$ from $\z_f$.
Writing $\hat{\uu} = \z/\|\z\|$, it is factored into a gain and a direction:
\begin{equation}
\Dop(\z) = \|\z\|\,\gamma(\hat{\uu})\,\bm{\delta}(\hat{\uu}), \qquad
\gamma(\hat{\uu}) = e^{k(\hat{\uu}) + k(-\hat{\uu})}, \qquad
\bm{\delta}(\hat{\uu}) = \frac{\A\hat{\uu} + h(\hat{\uu}) - h(-\hat{\uu})}{\|\A\hat{\uu} + h(\hat{\uu}) - h(-\hat{\uu})\|},
\label{eq:deriv}
\end{equation}
where $\A \in \R^{d\times d}$ is initialised to the identity, and $h$ and $k$ are multilayer perceptrons.
The gain $\gamma$ is even and positive and learns the ratio $n(f')/n(f)$, whereas the direction $\bm{\delta}$ is odd and of unit norm.

\begin{proposition}
\label{prop:homogeneity}
For every $\z \neq \mathbf{0}$ and every $c \in \R \setminus \{0\}$, the operator in Equation~\ref{eq:deriv} satisfies $\Dop(c\,\z) = c\,\Dop(\z)$.
\end{proposition}

The proof can be found in Appendix~\ref{app:proof}.
Thus, homogeneity is exact by construction, and powers $\Dop^2$ and polynomials $p(\Dop)$ are available by composition without being trained.
It is worth noting that $\Dop$ is not additive by construction: additivity is not enforced on the operator and is only learned indirectly, through the linearity of the latent.

\subsection{Training objective}
\label{sec:objective}

\begin{figure}[t]
\centering
\resizebox{\linewidth}{!}{\begin{tikzpicture}[
  font=\scriptsize,
  inp/.style={draw=black!55, rounded corners=2pt, fill=gray!8, align=center, minimum height=0.8cm, text width=2.3cm, inner sep=2pt},
  lat/.style={draw=blue!55!black, rounded corners=2pt, fill=blue!6, align=center, minimum height=0.8cm, text width=2.25cm, inner sep=2pt},
  loss/.style={draw, rounded corners=3pt, align=flush left, inner sep=3pt, text width=7.75cm},
  arr/.style={-{Stealth[length=1.6mm]}, semithick},
  lab/.style={font=\tiny, fill=white, inner sep=0.8pt}]

\node[inp] (f)  at (1.15, 6.1) {$f$ \;{\tiny (tokens)}\\[-1pt]\texttt{\tiny mul INT+ 3 sin x}};
\node[inp] (fp) at (1.15, 4.75) {$f'$ \;{\tiny (canonical derivative)}\\[-1pt]\texttt{\tiny mul INT+ 3 cos x}};
\node[inp] (h)  at (1.15, 3.4) {$h = af + bg$ \;or\; $cf$\\[-1pt]{\tiny written in prefix notation}};
\node[inp, fill=yellow!10, draw=orange!70!black] (vals) at (1.15, 1.65) {{\tiny values of $f$, $f'$, $h$, $h'$}\\[-1pt]{\tiny at 64 Gauss--Legendre nodes}};

\node[draw=blue!60!black, fill=blue!12, rounded corners=3pt, minimum width=0.9cm, minimum height=3.5cm, align=center] (enc) at (2.95, 4.75)
  {\rotatebox{90}{\parbox{3.3cm}{\centering shared encoder $E_\Omega$\\{\tiny transformer, 33M parameters}}}};
\draw[arr] (f.east) -- (f.east -| enc.west);
\draw[arr] (fp.east) -- (fp.east -| enc.west);
\draw[arr] (h.east) -- (h.east -| enc.west);

\node[lat] (zf)  at (4.75, 6.1) {$\mathbf{z}_f = n(f)\,\mathbf{u}_f$\\[-1pt]{\tiny $\mathbf{u}_f = \mathbf{v}_f/\|\mathbf{v}_f\|$}};
\node[lat] (zfp) at (4.75, 4.75) {$\mathbf{z}_{f'} = n(f')\,\mathbf{u}_{f'}$};
\node[lat] (zh)  at (4.75, 3.4) {$\mathbf{z}_{h} = n(h)\,\mathbf{u}_{h}$};
\draw[arr] (enc.east |- zf.west) -- (zf.west);
\draw[arr] (enc.east |- zfp.west) -- (zfp.west);
\draw[arr] (enc.east |- zh.west) -- (zh.west);

\node[draw=orange!70!black, fill=yellow!10, rounded corners=2pt, align=center, text width=2.75cm, minimum height=0.8cm, inner sep=2pt] (h1) at (4.75, 1.65)
  {$n(\cdot) = \|\cdot\|_{H^1}$\\[-1pt]$C_{ij} = \cos_{H^1}(f_i, f_j)$};
\draw[arr] (vals.east) -- (h1.west);
\draw[arr, orange!70!black, densely dotted] (h1.north) -- node[lab, right, text=orange!50!black, pos=0.45]{exact $n(\cdot)$} (zh.south);

\node[loss, draw=green!45!black, fill=green!6] (jepa) at (11.1, 6.1) {%
  \textbf{JEPA}\; predictor $\mathcal{D}(\mathbf{z}) = \|\mathbf{z}\|\,\gamma(\hat{\mathbf{u}})\,\bm{\delta}(\hat{\mathbf{u}})$, \; target $\bar{\mathbf{z}}_{f'}$, the latent of $f'$\\[2pt]
  $\mathcal{L}_{\mathrm{JEPA}} = 1 - \cos\big(\mathcal{D}(\mathbf{z}_f), \bar{\mathbf{z}}_{f'}\big) + \beta_g\big(\log\|\mathcal{D}(\mathbf{z}_f)\| - \log\|\bar{\mathbf{z}}_{f'}\|\big)^{2}$\\[1pt]
  {\tiny direction term trains $\bm{\delta}$ (odd, unit norm); log-gain term trains $\gamma$ (even, positive)}};
\node[loss, draw=blue!50!black, fill=blue!5] (lin) at (11.1, 4.6) {%
  \textbf{Linearity}\; combinations $h = af + bg$ and scaled copies $h = cf$\\[2pt]
  $\mathcal{L}_{\mathrm{lin}} = \|\mathbf{z}_{h} - (a\,\mathbf{z}_f + b\,\mathbf{z}_g)\|^2 \big/ \tfrac12\big(\|\mathbf{z}_h\|^2 + \|a\,\mathbf{z}_f + b\,\mathbf{z}_g\|^2\big)$};
\node[loss, draw=violet!60!black, fill=violet!5] (geo) at (11.1, 3.2) {%
  \textbf{Geometry}\; angles only, batch of directions $\mathbf{U} = [\mathbf{u}_1; \dots; \mathbf{u}_B]$\\[2pt]
  $\mathcal{L}_{\mathrm{geo}} = \tfrac{1}{B^2}\|\mathbf{U}\mathbf{U}^{\top} - \mathbf{C}\|_F^2 + \beta_c\big(1 - \rho^2(\mathbf{H}\mathbf{U}\mathbf{U}^{\top}\mathbf{H}, \mathbf{H}\mathbf{C}\mathbf{H})\big)$};
\node[loss, draw=orange!70!black, fill=orange!6] (var) at (11.1, 1.85) {%
  \textbf{Variance floor}\; $\mathcal{L}_{\mathrm{var}} = \tfrac1d\sum_{j=1}^{d} \max\big(0, \tau - \mathrm{std}_{\mathrm{batch}}(\sqrt{d}\,u_{j})\big)$};

\draw[arr] ([yshift=1.5mm]zf.east) -- ([yshift=1.5mm]zf.east -| jepa.west) node[lab, pos=0.5, above] {$\mathbf{z}_f$};
\draw[arr] (zfp.east) -- (6.95, 4.75) |- node[lab, pos=0.25, right] {target} ([yshift=-2.5mm]jepa.west);
\draw[arr] ([yshift=-1.5mm]zf.east) -| (6.15, 4.75) -- (lin.west |- 0, 4.75);
\draw[arr] ([yshift=1.5mm]zh.east) -| (6.35, 4.45) -- (lin.west |- 0, 4.45);
\draw[arr] ([yshift=-1.5mm]zh.east) -| (6.55, 3.35) -- node[lab, pos=0.55, above] {$\mathbf{U}$} (geo.west |- 0, 3.35);
\draw[arr] (6.55, 3.35) |- (var.west);
\draw[arr, orange!70!black] (h1.east) -| (6.75, 3.05) -- node[lab, pos=0.55, below, text=orange!50!black] {$\mathbf{C}$} (geo.west |- 0, 3.05);

\node[draw=black!60, rounded corners=3pt, fill=gray!4, inner sep=4pt, align=center] (tot) at (5.9, 0.35) {%
  $\mathcal{L} = \mathcal{L}_{\mathrm{geo}} + \mathcal{L}_{\mathrm{var}} + r(t)\,\big(\mathcal{L}_{\mathrm{lin}} + \mathcal{L}_{\mathrm{JEPA}}\big) + 0.05\,\mathcal{L}_{v}$, \qquad
  $\mathcal{L}_v = \big(\log(\|\mathbf{v}\|/\sqrt{d})\big)^2$};
\begin{scope}[shift={(12.3, -0.08)}]
  \draw[-{Stealth[length=1mm]}, thin] (0,0) -- (2.15,0) node[right, font=\tiny] {step $t$};
  \draw[-{Stealth[length=1mm]}, thin] (0,0) -- (0,0.75) node[above, font=\tiny] {$r(t)$};
  \draw[green!45!black, thick] (0,0.02) -- (0.6,0.02) -- (1.2,0.55) -- (2.0,0.55);
  \node[font=\tiny, below] at (0.6,0) {1k};
  \node[font=\tiny, below] at (1.2,0) {2k};
  \node[font=\tiny, left] at (0,0.55) {1};
\end{scope}
\end{tikzpicture}}
\caption{Construction of the training objective. The tokens of $f$, of its derivative $f'$, and of synthetic combinations $h$ are encoded by the same encoder, and latent norms and target cosines are computed exactly from the function values}
\label{fig:objective}
\end{figure}
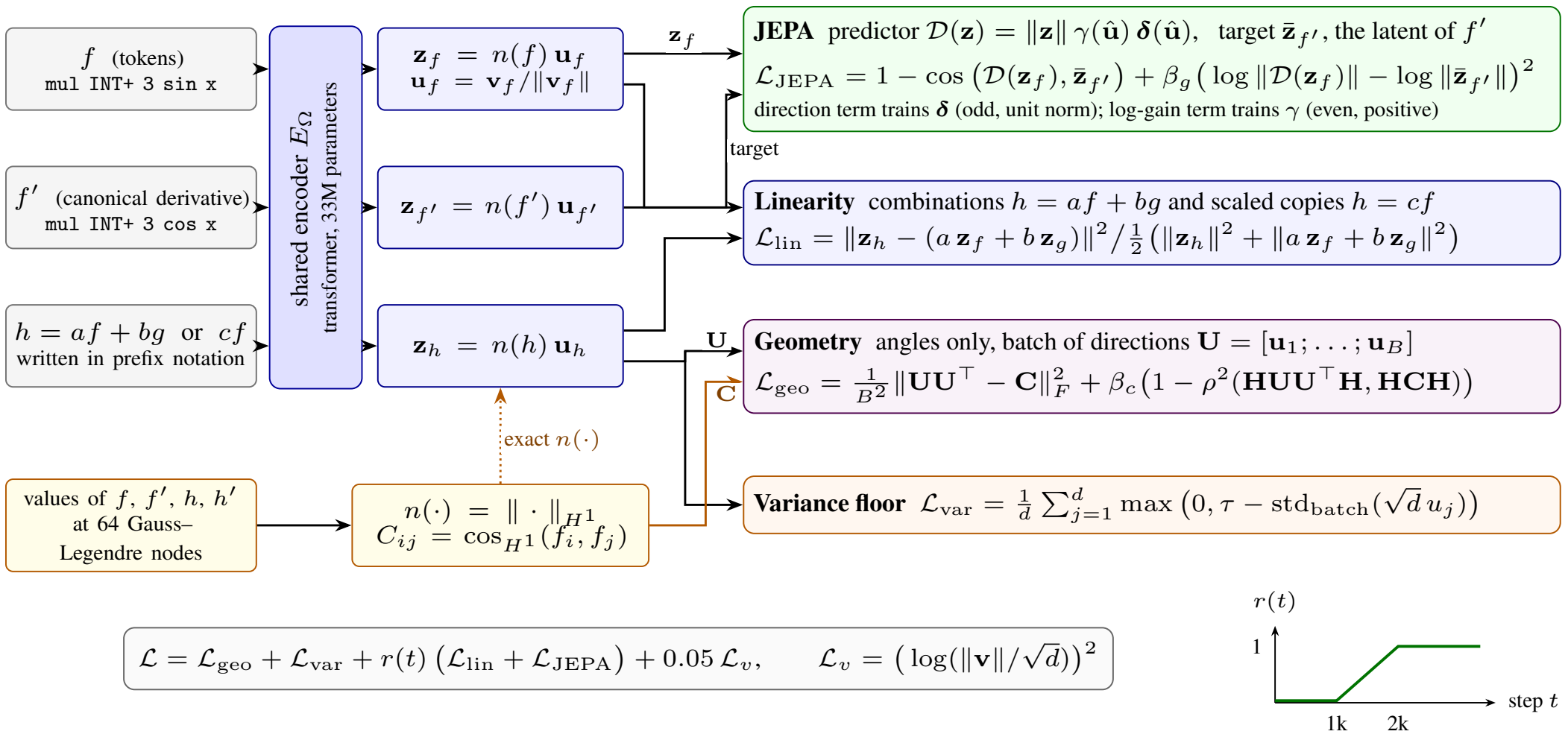

Figure~\ref{fig:objective} summarises how the objective is built.
Let $\uu_i$ be the directions of the $B$ functions of a batch, $\mathbf{U} \in \R^{B\times d}$ the matrix that stacks them, and $C_{ij} = \ip{f_i}{f_j}_{\Hone(\Omega)}/(n(f_i)\,n(f_j))$ the target cosines.
The geometry objective matches latent angles to Sobolev angles and adds a centring term whose maximum is reached by any rank-one latent:
\begin{equation}
\mathcal{L}_{\mathrm{geo}} = \frac{1}{B^2}\sum_{i,j}\big(\uu_i^\top\uu_j - C_{ij}\big)^2 + \beta_c\Big(1 - \rho^2\big(\mathbf{H}\mathbf{U}\mathbf{U}^\top\mathbf{H}, \mathbf{H}\mathbf{C}\mathbf{H}\big)\Big),
\label{eq:geo}
\end{equation}
where $\mathbf{H}$ is the centring matrix, $\rho$ is the Frobenius cosine and $\beta_c = 1.5$.
The first term is the mean squared angular error that bounds the inner-product error in Proposition~\ref{prop:ip}.
Since a freshly initialised transformer maps all inputs to almost parallel directions, a variance floor $\mathcal{L}_{\mathrm{var}} = \frac{1}{d}\sum_{j}\max\big(0,\, \tau - \mathrm{std}(\sqrt{d}\,u_{\cdot j})\big)$ with $\tau = 0.5$ is added, in the spirit of VICReg \cite{bardes2022vicreg}.

The linearity objective is applied to synthetic combinations $h = a f + b g$ and to scaled copies $h = c f$, built directly in prefix notation with exact values:
\begin{equation}
\mathcal{L}_{\mathrm{lin}} = \frac{\|\z_{h} - (a\,\z_f + b\,\z_g)\|^2}{\tfrac{1}{2}\big(\|\z_{h}\|^2 + \|a\,\z_f + b\,\z_g\|^2\big)},
\label{eq:lin}
\end{equation}
which is a relative squared error.

The JEPA objective predicts the latent of the derivative, and is split into a direction term and a log-gain term:
\begin{equation}
\mathcal{L}_{\mathrm{JEPA}} = \Big(1 - \cos\big(\Dop(\z_f), \bar{\z}_{f'}\big)\Big) + \beta_g\Big(\log\big(\|\Dop(\z_f)\| + \epsilon\|\z_f\|\big) - \log\big(\|\bar{\z}_{f'}\| + \epsilon\|\z_f\|\big)\Big)^2,
\label{eq:jepa}
\end{equation}
where the target $\bar{\z}_{f'}$ is the latent of $f'$ with a stop-gradient, $\beta_g = 1$, and $\epsilon = 0.05$ handles constant functions, for which $\z_{f'} = \mathbf{0}$.
The split is motivated by the following observation: for a prediction of norm $s$ at angle $\theta$ from the target $\mathbf{t}$, the squared error is minimised at $s = \|\mathbf{t}\|\cos\theta$, so a mean squared error shrinks the predicted norm whenever the direction is uncertain.
With Equation~\ref{eq:jepa}, the gain is trained independently of the direction.

The full objective is $\mathcal{L} = \mathcal{L}_{\mathrm{geo}} + \mathcal{L}_{\mathrm{var}} + r(t)\big(\mathcal{L}_{\mathrm{lin}} + \mathcal{L}_{\mathrm{JEPA}}\big) + 0.05\,\mathcal{L}_{v}$, where $\mathcal{L}_v = (\log(\|\vv\|/\sqrt{d}))^2$ only keeps the raw projection well scaled for bfloat16 precision.
The ramp $r(t)$ is 0 during the first 1000 steps, so that the geometry forms before the predictive terms act on the latent, and grows linearly to 1 over the next 1000 steps.
Only $f$ and $f'$ are used during pretraining: second derivatives, compositions, and ODEs are never seen, and serve as zero-shot tests.

\subsection{Modules on the frozen encoder}
\label{sec:modules}
Once pretrained, the encoder and $\Dop$ are frozen, and every downstream use of \fjepa{} is built from lightweight modules trained on top of them.
Every module outputs a latent as a unit direction times a magnitude, with the magnitude computed whenever possible and otherwise predicted as a log-gain.
Since the learned operators are accurate only up to a few percent (Section~\ref{sec:exp_geometry}), the modules are employed to propose candidates in latent space, which are then combined, refitted, and verified with exact numerical or symbolic steps.
This principle is shared by all the applications of Section~\ref{sec:experiments}, whose procedures are detailed in Appendix~\ref{app:procedures}.

\paragraph{Decoder}
A 6-layer transformer decoder (17.7M parameters) maps $(\uu, \log n)$ to prefix tokens under a grammar mask, so that every output parses.
Beam candidates are re-ranked by re-encoding them (round trip), and constants can be refined in latent space.

\paragraph{Composition operators}
For each outer function $h$, a network $\Mop_h(\z_g) \approx \z_{h\circ g}$ predicts a direction and a log-gain.
Seven operators are trained ($\sin g$, $\cos g$, $\log g$, $\sqrt{g}$, $g^2$, $g^3$, and $a g^2 + b g + c$), and the scaling laws of their targets, such as $(c\,g)^2 = c^2 g^2$, are satisfied by design for positive scalars (Proposition~\ref{prop:compose} in Appendix~\ref{app:compose_theory}).

\paragraph{Integrator}
$\mathcal{I}(\z) = \|\z\|\,e^{k_I(\hat{\uu})}\,\bm{\delta}_I(\hat{\uu})$ predicts the latent of the zero-mean antiderivative, and is trained with an additional cycle loss through the frozen $\Dop$.

\paragraph{Value encoder}
$\mathcal{V}$ maps 64 samples of $f$ to a unit direction, while the magnitude is computed from a spectral derivative of the samples, so that formulas can be recovered from data with the same decoder.
Further details can be found in Appendix~\ref{app:modules}.

\section{Experiments}
\label{sec:experiments}

The experiments follow the two stages of \fjepa{}: the pretrained representation is evaluated first (Section~\ref{sec:exp_geometry}), then the modules and the four downstream tasks built on the frozen encoder (Sections~\ref{sec:exp_decoding} to~\ref{sec:exp_discovery}), and finally the transfer beyond the pretraining distribution (Section~\ref{sec:exp_transfer}).
The encoder $\Epi$ was pretrained once for 40k steps with batches of 384 functions on a single 8~GB laptop GPU (6.6 hours), and it is never fine-tuned afterwards.
Unless stated otherwise, results are reported on a held-out test set of 12,146 functions that is disjoint from the training data of every component, and retrieval is measured on a fixed subset of 4,000 of them.
Only 9.2\% of the test functions lie within a relative $\Hone$ distance of $10^{-3}$ of a training function, and excluding them leaves the main results unchanged (derivative retrieval 0.835 instead of 0.820).
All models were trained once with a single seed.
Further details can be found in Appendix~\ref{app:details}, and time, memory, and model sizes are compared in Appendix~\ref{app:time}.

\subsection{Latent geometry and differentiation}
\label{sec:exp_geometry}

We first evaluate the pretrained representation itself, before any module is trained on it.
Table~\ref{tab:geometry} is organised in five blocks: whether latent angles and inner products are those of $\Hone$ (geometry), whether the latent of a combination is the combination of the latents (linearity), whether $\Dop$ maps the latent of a function to that of its derivative (derivative), whether relations that were never trained hold (zero-shot), and whether the latent space organises functions in an interpretable way (structure).

\begin{table}[!htb]
\caption{Latent space properties of the encoder $\Epi$ on held-out functions. Errors are median relative errors, and every row is defined in Appendix~\ref{app:metrics}}
\label{tab:geometry}
\centering
\small
\begin{tabular}{llc}
\toprule
 & Metric & \fjepa{} \\
\midrule
\multirow{3}{*}{Geometry} & Correlation of $\cos(\z_f, \z_g)$ with $\cos_{\Hone}(f, g)$ & 0.993 \\
 & Correlation of $\z_f^\top\z_g$ with $\ip{f}{g}_{\Hone}$ & 0.998 \\
 & Mean absolute cosine error & 0.017 \\
\midrule
\multirow{3}{*}{Linearity} & Scale error $\z_{cf}$ vs.\ $c\,\z_f$ ($c = 2$ / $10$ / $-1$) & 0.012 / 0.014 / 0.050 \\
 & Combinations $\z_{af+bg}$ vs.\ $a\,\z_f + b\,\z_g$ & 0.071 \\
 & Random sums $\z_{f+g}$ vs.\ $\z_f + \z_g$ & 0.207 \\
\midrule
\multirow{4}{*}{Derivative} & Retrieval of $f'$, top-1 / top-5 (3,956 candidates) & 0.820 / 0.960 \\
 & Relative error / cosine of $\Dop(\z_f)$ vs.\ $\z_{f'}$ & 0.070 / 0.998 \\
 & Gain ratio $\|\Dop(\z_f)\| / \|\z_{f'}\|$ & 1.004 \\
 & Best linear map, top-1 / top-5 & 0.460 / 0.680 \\
\midrule
\multirow{4}{*}{Zero-shot} & $\Dop(\Dop(\z_f))$ vs.\ $\z_{f''}$, error / top-1 (3,640 candidates) & 0.267 / 0.458 \\
 & $\Dop(\z_{f'})$ vs.\ $\z_{f''}$, error / top-1 & 0.232 / 0.501 \\
 & Eigenfunctions $\Dop\z = a\z$ ($e^{ax}$) / $\Dop^2\z = -k^2\z$ ($\sin kx$) & 0.122 / 0.174 \\
 & ODE annihilation $\|p(\Dop)\z\|$ relative to a shuffled ODE (chance 1) & 0.209 \\
\midrule
\multirow{3}{*}{Structure} & Coefficient-sweep ordering (Spearman; chance 0.57) & 0.828 \\
 & Distance between equivalent rewritings (relative; random pairs 1) & 0.071 \\
 & Linear probe: family accuracy / values $R^2$ & 0.742 / 0.998 \\
\bottomrule
\end{tabular}
\end{table}

As observed in Table~\ref{tab:geometry}, the latent space is close to an isometric embedding of $\Hone$: latent cosines and dot products match their Sobolev counterparts with correlations of 0.993 and 0.998, and the norm is exact by construction.
The latent is homogeneous to 1.2\% for positive scalars and to 5.0\% for negative ones, and linear combinations of the training type are reproduced to 7.1\%, whereas sums of two unrelated functions, which are longer and further from the training distribution, are reproduced only to 20.7\%.
The encoder $\Eunit$, trained from scratch on $[0, 5]$, reaches a comparable geometry, linearity, and derivative retrieval on the test split of its own interval, although its zero-shot operator algebra is weaker (Appendix~\ref{app:eunit}).

The learned derivative retrieves the true derivative among 3,956 candidates for 82.0\% of the functions (96.0\% within the top five), and its gain is calibrated, with a median ratio of 1.004 between predicted and true norms.
Notably, the best linear map fitted to $\Dop$ retrieves only 46.0\%, so differentiation is not a linear map of this latent space.
Relations that were never trained hold approximately: applying $\Dop$ twice retrieves the true $f''$ for 45.8\% of the functions, the eigenfunction relations of $e^{ax}$ and $\sin kx$ hold to 12\% and 17\%, and the residual $p(\Dop)\z$ of solutions of linear ODEs is five times smaller than for a shuffled ODE.
Without supervision, the latent also orders the members of coefficient sweeps (Spearman 0.83 against 0.57 by chance) and places equivalent rewritings of a function at 7\% of the distance between random pairs.

However, $\Dop$ is accurate only up to a noise floor.
Its median relative error is 0.070 and its smallest error on the validation split is 0.013, above the threshold $\varepsilon = 0.0051$ of Section~\ref{sec:encoder}, so no prediction of any learned operator satisfies the $\varepsilon$-criterion (Table~\ref{tab:eps} in Appendix~\ref{app:epscal}).
Retrieval succeeds nevertheless because the target only has to be the nearest candidate, and the median gap between distinct functions is 0.128.
Therefore, the latent operators are accurate enough to propose candidates but not to identify a function by themselves, which is why every application below verifies their proposals with an exact step.

Table~\ref{tab:embeddings} compares the latent space of \fjepa{} with the representations of two released encoders of symbolic expressions: SNIP \cite{meidani2024snip} and the encoder of the integrator of Lample and Charton \cite{lample2020deep}.
We report the property probes of SNIP with the same protocol for every representation, standard measures of how an embedding preserves the geometry of the functions, including trustworthiness \cite{venna2001trust}, and a derivative probe (Appendix~\ref{app:baselines}).
In order not to favour \fjepa{}, whose training target is the $\Hone$ cosine, the first geometry measure employs the plain cosine of the function values on a uniform grid.

\begin{table}[!htb]
\caption{Comparison of function representations computed from the symbolic text on the same 4,000 test functions. The property probes follow SNIP \cite{meidani2024snip} (normalised error, chance 1), and the derivative probe is fitted on the same 20k pairs for every representation}
\label{tab:embeddings}
\centering
\small
\resizebox{\linewidth}{!}{%
\begin{tabular}{lccccccccccc}
\toprule
 & & \multicolumn{4}{c}{Geometry} & \multicolumn{4}{c}{Property probes (NMSE)} & \multicolumn{2}{c}{Derivative probe} \\
\cmidrule(lr){3-6}\cmidrule(lr){7-10}\cmidrule(lr){11-12}
Representation & Dim. & Value cos. & Dist.\ rank & kNN@10 & Trust. & NCR & Upward. & Avg.\ $y$ & Osc. & Top-5 & Align. \\
\midrule
SNIP, symbolic encoder \cite{meidani2024snip} & 512 & 0.002 & -0.090 & 0.080 & 0.570 & 0.680 & 0.716 & 0.805 & 0.084 & 0.338 & 0.936 \\
Lample and Charton encoder \cite{lample2020deep} & 1024 & 0.012 & 0.133 & 0.087 & 0.601 & 0.319 & \textbf{0.217} & \textbf{0.210} & \textbf{0.044} & \textbf{0.979} & 0.908 \\
\fjepa{} & 256 & \textbf{0.946} & \textbf{1.000} & \textbf{0.900} & \textbf{0.994} & \textbf{0.235} & 0.389 & 0.583 & 0.390 & 0.929 & \textbf{0.997} \\
 
\bottomrule
\end{tabular}}
\end{table}

Only \fjepa{} reflects the geometry of the functions: its cosines follow those of the function values (correlation 0.946, against 0.012 and 0.002), and 90\% of its 10 nearest neighbours are also $\Hone$ neighbours, against fewer than 9\% for the other two encoders.
However, the encoder of Lample and Charton, with 2.4 times as many parameters and a latent of dimension 1,024, predicts three of the four SNIP properties more accurately and supports a more accurate derivative retrieval (top-5 0.979 against 0.929), whereas \fjepa{} predicts the non-convexity ratio best and has the best aligned derivative predictions.
A likely reason is the geometry itself, which places functions of the same family close to each other: 78\% of the retrieval errors of $\Dop$ return a different member of the correct family.
Thus, the representation of \fjepa{} is not the most informative one for every property, but it is the only one in which norms, angles, and linear combinations have an analytical meaning.

\begin{figure}[t]
\centering
\includegraphics[width=0.85\linewidth]{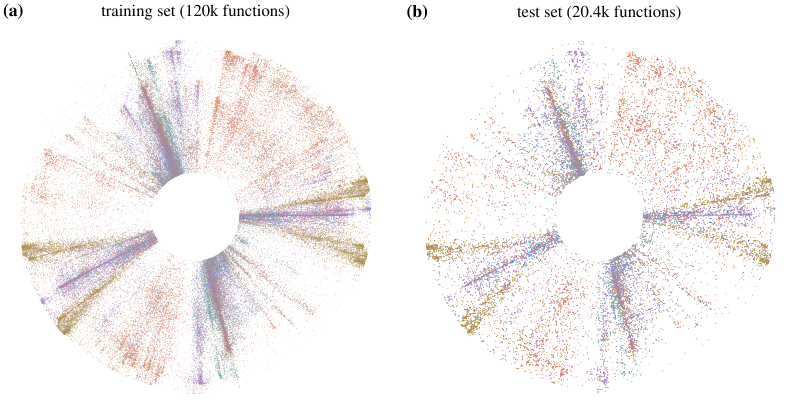}
\caption{Latent space of (a) 120k training functions and (b) the 20.4k functions of the test split of the training corpus. The angle is the code of the unit direction $\uu_f$ given by an autoencoder with a circular bottleneck, the radius is the rank of $\log\|f\|_{\Hone}$, and the colour encodes the elementary functions of each expression}
\label{fig:umap}
\end{figure}

Figure~\ref{fig:umap} shows the global organisation of the latent space, in which the unit directions are projected by a small autoencoder and the exact norm is kept as the radius.
All multiples of a function lie on one ray, oscillatory functions occupy different angular sectors from exponential, power, and rational functions, and the test set covers the same sectors as the training set.
It is worth noting that the projection is lossy, with a median angular error of 33$^\circ$ (Appendix~\ref{app:proj}).

\subsection{Decoding and composition}
\label{sec:exp_decoding}
\begin{figure}[t]
\centering
\includegraphics[width=0.82\linewidth]{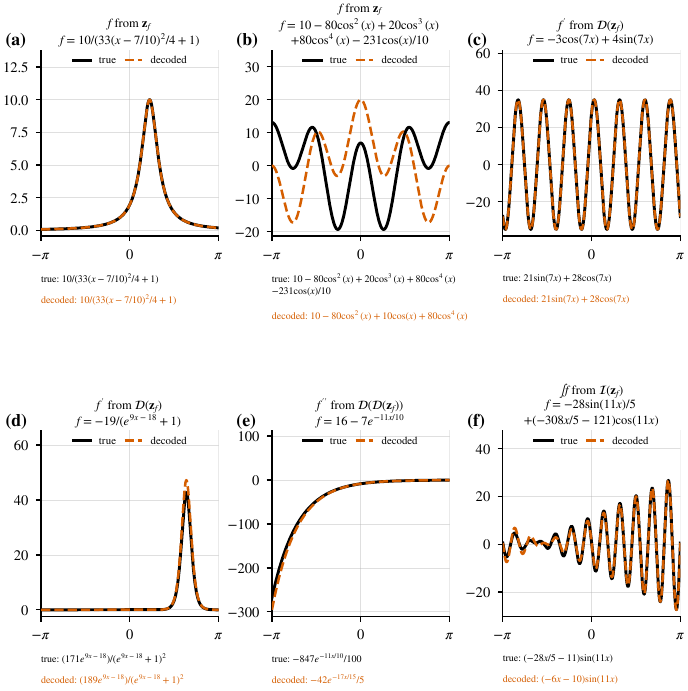}
\caption{True functions (black) against formulas decoded from latents (orange): (a, b) $f$ from $\z_f$, (c, d) $f'$ from $\Dop(\z_f)$, (e) $f''$ from $\Dop(\Dop(\z_f))$, and (f) the antiderivative from $\mathcal{I}(\z_f)$. The examples are taken in a fixed random order, and more are shown in Appendix~\ref{app:gallery}}
\label{fig:predops}
\end{figure}

From here on the encoder and $\Dop$ are frozen.
The decoder recovers the exact function from $\z_f$ for 52.6\% of the test functions with beam search and round-trip re-ranking, and for 67.0\% after latent constant refinement, while the shape of the function (constants refitted) is correct for 84.0\%.
Decoding the predicted latent $\Dop(\z_f)$ yields a derivative obtained entirely through the latent space: its shape is correct for 74.8\% of the functions, but its constants are exact only for 6.8\%, since $\Dop(\z_f)$ inherits the noise floor of $\Dop$.
Accuracy also decreases with expression length, from 77\% for 10 to 19 tokens to 10\% for 30 to 39 tokens (Appendix~\ref{app:decoder}).
Figure~\ref{fig:predops} illustrates both regimes: when decoding fails, the formula usually keeps the structure of the true one with a wrong constant, such as $189$ instead of $171$ in the predicted derivative of a logistic function.

Table~\ref{tab:composition} shows that composition is learnable on the frozen encoder, but not with a linear map: ridge regression retrieves at most 19.6\% of the targets, whereas $\Mop_h$ attains between 56.8\% and 92.2\% top-1 accuracy.
Remarkably, the quadratic $a g^2 + b g + c$ is retrieved with 44.4\% accuracy without any training on it, by combining the learned square operator with the linearity of the latent.
The predicted magnitudes also contribute to retrieval, as shown by the lower accuracy of the direction alone.

\begin{table}[!htb]
\caption{Accuracy comparison (top-1) of composition operators on the frozen encoder, retrieving $\z_{h\circ g}$ among about 3k unseen inner functions $g$. The last row gives the top-5 accuracy of $\Mop_h$}
\label{tab:composition}
\centering
\small
\resizebox{\linewidth}{!}{%
\begin{tabular}{lccccccc}
\toprule
Method & $\sin g$ & $\cos g$ & $\log g$ & $\sqrt{g}$ & $g^2$ & $g^3$ & $a g^2 + b g + c$ \\
\midrule
Identity $\z_g$ & 0.099 & 0.034 & 0.017 & 0.023 & 0.016 & 0.038 & 0.005 \\
Ridge regression & 0.032 & 0.021 & 0.041 & 0.019 & 0.096 & 0.196 & 0.118 \\
Linear part of $\Mop_h$ only & 0.095 & 0.000 & 0.032 & 0.201 & 0.149 & 0.682 & 0.000 \\
Zero-shot $a\,\Mop_{\mathrm{sq}}(\z_g) + b\,\z_g + c\,\z_1$ & -- & -- & -- & -- & -- & -- & 0.444 \\
$\Mop_h$, direction only & 0.578 & 0.508 & 0.481 & 0.253 & 0.703 & 0.675 & 0.660 \\
$\Mop_h$ & \textbf{0.690} & \textbf{0.647} & \textbf{0.584} & \textbf{0.568} & \textbf{0.922} & \textbf{0.919} & \textbf{0.796} \\
\midrule
$\Mop_h$, top-5 & 0.922 & 0.894 & 0.890 & 0.956 & 0.990 & 0.983 & 0.885 \\
\bottomrule
\end{tabular}}
\end{table}

A natural question is whether the learned operators add anything over decoding the input and applying computer algebra.
Table~\ref{tab:symcontrol} in Appendix~\ref{app:decoder} compares each learned operator with a symbolic path that employs the same frozen decoder and encoder but replaces the learned operator by the exact one, with both routes scored by retrieval among the same candidates.
The symbolic path has the higher top-1 accuracy on every operator, since it is almost always correct when the decoder reconstructs the input exactly.
However, on the 24\% to 45\% of the inputs that the decoder does not reconstruct, the learned operators are more accurate in top-5 for the derivative and for 6 of the 7 composition operators, the exception being the quadratic.
Hence, the learned operators do not replace exact symbolic calculus on formulas that can be recovered, but they complement it, and they are the only option when no formula is available, as in the tasks below.

\subsection{Symbolic integration}
\label{sec:exp_integration}
The four downstream tasks reuse the same frozen encoder, decoder, and operators, and each adds at most one lightweight module.
The integrator $\mathcal{I}$ retrieves the antiderivative among 11,196 candidates with a top-1 accuracy of 67.0\% (91.9\% top-5), against 42.7\% for a search that inverts $\Dop$ without training.
The solver of Algorithm~\ref{alg:integration} builds on it: latent proposals are decoded, refitted, and combined, and an answer is accepted only when SymPy verifies $F' = f$.
Symbolic preprocessing is limited to splitting sums, reducing trigonometric powers, and substituting inner functions, whereas every integration step is performed by the learned integrator and the decoder.

\begin{table}[!htb]
\caption{Symbolic integration on 300 test integrands and 30 hard elementary integrals, with cumulative levels of our solver. Solve rate: antiderivative verified exactly by SymPy; definite integral: the returned antiderivative, verified or not, reproduces $\int_{-\pi}^{\pi} f\,dx$ within 1\%}
\label{tab:integration}
\centering
\small
\begin{tabular}{lcccc}
\toprule
 & \multicolumn{2}{c}{Solve rate} & \multicolumn{2}{c}{Definite integral} \\
\cmidrule(lr){2-3}\cmidrule(lr){4-5}
Method & Test & Hard & Test & Hard \\
\midrule
Latent shortlists, decoding (beam 16) and constant refitting & 0.773 & 0.167 & 0.957 & 0.900 \\
+ exact combination of decoded pieces and pool functions & 0.787 & 0.200 & 0.977 & 0.900 \\
+ splitting sums into terms & 0.813 & 0.200 & 0.977 & 0.900 \\
+ residual peeling in latent space & 0.817 & 0.200 & 0.983 & 0.900 \\
+ latent tabular integration by parts & 0.827 & 0.433 & 0.983 & 0.967 \\
+ substitution and trigonometric reduction & 0.917 & 0.667 & 0.993 & 1.000 \\
\midrule
Lample and Charton, FWD+BWD+IBP, beam 10 \cite{lample2020deep} & 0.687 & 0.900 & -- & -- \\
SymPy \texttt{integrate} (reference, not a learned method) & 0.993 & 1.000 & -- & -- \\
\bottomrule
\end{tabular}
\end{table}

As observed in Table~\ref{tab:integration}, the solver verifies an antiderivative for 91.7\% of the test integrands and 66.7\% of the hard integrals, which require integration by parts or substitution, against 99.3\% and 100\% for SymPy.
The pretrained integrator of Lample and Charton \cite{lample2020deep} solves 68.7\% of our test integrands, whose coefficients often exceed the small integers of its training expressions, but 90.0\% of the hard integrals.
The unverified answers remain close approximations: the returned antiderivative reproduces the definite integral on $[-\pi, \pi]$ within 1\% for 99.3\% of the test integrands and for all the hard ones.
For integrands without an elementary antiderivative, such as $e^{-x^2}$ or $e^{\sin x}$, the derivative of a five-term approximation built from the same basis is within 1\% of the integrand for 7 of 12 functions (Table~\ref{tab:nonelem} in Appendix~\ref{app:results}).

\subsection{Symbolic regression}
\label{sec:exp_symreg}
Formulas are recovered from 64 samples with the value encoder, the decoder, constant refitting, and a selection by $\chi^2$ plus description length (Algorithm~\ref{alg:symreg}).
On 300 test functions, 80.0\% of the formulas are recovered exactly from clean samples, and 63.0\% and 44.7\% at relative noise levels of $10^{-3}$ and $10^{-2}$, against 87.0\% when the true latent is decoded instead of the output of the value encoder (Table~\ref{tab:symreg} in Appendix~\ref{app:sr}).
Table~\ref{tab:ext_sr} compares \fjepa{} with six external methods that receive the same samples of a fixed subset of 50 test functions (Appendix~\ref{app:standard}).

\begin{table}[!htb]
\caption{Symbolic regression from 64 samples on the same 50 test functions for every method, and on 24 classic univariate benchmarks. Entries with a slash give exact recovery / recovery within the function-space threshold, and the columns are defined in Appendix~\ref{app:metrics}}
\label{tab:ext_sr}
\centering
\small
\resizebox{\linewidth}{!}{%
\begin{tabular}{lccccccccc}
\toprule
 & \multicolumn{3}{c}{Exact / within threshold} & \multicolumn{3}{c}{$\sigma = 0$} & & \multicolumn{2}{c}{Classic (of 24)} \\
\cmidrule(lr){2-4}\cmidrule(lr){5-7}\cmidrule(lr){9-10}
Method & $\sigma = 0$ & $10^{-3}$ & $10^{-2}$ & NMSE$<10^{-3}$ & Extrap. & Nodes & Rel.\ err.\ $10^{-2}$ & $\sigma = 0$ & $10^{-2}$ \\
\midrule
E2E & 0.00 / 0.02 & 0.00 / 0.00 & 0.00 / 0.00 & 0.12 & 6.2e-01 & 20 & 1.7e-01 & 0 / 0 & 0 / 0 \\
NeSymReS & 0.06 / 0.14 & 0.00 / 0.12 & 0.00 / 0.08 & 0.22 & 8.0e-01 & \textbf{12} & 2.7e-01 & 3 / 4 & 0 / 2 \\
SNIP & 0.04 / 0.58 & 0.00 / 0.30 & 0.00 / 0.10 & 0.74 & 4.0e-04 & 22 & 6.5e-03 & 0 / 10 & 0 / 2 \\
MMSR & 0.00 / 0.12 & 0.00 / 0.08 & 0.00 / 0.02 & 0.28 & 4.1e-01 & 15 & 3.2e-01 & 3 / 5 & 1 / 3 \\
PySR & 0.06 / 0.74 & 0.00 / 0.58 & 0.00 / 0.18 & 0.94 & 4.0e-07 & 22 & 5.2e-03 & \textbf{9} / 19 & 1 / 4 \\
Operon & 0.06 / \textbf{0.96} & 0.00 / 0.26 & 0.00 / 0.00 & 0.96 & 2.3e-09 & 23 & 7.5e-03 & 3 / \textbf{20} & 0 / 0 \\
\fjepa{} & \textbf{0.76} / \textbf{0.96} & \textbf{0.76} / \textbf{0.92} & \textbf{0.62} / \textbf{0.74} & \textbf{0.98} & \textbf{$<10^{-20}$} & \textbf{12} & \textbf{2.8e-08} & 8 / 10 & \textbf{8} / \textbf{9} \\
 
\bottomrule
\end{tabular}}
\end{table}

\begin{figure}[t]
\centering
\includegraphics[width=0.82\linewidth]{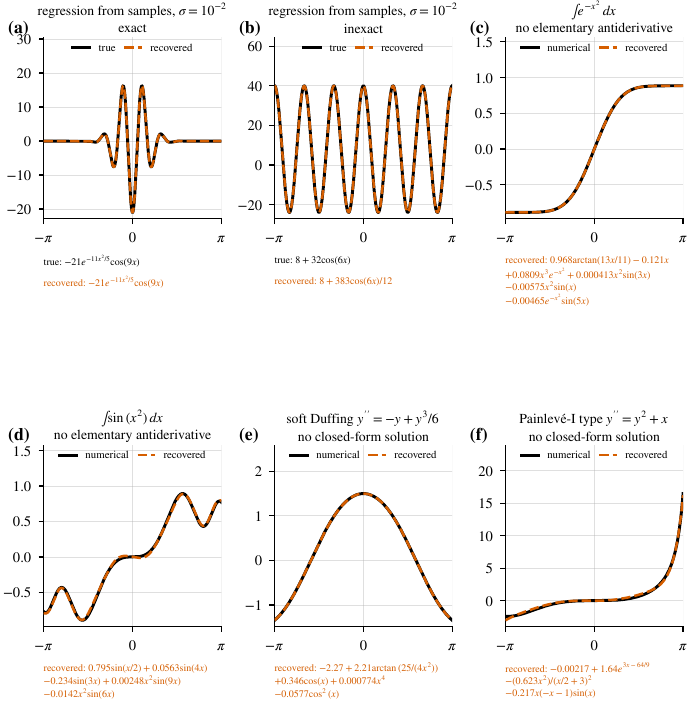}
\caption{True functions (black) against recovered formulas (orange): (a, b) symbolic regression at $\sigma = 10^{-2}$, (c, d) five-term approximations of antiderivatives without elementary closed form, and (e, f) four-atom formulas for initial-value problems without closed form. More examples are shown in Appendix~\ref{app:gallery}}
\label{fig:predapp}
\end{figure}

\fjepa{} is the only method that recovers the exact formula for most functions, and the only one that keeps doing so under noise (76\%, 76\%, and 62\%, against at most 6\% for every other method).
This advantage partly reflects the criterion and the data: the test functions belong to the pretraining families, whose rational constants the decoder emits, whereas the other methods return floating-point constants.
When near-exact formulas are also counted, through a threshold of 0.0037 on the relative $\Hone$ error, Operon ties \fjepa{} on clean samples (0.96 each), and on the 24 classic benchmarks Operon and PySR are within the threshold of 20 and 19 targets against 10 for \fjepa{}.
Under noise, in contrast, the advantage of \fjepa{} remains under both criteria (0.92 and 0.74 against at most 0.58 and 0.18), since the constants of the other methods fit the noise: their median relative error at $\sigma = 10^{-2}$ is between $5.2\cdot10^{-3}$ and 0.32, against $2.8\cdot10^{-8}$ for \fjepa{}.
Moreover, \fjepa{} is the fastest method, with 0.44 seconds per function against 1.6 to 93 seconds (Appendix~\ref{app:time}).
At 1\% noise, its typical failure is a single constant snapped to a wrong fraction, as in Figure~\ref{fig:predapp}b.

\subsection{Solving ordinary differential equations}
\label{sec:exp_odes}
For a linear ODE $p(\tfrac{d}{dx})\,y = 0$, the latent residual $\|p(\Dop)\z\|$ ranks a pool of functions, but true solutions have median relative residuals between 0.009 and 0.08, and gradient descent on $\z$ below this level does not improve the ranking (Table~\ref{tab:odelin} in Appendix~\ref{app:results}).
Following the principle of Section~\ref{sec:modules}, the latent residual is employed to shortlist the top 1\% of a pool of 2M functions, and exact steps on the stored values (re-ranking, sparse combinations, and refitting of constants) followed by SymPy verification produce the solutions (Algorithm~\ref{alg:odes}).
This yields 20 verified solutions among the top 20 for all 9 linear ODEs tested and for 4 nonlinear ones, including solutions absent from the pool (Table~\ref{tab:odec}).
Since SymPy solves such textbook ODEs directly, this experiment validates the principle rather than a use case.

We therefore consider 12 initial-value problems for which SymPy's \texttt{dsolve} fails, such as the pendulum, Duffing oscillators, and a Painlev\'e-I type equation.
A network-free spectral Newton method provides the solution values, and the task is to compress them into a formula with $k$ atoms, measured by the relative deviation from a high-precision numerical solution (Algorithm~\ref{alg:ivp}).
Every learned source of atoms (latent nearest neighbours, latent ODE shortlist, latent residual peeling, and atoms created by the composition operators) is compared with a network-free control that differs only in the learned part.

\begin{table}[!htb]
\caption{Performance comparison on 12 initial-value problems that SymPy cannot solve in closed form, with $k$ atoms. Ratios are medians over the ODEs of the relative deviations from the numerical solution}
\label{tab:odehard}
\centering
\small
\begin{tabular}{lccc}
\toprule
 & $k = 3$ & $k = 4$ & $k = 7$ \\
\midrule
Best learned vs.\ best network-free control: wins / ties / losses & 5 / 1 / 6 & 11 / 0 / 1 & 9 / 1 / 2 \\
Median ratio (control / learned) & 0.96 & 1.43 & 1.40 \\
\midrule
Composed atoms vs.\ latent atoms only: wins / ties / losses & 7 / 5 / 0 & 10 / 1 / 1 & 10 / 2 / 0 \\
Median ratio (latent / composed) & 1.15 & 1.55 & 2.00 \\
\midrule
Best learned vs.\ Pad\'e approximant with $k+1$ coefficients: wins & 12 / 12 & 12 / 12 & 12 / 12 \\
Median error ratio (learned / Pad\'e) & 0.037 & 0.011 & 0.004 \\
\bottomrule
\end{tabular}
\end{table}

As observed in Table~\ref{tab:odehard}, with four atoms the best learned method outperforms the best control on 11 of 12 ODEs (median ratio 1.43), and atoms created by the composition operators improve over latent atoms alone on 10 of 12.
With three atoms, however, learned and network-free methods are equivalent (5 wins, 1 tie, and 6 losses).
All learned formulas are one to two orders of magnitude more accurate than Pad\'e approximants \cite{baker1996pade} with the same number of coefficients.
The per-equation deviations are listed in Table~\ref{tab:odehardfull} in Appendix~\ref{app:results}.
Figure~\ref{fig:predapp}e,f illustrates the two extremes: the four-atom formula is indistinguishable from the solution of the soft Duffing oscillator, whereas for the Painlev\'e-I type equation every four-atom formula deviates by about 5\%.

\subsection{Discovering ordinary differential equations}
\label{sec:exp_discovery}
Table~\ref{tab:discovery} reports the discovery of the governing equation from 64 noisy samples, with the same term search, coefficient snapping, and verification for every route (Algorithm~\ref{alg:discovery}).
The strong-form and weak-form routes estimate the derivatives from the samples and employ no network, whereas the learned route maps the samples to a formula with the value encoder and the decoder, and then differentiates the formula exactly.

\begin{table}[!htb]
\caption{Accuracy comparison (valid@1) on ODE discovery from 64 noisy samples (150 functions per set). $^\dagger$External methods receive the true order of the equation and generous coefficient snapping (Appendix~\ref{app:baselines})}
\label{tab:discovery}
\centering
\small
\begin{tabular}{lcccccc}
\toprule
 & \multicolumn{3}{c}{Linear ODEs} & \multicolumn{3}{c}{Logistic-type ODEs} \\
\cmidrule(lr){2-4}\cmidrule(lr){5-7}
Method & $\sigma = 0$ & $10^{-3}$ & $10^{-2}$ & $\sigma = 0$ & $10^{-3}$ & $10^{-2}$ \\
\midrule
Strong form (SINDy-like, no network) & 0.520 & 0.200 & 0.073 & \textbf{0.853} & 0.513 & 0.200 \\
Weak form (no network) & \textbf{0.673} & 0.233 & 0.053 & 0.473 & 0.380 & 0.127 \\
Samples $\to$ formula $\to$ exact derivatives & 0.487 & \textbf{0.440} & \textbf{0.440} & 0.633 & \textbf{0.647} & \textbf{0.600} \\
\midrule
SINDy, PySINDy \cite{desilva2020pysindy}$^\dagger$ & 0.073 & 0.000 & 0.000 & 0.200 & 0.060 & 0.013 \\
Weak SINDy, PySINDy \cite{messenger2021weaksindy}$^\dagger$ & 0.167 & 0.047 & 0.013 & 0.580 & 0.433 & 0.080 \\
ODEFormer, pretrained \cite{dascoli2024odeformer}$^\dagger$ & 0.047 & 0.020 & 0.020 & 0.033 & 0.040 & 0.027 \\
\bottomrule
\end{tabular}
\end{table}

The learned route is almost insensitive to noise: at 1\% noise it is the most accurate method on both families, 6.0 and 3.0 times more accurate than the best network-free route.
On clean samples, however, the network-free routes are more accurate (0.673 and 0.853 against 0.487 and 0.633), since spectral derivatives of clean samples already identify most equations.
The external methods, which receive the true order of the equation, identify at most 16.7\% of the linear ODEs, whose states require estimated higher derivatives; on logistic-type equations weak SINDy reaches 0.580 on clean samples but drops to 0.080 at 1\% noise.

\subsection{Transfer beyond the pretraining distribution}
\label{sec:exp_transfer}
In order to measure how far the frozen stack transfers, it is evaluated on out-of-distribution expressions and on standard benchmarks, restricted to the items that \fjepa{} can represent (univariate, without poles on $[-\pi, \pi]$, and within our vocabulary; Appendix~\ref{app:standard}).
Every method is evaluated on the same items.

\paragraph{Out-of-distribution expressions}
The out-of-distribution integrands are derivatives of random non-polynomial expression trees with 6 to 8 operators, whereas the training trees have at most 5.
The solver verifies an antiderivative for 44.3\% of them, against 71.8\% for SymPy, and only 17.4\% before the substitution level (Table~\ref{tab:ood_int} in Appendix~\ref{app:results}), which confirms the brittleness of learned integrators outside their training distribution \cite{welleck2022brittleness}.
Notably, 8 of the solved integrals are not solved by SymPy within 10 seconds (Table~\ref{tab:oodnosympy}).
However, these integrands are generated like the backward training data of the integrator of Lample and Charton, which solves 99.3\% of them, so this benchmark is out of distribution only for \fjepa{}.
For symbolic regression on 100 such trees, \fjepa{} recovers 10\% of the formulas exactly from clean samples, against 17\% for PySR, and 7\% at 1\% noise, where PySR recovers none (Table~\ref{tab:ood_sr} in Appendix~\ref{app:sr}).

\paragraph{Standard integration benchmarks}
On the test sets of Lample and Charton, the MIT Integration Bee, and the Rubi suite \cite{rich2018rubi} (Table~\ref{tab:ext_integration}), \fjepa{} is an approximate integrator rather than an exact one.
Between 36.5\% and 76.0\% of its antiderivatives have a derivative within a relative error of $10^{-2}$ of the integrand, but only between 16.5\% and 35.9\% are verified as exact, against 97.5\% to 99.5\% for the integrator of Lample and Charton on its own test sets and 56.5\% and 28.0\% on the other two.
Thus, the integration results of Section~\ref{sec:exp_integration} do not transfer to the distributions on which sequence-to-sequence integrators are trained.

\begin{table}[!htb]
\caption{Symbolic integration on standard benchmarks (200 representable items per benchmark, 92 for MIT Bee). Top: shares of the approximate antiderivatives of \fjepa{} that are close to the true one; bottom: shares verified exactly by SymPy}
\label{tab:ext_integration}
\centering
\small
\resizebox{\linewidth}{!}{%
\begin{tabular}{lccccc}
\toprule
 & \multicolumn{3}{c}{Lample and Charton \cite{lample2020deep}} & & \\
\cmidrule(lr){2-4}
Method & FWD & BWD & IBP & MIT Bee & Rubi \cite{rich2018rubi} \\
\midrule
\multicolumn{6}{l}{\textit{Approximate antiderivatives (only \fjepa{} produces them)}} \\
\fjepa{}, relative error of $F'$ below $10^{-2}$ & 0.700 & 0.530 & 0.760 & 0.500 & 0.365 \\
\fjepa{}, within $\varepsilon$ of the true antiderivative ($e_{\Hone} \le \varepsilon$) & 0.660 & 0.485 & 0.735 & 0.511 & \textbf{0.375} \\
\midrule
\multicolumn{6}{l}{\textit{Exact antiderivatives verified by SymPy}} \\
\fjepa{} & 0.250 & 0.240 & 0.195 & 0.359 & 0.165 \\
Lample and Charton \cite{lample2020deep}, rerun on these items & \textbf{0.975} & \textbf{0.985} & \textbf{0.995} & \textbf{0.565} & 0.280 \\
 
\bottomrule
\end{tabular}}
\end{table}

\paragraph{Standard ODE benchmarks}
The ODE test sets of Lample and Charton are posed as initial-value problems, and \fjepa{} only receives samples of their numerical solution (Table~\ref{tab:ext_ode}).
A solution is counted as verified when its numeric residual is below $10^{-6}$ and it meets the initial conditions, because a symbolic proof rejects 29.5\% and 18.5\% of the reference solutions themselves.
The pretrained solvers of Lample and Charton verify 94.5\% and 91.5\% of the problems and SymPy 62\% and 32\%, whereas \fjepa{} verifies 18.5\% and 16\%, and 25\% and 23\% after fine-tuning the decoder and the value encoder on expressions of this distribution and adding an ODE-aware search.
The gap originates in the pretrained encoder: these solutions combine products and compositions that are absent from the pretraining families, and decoding their true latent recovers only 7.5\% and 4.5\% of them exactly, against 87\% for our test functions (Appendix~\ref{app:odediag}).

\begin{table}[!htb]
\caption{Solving ODEs on the Lample and Charton test sets, posed as initial-value problems (200 each). Verified: numeric residual below $10^{-6}$ and initial conditions met within $10^{-6}$}
\label{tab:ext_ode}
\centering
\small
\resizebox{\linewidth}{!}{%
\begin{tabular}{lcccccc}
\toprule
 & \multicolumn{3}{c}{ODE1} & \multicolumn{3}{c}{ODE2} \\
\cmidrule(lr){2-4}\cmidrule(lr){5-7}
Method & Verified & Res.\ $<10^{-2}$ & Res.\ $<10^{-1}$ & Verified & Res.\ $<10^{-2}$ & Res.\ $<10^{-1}$ \\
\midrule
\fjepa{} (numerical IVP $\to$ samples-to-formula) & 0.185 & 0.240 & 0.305 & 0.160 & 0.220 & 0.280 \\
\fjepa{} + fine-tuned decoder + ODE-aware search & 0.250 & 0.295 & 0.365 & 0.230 & 0.280 & 0.350 \\
Lample and Charton, rerun on these IVPs & \textbf{0.945} & \textbf{0.960} & \textbf{0.965} & \textbf{0.915} & \textbf{0.950} & \textbf{0.950} \\
SymPy \texttt{dsolve} (10 s) & 0.620 & 0.620 & 0.620 & 0.320 & 0.335 & 0.335 \\
 
\bottomrule
\end{tabular}}
\end{table}

On ODEBench \cite{dascoli2024odeformer} (Table~\ref{tab:ext_odebench} in Appendix~\ref{app:results}), the learned discovery route reconstructs 80\%, 87\%, and 89\% of the trajectories at the three noise levels.
It is more accurate than ODEFormer under noise and ties with weak SINDy at 1\% noise, but every SINDy variant is more accurate on clean samples.
In contrast, \fjepa{} recovers the exact equation more often than every other method (15\% of the trajectories on clean samples and 7\% under noise, against at most 7\% and 4\%), although these shares are low in absolute terms.

\section{Limitations}
\label{sec:limitations}

One limitation of our work is that the latent operators are accurate only up to a noise floor of a few percent: no prediction satisfies the calibrated $\varepsilon$-criterion, linear relations between latents hold up to about 7\%, and the learned derivative is not a linear map of the latent space.
Any optimisation in latent space below this level fits the errors of the model rather than the problem, which is why the latent space only proposes candidates and exact numerical or symbolic steps combine and verify them.
Where a closed form exists, computer algebra remains more reliable, and when the decoder recovers the input formula, applying the exact operator to it is more accurate than the learned operator (Table~\ref{tab:symcontrol}).
The transfer beyond the pretraining families is partial: on standard integration and ODE benchmarks \fjepa{} verifies at most 35.9\% and 25\% of the items, far behind sequence-to-sequence models trained on those distributions, and closing this gap would require pretraining the encoder on such expressions (Section~\ref{sec:exp_transfer}).
Its advantage in exact symbolic regression is measured on functions of the pretraining families and narrows on clean samples when near-exact formulas are counted.
Moreover, the study is restricted to univariate functions on a fixed interval, to a 37-token vocabulary without special functions, and to synthetic data, and the decoder rarely reproduces expressions longer than 30 tokens exactly.
Finally, all models were trained once on a single 8~GB laptop GPU, so the variability across training runs and the scaling behaviour are not measured, and Proposition~\ref{prop:ip} controls the inner-product error only through the angular error of the encoder, which is measured rather than bounded.

\section{Conclusions}
\label{sec:conclusions}

In this work, we examined whether a single transformer encoder, pretrained once from symbolic text, can learn a vector space of functions on which calculus acts through latent operators, and whether this representation can be reused across tasks.
The proposed \fjepa{} computes the magnitude of each latent exactly as the $\Hone$ norm and learns only its direction, matching Sobolev angles, enforcing linearity, and predicting the derivative with a factored and exactly homogeneous operator.
The resulting latent space closely reproduces Sobolev angles and inner products, and its operators are accurate enough to propose solutions that exact steps then verify.

Without any fine-tuning of the encoder, the same representation supports symbolic integration, symbolic regression, and the solution and discovery of ODEs, with its clearest gains on noisy data, whereas computer algebra and task-specific models remain stronger on standard benchmarks.
To the best of our knowledge, this is the first joint-embedding predictive architecture trained on the symbolic form of functions in which the predictor is the derivative operator and collapse is prevented by a data-defined Sobolev geometry.

For future work, we would like to extend the approach to multivariate functions with commuting partial derivative operators, to reduce the noise floor of the derivative with structured operators, and to pretrain the encoder on broader distributions of expressions.

{
\small
\bibliographystyle{plainnat}
\bibliography{refs}
}

\appendix
\section{Experimental details}
\label{app:details}

\subsection{Data generation}
\label{app:data}
Functions are generated with SymPy \cite{meurer2017sympy} on $\Omega = [-\pi, \pi]$ and written in prefix notation \cite{lample2020deep} with a 37-token vocabulary:
\begin{itemize}
  \item special tokens;
  \item the binary operators \texttt{add sub mul div pow};
  \item the unary operators \texttt{neg inv sqrt pow2 pow3 exp log sin cos tan atan};
  \item the symbols \texttt{x E pi}, the sign tokens \texttt{INT+ INT-}, the ten digits and a fraction separator.
\end{itemize}
A rational coefficient is a single contiguous literal (\texttt{INT+ 3 / 1 0} for $3/10$), and numeric coefficients come first in products, so that functions that differ in one coefficient differ in one literal.
For instance, $e^{-x^2}\cos 3x$ is written \texttt{mul exp mul INT- 1 pow2 x cos mul INT+ 3 x}.

The families are listed in Table~\ref{tab:families}.
Coefficients are integers between 1 and 10 (50\%), integers between 11 and 100 (30\%) or fractions $p/q$ with $q \in \{2, 3, 4, 5, 10\}$ (20\%), with random signs.
Frequencies are at most 16 and exponential rates at most 2, so that 64 quadrature nodes resolve every function.
The ODE family contains solutions of $\sum_k c_k y^{(k)} = 0$ with a monic integer characteristic polynomial of order 1 to 4, built from integer real roots with multiplicity and complex pairs $a \pm bi$ with $a \in \{-1, 0, 1\}$ and $b \leq 10$; every row satisfies its ODE exactly.
The family of polynomials of a component contains $a g^3 + b g^2 + c g + d$ for a single component $g \in \{x, \sin kx, \cos kx, e^{rx}, \log(x^2+1), \arctan x, 1/(x^2+1)\}$.
Moreover, constants, linear and quadratic functions are guaranteed minimum shares of 1\%, 4\% and 6\%, and the length distribution decreases from 10 tokens on (mean length 19.9 tokens).

\begin{table}[H]
\caption{Composition of the 2.0M training functions by family (\%)}
\label{tab:families}
\centering
\small
\resizebox{\linewidth}{!}{%
\begin{tabular}{lclclc}
\toprule
Family & Share & Family & Share & Family & Share \\
\midrule
linear ODE solutions & 19.3 & Chebyshev series & 5.7 & logistic & 2.7 \\
polynomials of a component & 13.1 & exponential & 4.7 & damped oscillation & 2.7 \\
polynomial & 7.5 & Gaussian & 4.2 & $x\,e^{-bx^2}$ type & 2.5 \\
Fourier sum & 6.4 & Lorentzian & 3.5 & $x^n e^{kx}$ & 2.4 \\
polynomial $\times$ trigonometric & 5.9 & $\arctan$ & 3.5 & random expression trees & 4.2 \\
harmonic & 5.3 & rational & 3.5 & & \\
 & & $\log(1+bx^2)$ & 2.8 & & \\
\bottomrule
\end{tabular}}
\end{table}

A function is accepted under the following conditions:
\begin{itemize}
  \item it is finite on a 1024-point grid and bounded by $10^5$ in absolute value;
  \item its norm agrees between 64 and 128 quadrature nodes (relative tolerance $10^{-3}$);
  \item its $L^2$ norm lies in $[0.05, 2000]$;
  \item it has at most 126 tokens.
\end{itemize}
The derivative $f'$ is stored in its shortest equivalent form.
The test set contains 12,146 functions generated with the same families and filtered so that none of them appears in the training data.
The corpus also has its own validation and test splits (20.4k functions each): the former is used for calibration and model selection, and the latter only in Figure~\ref{fig:umap}.
The composition operators are trained on 4.17M pairs $(g, h\circ g)$, whose inner functions $g$ are drawn from the training split.

\subsection{Encoder training}
\label{app:training}

The hyperparameters are listed in Table~\ref{tab:hparams}.
The networks $h$ and $k$ in Equation~\ref{eq:deriv} have hidden sizes 1024 and 256, respectively, and their gradients are clipped separately from those of the encoder.
Batches are encoded in four chunks of similar length, since random batches contain about 70\% padding; this makes training 4.9 times faster.

\begin{table}[H]
\caption{Hyperparameters of the encoder}
\label{tab:hparams}
\centering
\small
\begin{tabularx}{\linewidth}{@{}lX@{}}
\toprule
Hyperparameter & Value \\
\midrule
Architecture & BERT, 10 layers, hidden size 512, 8 heads, feed-forward 2176, $d = 256$ \\
Parameters & 33M (encoder), 1.6M ($\Dop$) \\
Initialisation & truncated normal (std 0.02), residual outputs scaled by $1/\sqrt{2L}$ \\
Optimizer & AdamW, learning rate $3\cdot10^{-4}$ ($10^{-3}$ for $\Dop$), 700 warm-up steps, cosine \\
Steps / batch size & 40,000 / 384 \\
Batch composition & 15\% equivalence groups, 25\% sweeps, plus 15\% synthetic rows (half combinations, half scaled copies) \\
Combination coefficients & $\pm\{1, 2, 3, 5, 10, \nicefrac{1}{2}\}$ \\
$\lambda$, $\beta_c$, $\tau$, $\beta_g$, $\epsilon$ & 0.5, 1.5, 0.5, 1, 0.05 \\
Ramp of $\mathcal{L}_{\mathrm{lin}}$ and $\mathcal{L}_{\mathrm{JEPA}}$ & 0 until step 1000, linear to 1 at step 2000 \\
Precision & bfloat16 autocast, gradient checkpointing, length-bucketed batches \\
\bottomrule
\end{tabularx}
\end{table}

\subsection{Modules on the frozen encoder}
\label{app:modules}
The decoder has 6 layers, model size 384, 6 heads and 8 memory tokens (17.7M parameters).
It is trained for 105k steps with batches of 32k tokens and a warm-up, stable and decay schedule, with dropout 0.05 and label smoothing 0.02.
Its inputs are randomly rotated (cosine at least 0.99) for 40\% of the rows, and 30\% of the derivative rows use the predicted latent $\Dop(\z_f)$ as input.
The composition bank contains seven networks, trained jointly with disjoint parameters and per-network gradient clipping for 40k steps; their construction is described in Appendix~\ref{app:compose}.
The integrator and the value encoder share the same MLP size, and are trained for 60k and 20k steps, respectively.
The value encoder receives the normalised samples and their spectral derivative (Legendre degree up to 40), noise with $\sigma \leq 0.02$ is added to half of the training batches, and it is fine-tuned with 0.5 times the cross-entropy of the frozen decoder.

\paragraph{Construction of the composition operators}
\label{app:compose}
Each composition operator $\Mop_h$, for $h \in \{\sin, \cos, \log, \sqrt{\cdot}, (\cdot)^2, (\cdot)^3\}$ and for the quadratic $a g^2 + b g + c$, is a separate network on the frozen latent $\z_g$, with $n_g = \|\z_g\|$ and $\hat{\uu} = \z_g/n_g$.
Its output is built in the same factored form as the latent and as $\Dop$:
\begin{equation}
\Mop_h(\z_g) = e^{\kappa_h(\mathbf{x}) + p_h \log n_g}\,\bm{\delta}_h(\mathbf{x}) + a_h\,(\log n_g)\,\z_1, \qquad
\bm{\delta}_h(\mathbf{x}) = \frac{\mathbf{W}_h\hat{\uu} + \mathbf{r}_h(\mathbf{x})}{\|\mathbf{W}_h\hat{\uu} + \mathbf{r}_h(\mathbf{x})\|},
\label{eq:compose}
\end{equation}
where $\z_1$ is the latent of the constant function 1, $\mathbf{W}_h \in \R^{d\times d}$ is a linear skip initialised to the identity, and $(\mathbf{r}_h, \kappa_h)$ are the $d + 1$ outputs of a residual MLP $T_h$ (input layer to width 768, three pre-LayerNorm residual blocks with GELU, expansion factor 2 and dropout 0.1, and a final LayerNorm).
The input $\mathbf{x}$ of $T_h$ and the constants $(p_h, a_h)$ depend on the operator:
\begin{itemize}
\setlength\itemsep{0pt}
\item $g^2$, $g^3$ and $\sqrt{g}$: $\mathbf{x} = \hat{\uu}$, $a_h = 0$ and $p_h = 2$, $3$ and $\nicefrac12$, respectively;
\item $\log g$: $\mathbf{x} = \hat{\uu}$, $p_h = 0$ and $a_h = 1$;
\item $\sin g$ and $\cos g$: $\mathbf{x} = (\hat{\uu}, \ell)$, where $\ell$ is the standardised $\log n_g$, $p_h = 1$ and $a_h = 0$;
\item $a g^2 + b g + c$: $\Mop_{\mathrm{quad}}(\z_g) = \mathbf{b} + \|\mathbf{b}\|\,\mathbf{r}_{\mathrm{quad}}(\hat{\uu}, \ell, \tfrac13(a, b, c))$, with the base $\mathbf{b} = a\,\Mop_{\mathrm{sq}}(\z_g) + b\,\z_g + c\,\z_1$, in which $\Mop_{\mathrm{sq}}$ is evaluated without gradient.
\end{itemize}

The scaling laws that this construction satisfies by design are stated in Proposition~\ref{prop:compose} (Appendix~\ref{app:compose_theory}). Each design choice reproduces a property of the target, so that the network only has to learn what the property does not determine:
\begin{itemize}
\setlength\itemsep{0pt}
\item Equivariance under positive scaling. Since $(c g)^p = c^p g^p$, $\sqrt{c g} = c^{1/2}\sqrt{g}$ and $\log(c g) = \log g + \log c$, and since the latent has an exact norm and is linear up to the scale error of the encoder (Table~\ref{tab:geometry}), the targets satisfy the laws of Proposition~\ref{prop:compose} up to that error. The architecture satisfies them exactly, so these networks learn a function of the direction $\hat{\uu}$ alone, and the logarithm learns its value at unit norm, with training target $\z_{\log g} - (\log n_g)\,\z_1$.
\item No scaling law for $\sin$ and $\cos$. Since $\sin(c g)$ is not determined by $\sin g$, these networks receive $\log n_g$ as an input and predict the gain freely.
\item Sign changes. For $c < 0$ the laws require the parities $(-g)^2 = g^2$, $(-g)^3 = -g^3$, $\sin(-g) = -\sin g$ and $\cos(-g) = \cos g$, which Equation~\ref{eq:compose} does not enforce because $\kappa_h$ and $\bm{\delta}_h$ are not symmetrised as in $\Dop$ (Proposition~\ref{prop:homogeneity}). They are taught by sign augmentation: for a random half of each batch, $\z_g$ is replaced by $-\z_g$ and the target by $-\z_{h\circ g}$ for the odd operators ($\sin$, $g^3$) and by $\z_{h\circ g}$ for the even ones ($\cos$, $g^2$), and $b$ by $-b$ for the quadratic; $\log$ and $\sqrt{\cdot}$, whose inner functions are nonnegative on $\Omega$ in the data, are not augmented. The integrator is augmented in the same way, since integration is odd.
\item Linearity and closure under composition. The quadratic is not learned from scratch: by the linearity of the latent, $\z_{a g^2 + b g + c} \approx a\,\z_{g^2} + b\,\z_g + c\,\z_1$, so it is the composition of the learned square operator with exact latent linear algebra plus a residual scaled by the norm of the base. The base alone retrieves 44.4\% of the targets without any training on the quadratic (Table~\ref{tab:composition}).
\item Factored loss. As for $\Dop$ (Section~\ref{sec:objective}), the direction and the log-gain are trained by separate terms, $(1 - \cos(\bm{\delta}_h, \hat{\mathbf{t}})) + (\kappa_h - \log\|\mathbf{t}\| + p_h\log n_g)^2 + 0.1\,\mathcal{L}_{\mathrm{NCE}}(\bm{\delta}_h, \hat{\mathbf{t}})$ with temperature 0.05, where $\mathbf{t} = \z_{h\circ g} - a_h(\log n_g)\,\z_1$, which avoids the shrinkage of the predicted norm under an uncertain direction; the quadratic uses the relative squared error of its full prediction plus the same contrastive term.
\item Identity initialisation. With $\mathbf{W}_h = \mathbf{I}$ and a residual head initialised at about 1\% of the norm of $\hat{\uu}$, every direction starts at the direction of the inner function, and the linear skip carries the part of the map that is linear in the latent (Table~\ref{tab:composition}, linear part only).
\end{itemize}

\subsection{External baselines}
\label{app:baselines}
All symbolic-regression baselines receive exactly the 64 samples (and noise draws) used by \fjepa{} and are scored with the same criteria; floating-point constants are snapped to simple fractions before the symbolic check, as \fjepa{} does with its own constants.
PySR \cite{cranmer2023pysr} (version 2.7) uses the binary operators $+, -, \times, \div$ and the unary operators $\sin$, $\cos$, $\exp$, $\log$, $\sqrt{\cdot}$, square, cube and $\arctan$, a maximum expression size of 30, a budget of 20 seconds per function, deterministic serial search and its default model selection.
The end-to-end transformer \cite{kamienny2022e2e} is the publicly released pretrained model with the settings of its official example (200 input points, rescaling, refinement of 100 candidate trees by BFGS).
Since its checkpoint is a pickled Python object, it was loaded with a restricted unpickler that resolves only the 42 classes and functions found by a static scan of the file, and two incompatibilities of the official code with NumPy~2 were fixed (an unused import and the alias \texttt{np.infty}).

The integrator of Lample and Charton \cite{lample2020deep} is the released FWD+BWD+IBP model, run with the official code and the inference settings of its example notebook (input $\partial_x y(x) - f(x)$ in its prefix notation, beam 10, at most 200 output tokens); an integrand is solved if any hypothesis $H$ satisfies $H' = f$ with our SymPy check.
The symbolic path of Table~\ref{tab:symcontrol} decodes the input latent (greedy and beam 8, round-trip re-ranking), applies \texttt{sympy.diff} or the composition at token level, writes the derivative in the canonical form of the data (shortest of the raw, \texttt{together} and \texttt{factor\_terms} forms), and re-encodes it with the exact norm.

For ODE discovery, ODEFormer \cite{dascoli2024odeformer} is the released pretrained model with the beam size 50 and temperature 0.1 of its official example, and SINDy \cite{brunton2016sindy} and weak SINDy \cite{messenger2021weaksindy} are the implementations of PySINDy~2.1 \cite{desilva2020pysindy, kaptanoglu2022pysindy}, with the library $\{1, s_i, s_i^2, s_i^3\}$ over the state components $s_i$, the sequentially thresholded least-squares optimiser, and the model chosen among 7 thresholds by the same parsimony-weighted residual as our search.
These methods identify first-order systems of observed states, so for a linear ODE of order $k$ they receive the state $(y, y', \ldots, y^{(k-1)})$, with derivatives estimated by the finite differences of PySINDy (smoothed for noisy data) and with the true order $k$, which our routes have to find.
Moreover, their coefficients are snapped generously: an equation is valid if snapping with any of the denominator bounds of our search makes it exact.
Both choices favour the baselines.

For the representation comparison of Table~\ref{tab:embeddings}, SNIP \cite{meidani2024snip} is the released SNIP-1d-normalized model, of which the symbolic encoder is used, with expressions written in the format of its own generator (binary sums and products, unary powers, all constants in its floating-point encoding).
As a check of this input format, its symbolic and numeric embeddings retrieve each other with 42\% top-1 accuracy among 200 functions of its own generator, but only 3\% among 200 of our test functions, with any input range.
The Lample and Charton representation is the mean of the final encoder states for the integration input.
The probes of Table~\ref{tab:embeddings} are fitted on the same 20k training pairs for every representation: ridge regression with the regularisation chosen on a held-out 10\%, and a 2-layer MLP with 1,024 hidden units trained for 3k Adam steps.
The derivative probe takes $[e/\|e\|, \log\|e\|]$ as input and predicts a unit direction (cosine loss) and the log-ratio $\log(\|e(f')\|/\|e(f)\|)$, with the prediction $\|e(f)\|\,e^{k}\,\mathbf{d}$, so that it is not biased towards small norms.

E-Gen \cite{zheng2025egen} releases code and data but no trained model, so it is not included.
The checkpoints of the Lample and Charton model and of SNIP contain only tensors (and, for SNIP, a path object), so they were loaded with \texttt{weights\_only=True}, whereas the ODEFormer checkpoint is a pickled Python object and was loaded with a restricted unpickler that resolves only the 32 globals found by a static scan.
The official code of ODEFormer and SNIP required the same two NumPy~2 fixes as above.

\subsection{Standard benchmarks}
\label{app:standard}
The standard benchmarks of Section~\ref{sec:exp_transfer} are the test splits of Lample and Charton \cite{lample2020deep} (integration FWD, BWD and IBP; ODE1 and ODE2), the indefinite integrals of the Rubi test suite \cite{rich2018rubi} and of the MIT Integration Bee as collected in the SymPy integration test suites \cite{sympy2026integrationsuites}, and the 23 one-dimensional systems of ODEBench \cite{dascoli2024odeformer}.
Each item is kept only if it can be represented by \fjepa{}: a univariate expression in $x$ (hyperbolic and inverse functions are rewritten exactly with $\exp$, $\log$ and $\arctan$, and decimal constants as rationals), at most 126 tokens of our vocabulary, real and finite on $[-\pi, \pi]$ without poles (no step between neighbouring points of a 20,001-point grid exceeds 5\% of the range), and with a norm in the training range of the encoder.
Table~\ref{tab:ext_coverage} gives the number of items kept and the main reasons for rejection; the benchmarks are scanned in a seed-0 random order, and the first 200 kept items are used.

\begin{table}[H]
\caption{Coverage of the standard benchmarks: representable items kept, items scanned, and the three main reasons for rejection}
\label{tab:ext_coverage}
\centering
\scriptsize
\begin{tabular}{lrrp{8.3cm}}
\toprule
Benchmark & Kept & Scanned & Main rejection reasons (counts) \\
\midrule
L\&C integration FWD & 1000 & 2764 & not finite/real on [-pi, pi] (1096); outside norm range (283); operator outside vocabulary (212) \\
L\&C integration BWD & 1000 & 8203 & not finite/real on [-pi, pi] (3463); too long (1278); operator outside vocabulary (1038) \\
L\&C integration IBP & 1000 & 2098 & not finite/real on [-pi, pi] (580); outside norm range (283); operator outside vocabulary (143) \\
MIT Integration Bee & 92 & 322 & not finite/real on [-pi, pi] (163); outside norm range (35); pole / discontinuity in [-pi, pi] (18) \\
Rubi (indefinite) & 1000 & 58104 & not univariate in x (51687); not finite/real on [-pi, pi] (3570); outside norm range (923) \\
L\&C ODE1 (as IVPs) & 530 & 6523 & not finite/real on [-pi, pi] (3317); operator outside vocabulary (964); pole / discontinuity in [-pi, pi] (517) \\
L\&C ODE2 (as IVPs) & 383 & 4346 & not finite/real on [-pi, pi] (2147); ill-conditioned IVP on [-pi, pi] (617); outside norm range (428) \\
 
\bottomrule
\end{tabular}
\end{table}

The ODE test items of Lample and Charton ask for the general solution of an equation $F(x, y, y', y'') = 0$.
They are converted to initial-value problems by fixing the constants of the reference solution to the first values in $\{1, 2, -1, \tfrac12, 3\}$ (pairs for ODE2) for which the particular solution is representable, and by imposing $y(x_0)$ (and $y'(x_0)$) at the first point $x_0 \in \{0, \tfrac12, 1, -\tfrac12\}$ where the equation can be solved for the highest derivative; for implicit equations, the IVP also selects the branch taken by the reference solution at $x_0$.
Since \fjepa{} works from samples, we keep only the IVPs whose numerical solution (DOP853, relative tolerance $10^{-12}$) reproduces the reference solution on $[-\pi, \pi]$ to $10^{-6}$, which removes equations that are singular inside the interval.
The pretrained solvers of Lample and Charton receive the original equation and return a general solution, whose constants are then solved from the initial conditions with SymPy; SymPy \texttt{dsolve} receives the initial conditions directly.

For symbolic regression, NeSymReS \cite{biggio2021nesymres} is the released 100M model with the inference configuration of its repository (beam search and BFGS constant fitting), SNIP \cite{meidani2024snip} is the released encoder-decoder model with latent space optimisation by Grey Wolf optimisation (population 50, 80 iterations, as in its example), MMSR \cite{li2025mmsr} is the released model with the settings of its benchmark script (run in a separate TensorFlow environment; the released model is bivariate, so the second input is a dummy variable drawn uniformly from $[-5, 5]$), and Operon \cite{burlacu2020operon} uses our operator set and a budget of 20 seconds per function, as PySR.
Every method is given at most 600 seconds per function, after which the function counts as unsolved.
On ODEBench, every method searches first-order autonomous equations $y' = g(y)$, the form of all one-dimensional ODEBench systems: the learned route of \fjepa{} and SINDy use the library $\{1, y, y^2, y^3\}$ with floating-point coefficients (the constants of ODEBench are decimals), ODEFormer \cite{dascoli2024odeformer} receives the samples with their original times, and the derivative-estimation baseline fits PySR to $y' = g(y)$ on smoothed finite differences.
The checkpoints of NeSymReS, SNIP and the ODE solvers of Lample and Charton contain only tensors (and, for NeSymReS, a checkpoint-callback object) and were loaded with \texttt{weights\_only=True} after a static scan; the server that hosts the Lample and Charton files had an expired TLS certificate at the time of download, so their SHA-256 hashes are recorded and the files were only loaded in this restricted mode.
The official code of these baselines required only compatibility fixes for NumPy~2, Python~3.11 and PyTorch~2 (deprecated aliases, a mutable dataclass default and byte masks), which do not change their computations.
As context, Lample and Charton report accuracies of 95.6\% to 99.6\% on their complete integration test sets and general-solution accuracies of 97.2\% (ODE1) and 88.2\% (ODE2) with beam 10; their protocols differ from ours, so these numbers are not included in the tables. The reference solutions pass the numeric verification of Table~\ref{tab:ext_ode} for 97.5\% (ODE1) and 96\% (ODE2) of the IVPs, where the residual is the maximum of $|F(x, y, y', y'')|$ over 256 points of $[-\pi, \pi]$ relative to the size of the terms. ODEFormer reports its ODEBench accuracies only graphically and aggregated over all dimensions, so they are not reproduced.
For the integration benchmarks of Table~\ref{tab:ext_integration}, the answer of \fjepa{} combines, per integrand, the solver of Algorithm~\ref{alg:integration} with a second latent proposal obtained by inverting the derivative operator, $\z^\star = \arg\min_{\z} \lVert \Dop(\z) - \z_f \rVert^2 / \lVert \z_f \rVert^2$, followed by two rounds of refinement on the residual $f - \hat F'$; the answer with the smaller residual $\lVert \hat F' - f \rVert/\lVert f \rVert$ is kept, which does not use the true antiderivative, and these settings were chosen on 150 validation integrands. The second proposal alone verifies far fewer exact antiderivatives than the solver of Algorithm~\ref{alg:integration}, since it lacks the substitution level, so the combination only improves the approximate rows of the table.
The timeout of the ODEFormer simplifier schedules an alarm signal that it never cancels; in the first evaluation this pending alarm shortened the 60-second limit of the next search of the learned route of \fjepa{} to about one second, so that 67 of its 138 searches were interrupted. The harness now cancels any pending alarm after every ODEFormer call, and the learned route was re-evaluated on the same noisy samples; the affected evaluation gave 37\%, 30\%, and 65\% instead of the 80\%, 87\%, and 89\% of Table~\ref{tab:ext_odebench}. Three inference-only variants of the learned route (falling back to the next decoded candidate, refitting the coefficients of the decoded support in the weak form, and a wider beam) were compared on 40 synthetic validation systems, and none was more accurate than the original route, which is therefore kept.

\subsection{Evaluation metrics}
\label{app:metrics}
Scale errors are median values of $\|\z_{cf} - c\,\z_f\|/\|c\,\z_f\|$ with inputs \texttt{mul c f}.
Linearity compares $\z_{af+bg}$ with $a\,\z_f + b\,\z_g$ for training-type combinations, and additivity compares $\z_{f+g}$ with $\z_f + \z_g$ for random pairs.
Derivative retrieval finds the nearest neighbour of $\Dop(\z_f)$ among all distinct test derivatives.
ODE annihilation is the residual $\|p(\Dop)\z\|$ of ODE-family functions relative to that obtained with a shuffled ODE (1 is chance level), and ODE precision at 20 ranks the pool by $r(\z)$.
Decoding is exact when the relative $L^2$ error of the decoded expression is below $10^{-6}$, and the shape is correct when the error is below $10^{-3}$ after refitting all constants to the true values.
For ODE discovery, valid@1 is the fraction of functions whose top-ranked equation is satisfied exactly by the true function.

The rows of Table~\ref{tab:geometry} are defined as follows.
\begin{itemize}
\setlength\itemsep{0pt}
\item Geometry: Pearson correlation, over random pairs of test functions, between the cosine (inner product) of the two latents and the cosine (inner product) of the two functions in $\Hone$, and mean absolute difference between the two cosines; an exact isometry would give 1, 1, and 0.
\item Linearity: the scale error above, the same error for combinations $af + bg$ of two functions of the same family (the type seen in pretraining), and for sums $f + g$ of two unrelated test functions.
\item Derivative: top-1 and top-5 retrieval of $f'$ among 3,956 candidates by nearest neighbour of $\Dop(\z_f)$; relative error and cosine between $\Dop(\z_f)$ and $\z_{f'}$; ratio of their norms; and retrieval with the best linear map fitted to $\Dop$ by ridge regression on 60k training pairs.
\item Zero-shot: second derivatives obtained by applying $\Dop$ twice, or once to $\z_{f'}$, retrieved among 3,640 candidates; relative errors of the eigenfunction relations $\Dop\z = a\z$ for $e^{ax}$ and $\Dop^2\z = -k^2\z$ for $\sin kx$; and ODE annihilation.
\item Structure: Spearman correlation between the order of a swept coefficient and the order of the latents along the sweep; distance between the latents of two rewritings of the same function, relative to random pairs; and accuracy of a linear probe for the family of a function and $R^2$ of a linear probe for its sampled values.
\end{itemize}

In the symbolic-regression comparison of Table~\ref{tab:ext_sr}, a formula is exact when its clean relative error is below $10^{-6}$ and SymPy proves its equality with the target, and it is within the threshold when its relative $\Hone$ error with respect to the clean target is at most 0.0037, the function-space analogue of $\varepsilon$ (Appendix~\ref{app:epscal}), which also counts near-exact formulas with floating-point constants.
The remaining columns are, at $\sigma = 0$, the share of functions with a held-out NMSE below $10^{-3}$, the median NMSE on $[-1.25\pi, 1.25\pi] \setminus [-\pi, \pi]$ (Extrap.; 35 of the 50 formulas of \fjepa{} coincide with the target), and the median number of nodes of the formula, and, at $\sigma = 10^{-2}$, the median relative error with respect to the clean function.
The same function-space threshold is used for the approximate antiderivatives of Table~\ref{tab:ext_integration}, with the constant fixed by $F(-\pi) = 0$.
In Tables~\ref{tab:integration} and~\ref{tab:ood_int}, the definite-integral column counts the integrands for which $|F(\pi) - F(-\pi) - \int_{-\pi}^{\pi} f\,dx| < 10^{-2} \int_{-\pi}^{\pi} |f|\,dx$.
In Table~\ref{tab:symcontrol}, an input is not decoded when the relative $L^2$ error of its decoded formula is above $10^{-6}$, and the evaluation uses 4,000 functions for the derivative and 1,000 inner functions per composition operator.

\subsection{Inference procedures}
\label{app:procedures}
This section specifies the procedures used on the frozen encoder in Section~\ref{sec:experiments}. In them, $r(\z) = \|\Dop^k\z - \sum_j a_j T_j(\z)\| / (\|\Dop^k\z\| + \sum_j |a_j|\,\|T_j(\z)\|)$ is the scale-free latent residual of an ODE $y^{(k)} = \sum_j a_j T_j(y)$, and $\hat{\sigma}$ is the noise level estimated from the flat floor of the Legendre spectrum of the samples.
Every procedure ends with an exact check, either a SymPy proof or a high-precision numerical comparison, and no procedure uses the ground truth to select its output.

\begin{algorithm}[h]
\caption{Symbolic integration on the frozen stack (levels of Table~\ref{tab:integration})}
\label{alg:integration}
\small
\textbf{Input:} integrand $f$ (tokens and values at the 64 nodes). \textbf{Output:} a SymPy-verified antiderivative $F$, or none.
\begin{enumerate}
\setlength\itemsep{0pt}
\item $\z_f \leftarrow E_\Omega(f)$, $\hat{\z}_F \leftarrow \mathcal{I}(\z_f)$; shortlist the 16 training functions $g$ nearest to $\hat{\z}_F$ and the 16 with $\Dop(\z_g)$ nearest to $\z_f$.
\item Decode $\hat{\z}_F$ with beam 16 and refit the constants of every candidate so that $F' = f$ on the nodes.
\item Exact combination: orthogonal matching pursuit (at most 6 atoms) over decoded pieces, shortlisted functions and $\{x, x^2, x^3\}$ for $F' = f$; coefficients snapped to fractions.
\item Term splitting: integrate each term of a sum separately and add the results.
\item Residual peeling (2 rounds): encode the residual $f - \sum_j a_j g_j'$, integrate it with $\mathcal{I}$ and repeat steps 2--3.
\item Integration by parts: for each product $u\,h$ in $f$, build the chain $H_1, H_2, H_3$ of antiderivatives of $h$ with $\mathcal{I}$ and the decoder (constants dropped) and add the atoms $u^{(i)} H_{i+1}$ to a small basis before the full one.
\item Substitution and reduction: reduce powers and products of trigonometric functions to sums; for every inner function $\varphi$ of $f$, rewrite $f/\varphi'$ in $U = \varphi(x)$, integrate it with steps 1--6 and substitute back.
\item Return the first candidate $F$ for which SymPy proves $F' - f = 0$ within the time limit.
\end{enumerate}
\end{algorithm}

\begin{algorithm}[h]
\caption{Symbolic regression from samples}
\label{alg:symreg}
\small
\textbf{Input:} values $y$ at the 64 nodes. \textbf{Output:} a formula.
\begin{enumerate}
\setlength\itemsep{0pt}
\item Estimate $\hat{\sigma}$; compute $\hat{n}$ as the $\Hone$ norm of $y$ with a spectral derivative (low-pass to degree 40 if $\hat{\sigma} > 0$); $\hat{\z} \leftarrow \hat{n}\,\mathcal{V}(y)$.
\item Decode $\hat{\z}$ with beam 16; refit the constants of every candidate to $y$ by weighted least squares and snap them to fractions with denominator at most 12.
\item Exact combination of the candidates with the atoms $\{1, x, x^2\}$.
\item Residual peeling (2 rounds): encode the residual values with $\mathcal{V}$, decode them and add the results to the basis.
\item Return the candidate minimising $\tfrac{n}{2\hat{\sigma}^2}\,\mathrm{err}^2 + 4$ bits per operator $+$ the bits of its constants (for $\hat{\sigma} = 0$, the shortest exact candidate).
\end{enumerate}
\end{algorithm}

\begin{algorithm}[h]
\caption{Solving ODEs with closed-form solutions}
\label{alg:odes}
\small
\textbf{Input:} ODE $y^{(k)} = \sum_j a_j T_j(y)$, pool of 2M functions with latents and stored values. \textbf{Output:} verified solutions.
\begin{enumerate}
\setlength\itemsep{0pt}
\item For every pool function compute $r(\z)$, with $T_j(\z) \in \{\z, \Dop\z, \Dop^2\z, \Mop_{\mathrm{sq}}(\z), \Mop_{\mathrm{cube}}(\z), \Mop_{\mathrm{sqrt}}(\z), \z_1\}$, and keep the 1\% with the lowest residual.
\item Re-rank the shortlist by the exact residual of the ODE on the stored values of $f$, $f'$ and $f''$.
\item Linear ODEs: for all subsets of at most 3 functions among the 16 best and $\{1, x\}$, compute the exact null vector of the residual matrix and snap it to fractions.
\item Refit the constants of the 50 best skeletons so that the ODE holds (scale-free residual below $10^{-9}$, every term active).
\item Return the candidates for which SymPy proves that the ODE holds.
\end{enumerate}
\end{algorithm}

\begin{algorithm}[h]
\caption{Compact formulas for initial-value problems without closed form}
\label{alg:ivp}
\small
\textbf{Input:} ODE, initial conditions, solution values $y_N$ from a spectral Newton method. \textbf{Output:} formulas with $k = 1, \dots, 7$ atoms.
\begin{enumerate}
\setlength\itemsep{0pt}
\item Build a dictionary of atoms. Learned: the 200 nearest latents to $\mathcal{V}(y_N)$, the 200 best functions by the latent ODE residual, decoded composition outputs $\Mop_h(\z_g)$ of the 30 best, and latent residual peeling ($\mathcal{V}$ of the current residual, 16 neighbours and 8 decoded beams). Controls: the 200 nearest functions in value space, 200 random pool functions, and value-space residual peeling.
\item Greedily add the atom that most decreases $10\|y - y_N\|^2 + \text{ODE residual} + \text{initial-condition penalty}$, refit all coefficients by nonlinear least squares, and refine the set by swaps.
\item Report the relative deviation of each formula from a high-precision numerical solution (not used in steps 1--2).
\end{enumerate}
\end{algorithm}

\begin{algorithm}[h]
\caption{ODE discovery from noisy samples (one route)}
\label{alg:discovery}
\small
\textbf{Input:} 64 noisy samples. \textbf{Output:} an equation $y^{(k)} = \sum_j a_j T_j$, $T \in \{y, \dots, y^{(k-1)}, y^2, y^3, 1\}$.
\begin{enumerate}
\setlength\itemsep{0pt}
\item Build the library of the route: spectral derivatives of the samples (strong form), weak-form projections on 40 test functions, or exact derivatives of the formula returned by Algorithm~\ref{alg:symreg} (samples $\to$ formula).
\item Fit all supports of at most 4 terms by least squares, rank them by relative residual, and snap each coefficient to the fraction with the smallest denominator that keeps the residual within 5\%.
\item Return the top-ranked equation, on which valid@1 is scored.
\end{enumerate}
\end{algorithm}

\section{Theoretical results}
\label{app:theory}

\subsection{Proof of Proposition~\ref{prop:homogeneity}}
\label{app:proof}
Consider $\z \neq \mathbf{0}$ and $c \neq 0$, and let $\hat{\uu} = \z/\|\z\|$ and $s = \mathrm{sign}(c)$.
Then $\|c\,\z\| = |c|\,\|\z\|$ and $c\,\z/\|c\,\z\| = s\,\hat{\uu}$.
Since $\gamma(\hat{\uu}) = e^{k(\hat{\uu}) + k(-\hat{\uu})}$ is symmetric in $\hat{\uu}$ and $-\hat{\uu}$, we have $\gamma(s\,\hat{\uu}) = \gamma(\hat{\uu})$.
Since $\A$ is linear and $h(\hat{\uu}) - h(-\hat{\uu})$ is odd, the numerator of $\bm{\delta}$ is odd, and so is its normalisation, $\bm{\delta}(s\,\hat{\uu}) = s\,\bm{\delta}(\hat{\uu})$.
Therefore
\begin{equation*}
\Dop(c\,\z) = |c|\,\|\z\|\,\gamma(\hat{\uu})\,s\,\bm{\delta}(\hat{\uu}) = c\,\|\z\|\,\gamma(\hat{\uu})\,\bm{\delta}(\hat{\uu}) = c\,\Dop(\z),
\end{equation*}
where $|c|\,s = c$. \hfill $\square$

For $c > 0$ only the first step is needed, since then $s = 1$; this is the argument behind the scaling laws of the integrator and of the composition operators (Proposition~\ref{prop:compose}), whose gains and directions are not symmetrised and which therefore satisfy their laws exactly only for positive scalars.

\subsection{Why a finite latent vector can reproduce the Sobolev inner product}
\label{app:ip}
This section answers two questions in plain terms.
First, can a vector of finite dimension behave exactly like a function in the Sobolev space, with dot products equal to Sobolev inner products and linear combinations equal to linear combinations of functions?
Proposition~\ref{prop:oracle} shows that it can: the coefficients of a function on a Sobolev-orthonormal polynomial basis form such a vector, up to a truncation error and a quadrature error that both vanish quickly for smooth functions, so the target of the pretraining is attainable.
Second, how far can the learned encoder be from this ideal?
Proposition~\ref{prop:ip} and Corollary~\ref{cor:dist} show that, because the norm of a latent is computed exactly, the error of every latent dot product and latent distance comes from only two sources: the angle error of the network, which training reduces, and the error of the 64-node quadrature, which is negligible for the functions used here.
Hence, any remaining error of the latent geometry is the angular error of the encoder, which is measured in Appendix~\ref{app:ipcheck}.

Let $\Omega = [a, b]$, let $I(F) = \frac{1}{b-a}\int_a^b F(x)\,dx$, and let $Q_N(F) = \sum_{i=1}^{N} w_i F(x_i)$ be the $N$-node Gauss-Legendre rule mapped to $\Omega$ with weights normalised to $\sum_i w_i = 1$, which is the rule used by the encoder ($N = 64$).
$Q_N$ is exact for polynomials of degree at most $2N - 1$.
Let $t(x) = (2x - a - b)/(b - a)$ map $\Omega$ onto $[-1, 1]$, and let $E_\rho$ denote the open Bernstein ellipse with parameter $\rho > 1$ in the variable $t$.
We use the following classical bound \cite[Theorem~19.3]{trefethen2013atap}, divided by 2 because of the normalised measure.

\paragraph{Quadrature bound}
If $F \circ t^{-1}$ is analytic in $E_\rho$ and bounded there by $M$, then
\begin{equation}
|Q_N(F) - I(F)| \leq \frac{32}{15}\,\frac{M\,\rho^{-2N}}{\rho^2 - 1}.
\label{eq:gaussbound}
\end{equation}

The bound decays geometrically with the number of nodes; for functions that are only $m$ times differentiable the decay is algebraic in $N$ instead.
The data filter of Appendix~\ref{app:data} additionally checks, for every stored function, that its norm agrees between 64 and 128 nodes to a relative tolerance of $10^{-3}$.

\paragraph{Reference embedding}
Let $\phi_0, \dots, \phi_{K-1}$ be the basis of polynomials of degree less than $K$ that is orthonormal in $\ip{\cdot}{\cdot}_{\Hone(\Omega)}$ (Gram-Schmidt applied to $1, x, \dots, x^{K-1}$), let $\Pi_K$ be the orthogonal projection onto their span, $T_K(f) = \|f - \Pi_K f\|_{\Hone(\Omega)}$ the truncation error, $c_k(f) = \ip{f}{\phi_k}_{\Hone(\Omega)}$ the exact coefficients, and
\begin{equation}
\hat{c}_k(f) = Q_N\big(f\,\phi_k + \lambda^2 f'\phi_k'\big), \qquad L_K(f) = \big(\hat{c}_0(f), \dots, \hat{c}_{K-1}(f)\big) \in \R^K,
\label{eq:oracle}
\end{equation}
so that each coordinate of $L_K(f)$ is the discrete quadrature of $\ip{f}{\phi_k}_{\Hone(\Omega)}$, and $L_K$ is linear in $f$.

\begin{proposition}
\label{prop:oracle}
For $f, g \in \Hone(\Omega)$ whose values and derivatives are defined at the nodes, let $e(f) = \|L_K(f) - \mathbf{c}(f)\|$ be the quadrature error of the coefficients. Then
\begin{equation}
\big|L_K(f)^\top L_K(g) - \ip{f}{g}_{\Hone(\Omega)}\big| \leq T_K(f)\,T_K(g) + e(f)\,\|L_K(g)\| + \|f\|_{\Hone(\Omega)}\,e(g).
\label{eq:oraclebound}
\end{equation}
Moreover, $T_K(f) \to 0$ as $K \to \infty$; $e(f) = 0$ if $f$ is a polynomial of degree $p$ with $p + K \leq 2N$; and if $f$ and $f'$ are analytic in $E_\rho$ with $|f\phi_k + \lambda^2 f'\phi_k'| \leq M_k(f)$ there, then $e(f) \leq \frac{32}{15}\frac{\rho^{-2N}}{\rho^2 - 1}\big(\sum_{k<K} M_k(f)^2\big)^{1/2}$.
\end{proposition}

\paragraph{Proof}
Since $f - \Pi_K f$ and $g - \Pi_K g$ are orthogonal to the span of the $\phi_k$, the cross terms vanish in
\begin{equation*}
\ip{f}{g}_{\Hone} = \ip{\Pi_K f}{\Pi_K g}_{\Hone} + \ip{f - \Pi_K f}{g - \Pi_K g}_{\Hone},
\end{equation*}
and by orthonormality $\ip{\Pi_K f}{\Pi_K g}_{\Hone} = \mathbf{c}(f)^\top\mathbf{c}(g)$, which is the Parseval identity on the span.
The Cauchy-Schwarz inequality gives $|\mathbf{c}(f)^\top\mathbf{c}(g) - \ip{f}{g}_{\Hone}| \leq T_K(f)\,T_K(g)$.
Next, $L_K(f)^\top L_K(g) - \mathbf{c}(f)^\top\mathbf{c}(g) = \big(L_K(f) - \mathbf{c}(f)\big)^\top L_K(g) + \mathbf{c}(f)^\top\big(L_K(g) - \mathbf{c}(g)\big)$, whose absolute value is at most $e(f)\,\|L_K(g)\| + \|\mathbf{c}(f)\|\,e(g)$, and $\|\mathbf{c}(f)\| \leq \|f\|_{\Hone}$ by the Bessel inequality.
The triangle inequality gives Equation~\ref{eq:oraclebound}.
Since polynomials are dense in $\Hone(\Omega)$ on a bounded interval, $T_K(f) \to 0$.
If $f$ is a polynomial of degree $p$, the integrands $f\phi_k$ and $f'\phi_k'$ have degree at most $p + K - 1 \leq 2N - 1$, so $Q_N$ is exact and $e(f) = 0$.
Finally, applying Equation~\ref{eq:gaussbound} to each coordinate and summing the squares gives the analytic bound. \hfill $\square$

Since $L_K$ is linear and reproduces the inner product up to Equation~\ref{eq:oraclebound}, it has all the properties sought for the latent space, and for $K \leq d$ it can be embedded in $\R^d$ by padding with zeros.
\paragraph{Learned encoder}
The encoder does not compute $L_K$: its direction is produced by a transformer from the symbolic text.
Its latent (Equation~\ref{eq:latent}) uses the quadrature norm $\hat{n}(f) = Q_N(f^2 + \lambda^2 f'^2)^{1/2}$, computed from the values of $f$ and of its symbolic derivative $f'$ at the nodes, so that no numerical differentiation is involved.
Let $\hat{G}(f, g) = Q_N(f g + \lambda^2 f' g')$ be the quadrature Gram entry, $\hat{C}(f, g) = \hat{G}(f, g)/(\hat{n}(f)\,\hat{n}(g))$ the target cosine of Equation~\ref{eq:geo}, and $\alpha(f, g) = |\uu_f^\top\uu_g - \hat{C}(f, g)|$ the angular error of the encoder.

\begin{proposition}
\label{prop:ip}
For the latent of Equation~\ref{eq:latent},
\begin{equation}
\big|\z_f^\top\z_g - \ip{f}{g}_{\Hone(\Omega)}\big| \leq \hat{n}(f)\,\hat{n}(g)\,\alpha(f, g) + \big|\hat{G}(f, g) - \ip{f}{g}_{\Hone(\Omega)}\big|,
\label{eq:ipbound}
\end{equation}
and if $f g + \lambda^2 f' g'$ is analytic in $E_\rho$ and bounded there by $M_{fg}$, the last term is at most $\frac{32}{15}M_{fg}\,\rho^{-2N}/(\rho^2 - 1)$.
\end{proposition}

\paragraph{Proof}
By Equation~\ref{eq:latent}, $\z_f^\top\z_g = \hat{n}(f)\,\hat{n}(g)\,\uu_f^\top\uu_g = \hat{n}(f)\,\hat{n}(g)\big(\uu_f^\top\uu_g - \hat{C}(f, g)\big) + \hat{G}(f, g)$, since $\hat{n}(f)\,\hat{n}(g)\,\hat{C}(f, g) = \hat{G}(f, g)$ by definition. Subtracting $\ip{f}{g}_{\Hone}$ and applying the triangle inequality gives Equation~\ref{eq:ipbound}, and Equation~\ref{eq:gaussbound} bounds the last term. \hfill $\square$

\begin{corollary}
\label{cor:dist}
With $q(f, g) = |\hat{G}(f, g) - \ip{f}{g}_{\Hone(\Omega)}|$,
\begin{equation*}
\big|\|\z_f - \z_g\|^2 - \|f - g\|^2_{\Hone(\Omega)}\big| \leq 2\,\hat{n}(f)\,\hat{n}(g)\,\alpha(f, g) + q(f, f) + q(g, g) + 2\,q(f, g).
\end{equation*}
\end{corollary}

\paragraph{Proof}
Expand $\|\z_f - \z_g\|^2 = \hat{n}(f)^2 + \hat{n}(g)^2 - 2\,\z_f^\top\z_g$ and $\|f - g\|^2_{\Hone} = \|f\|^2_{\Hone} + \|g\|^2_{\Hone} - 2\ip{f}{g}_{\Hone}$, use $\hat{n}(f)^2 = \hat{G}(f, f)$, and apply Proposition~\ref{prop:ip} to the cross term. \hfill $\square$

Thus, the only term in Equation~\ref{eq:ipbound} that depends on the network is the angular error $\alpha$, whose square, averaged over the pairs of a batch, is the first term of the geometry objective (Equation~\ref{eq:geo}).

\subsection{Scaling laws of the composition operators and of the integrator}
\label{app:compose_theory}
The composition operators and the integrator, whose construction is given in Appendix~\ref{app:compose}, satisfy the scaling laws of their targets exactly for positive scalars.

\begin{proposition}
\label{prop:compose}
For every $\z \neq \mathbf{0}$ and every $c > 0$, $\Mop_{\mathrm{sq}}(c\,\z) = c^2\,\Mop_{\mathrm{sq}}(\z)$, $\Mop_{\mathrm{cube}}(c\,\z) = c^3\,\Mop_{\mathrm{cube}}(\z)$, $\Mop_{\mathrm{sqrt}}(c\,\z) = c^{1/2}\,\Mop_{\mathrm{sqrt}}(\z)$, $\Mop_{\log}(c\,\z) = \Mop_{\log}(\z) + (\log c)\,\z_1$, and the integrator satisfies $\mathcal{I}(c\,\z) = c\,\mathcal{I}(\z)$.
\end{proposition}

\paragraph{Proof}
For $c > 0$, $\|c\,\z\| = c\,\|\z\|$ and $c\,\z/\|c\,\z\| = \hat{\uu}$. For these operators $\mathbf{x} = \hat{\uu}$, so $\kappa_h$ and $\bm{\delta}_h$ are unchanged, and Equation~\ref{eq:compose} gives $e^{\kappa_h + p_h\log(c\,n)}\bm{\delta}_h = c^{p_h}e^{\kappa_h + p_h\log n}\bm{\delta}_h$, while the affine term of the logarithm becomes $(\log c + \log n)\,\z_1$. The integrator $\mathcal{I}(\z) = \|\z\|\,e^{k_I(\hat{\uu})}\bm{\delta}_I(\hat{\uu})$ is the case $p = 1$. \hfill $\square$

\section{Additional experimental results}
\label{app:results}
\subsection{A second encoder trained on $[0, 5]$}
\label{app:eunit}
In order to check that learning a Sobolev space does not depend on the interval, a second encoder $\Eunit$ is trained once, from scratch, with the configuration of $\Epi$ (Table~\ref{tab:hparams}) on 2.0M functions generated on $[0, 5]$, and its checkpoint is selected on its validation split at step 38,000 of 40,000.
Table~\ref{tab:eunit} compares both encoders, each on the held-out test split of its own interval; since the two test sets contain different functions, the differences mix the effect of the encoder with that of the data.
$\Eunit$ matches $\Epi$ in geometry, scaling, linearity and derivative retrieval, and it is even slightly better on several of these metrics (cosine correlation 0.996 against 0.991, scale error at $c = -1$ of 0.037 against 0.054, error on random sums 0.141 against 0.216), so the approach does not depend on the interval $[-\pi, \pi]$. The values of $\Epi$ in Table~\ref{tab:eunit} are computed on a smaller sample than those of Table~\ref{tab:geometry}, hence the small differences between the two tables. It is weaker in the zero-shot operator algebra: the annihilation residual of ODEs is 2.5 times larger, the chain error of $\Dop^2$ is 32\% larger and the effective rank of the directions is lower.
The family probe is also lower, which we attribute to the data rather than to the encoder, since 20\% of the functions generated on $[0, 5]$ are numerically close to linear, against 11\% on $[-\pi, \pi]$. A shorter interval was discarded for the same reason: on $[0, 1]$, 44\% of the generated functions are close to linear and 5.4\% of random pairs have $|\cos| > 0.99$, against 2.4\% for the training data of $\Eunit$ and 0.6\% for that of $\Epi$.

\begin{table}[H]
\caption{$\Epi$ and $\Eunit$, each on the test split of its own interval (1,024 functions for the geometry, scaling and ODE rows, 4,000 for retrieval). Errors are relative}
\label{tab:eunit}
\centering
\small
\begin{tabular}{lcc}
\toprule
Metric & $\Epi$ & $\Eunit$ \\
\midrule
$\Hone$ cosine correlation & 0.991 & 0.996 \\
$\Hone$ inner-product correlation & 0.999 & 1.000 \\
effective rank of $\uu$ & 54.7 & 45.4 \\
scale error, $c = 10$ & 0.015 & 0.012 \\
scale error, $c = -1$ & 0.054 & 0.037 \\
linearity error (combinations) & 0.070 & 0.070 \\
error on random sums & 0.216 & 0.141 \\
derivative retrieval, top-1 (4,000) & 0.820 & 0.825 \\
derivative retrieval, top-5 (4,000) & 0.960 & 0.956 \\
relative error of $\Dop$ (median) & 0.070 & 0.076 \\
gain ratio of $\Dop$ & 1.004 & 0.991 \\
$\Dop^2$ chain error & 0.240 & 0.316 \\
$\Dop(\Dop(\z_f))$, top-1 & 0.458 & 0.398 \\
ODE annihilation & 0.208 & 0.530 \\
eigenfunction error, trigonometric & 0.174 & 0.258 \\
sweep ordering (Spearman) & 0.828 & 0.868 \\
linear probe, family accuracy & 0.742 & 0.445 \\
calibrated $\varepsilon_{\mathrm{rel}}$ (own validation split) & 0.0051 & 0.0038 \\
 
\bottomrule
\end{tabular}
\end{table}

Figure~\ref{fig:umap3d} shows the latent of $\Eunit$ next to that of $\Epi$, with the projection described in Appendix~\ref{app:proj}, trained in the same configuration on 120k training directions of $\Eunit$: no angle term and radius equal to the rank of $\log\|f\|_{\Hone}$.
On the test split, the mean of $1 - \cos(\uu_f, \hat{\uu}_f)$ is 0.182 for $k = 2$ and 0.080 for $k = 3$ (median angular errors of 20$^\circ$ and 10$^\circ$), lower than for $\Epi$ because the directions of $\Eunit$ have a lower effective rank (Table~\ref{tab:eunit}), and the validation family accuracy is 0.367 and 0.448, against 0.431 and 0.518 for $\Epi$.

\subsection{Calibration of the $\varepsilon$-criterion}
\label{app:epscal}
\begin{figure}[H]
\centering
\includegraphics[width=\linewidth]{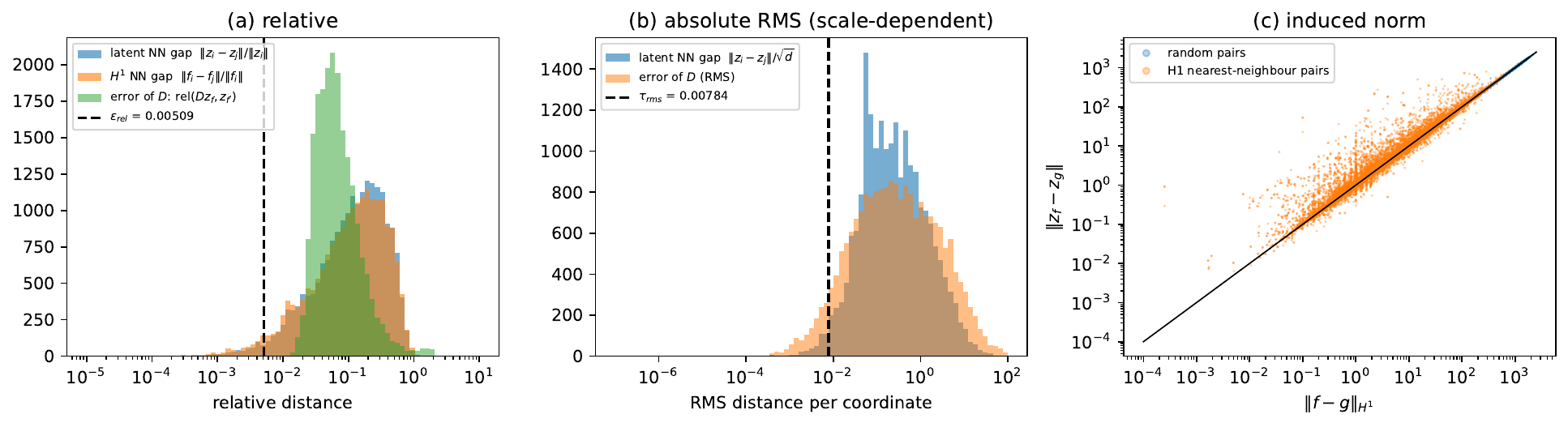}
\caption{Calibration of the $\varepsilon$-criterion on the validation split: (a) relative gaps to the nearest distinct function and relative error of $\Dop$, (b) the same quantities as absolute root mean square distances, and (c) latent against $\Hone$ distance}
\label{fig:epscal}
\end{figure}

Figure~\ref{fig:epscal} shows the data behind the threshold of Section~\ref{sec:encoder}, computed on the 19,765 distinct functions of the validation split.
The relative latent gap to the nearest distinct function and the corresponding $\Hone$ gap have almost the same distribution (medians 0.128 and 0.121), and the latent and $\Hone$ nearest neighbours coincide for 77\% of the functions.
The relative error of $\Dop$ lies between the 5th percentile and the median of these gaps, which is why retrieval succeeds while the $\varepsilon$-ball is never reached.
The absolute root mean square gaps span 2.5 orders of magnitude between their 5th and 95th percentiles, because the norms of the functions do, which rules out a single absolute threshold.
Finally, latent distances reproduce $\Hone$ distances on random pairs (Pearson 0.99997), but only to a median relative error of 7\% on nearest-neighbour pairs, so that very close functions are not resolved as accurately as distant ones.
When the criterion is applied to a formula rather than to a latent, the relative $\Hone$ error of the formula is compared with the threshold obtained by the same rule from the function-space gaps, 0.0037.

Table~\ref{tab:eps} re-scores the predictions of every learned operator with this criterion.
No prediction reaches $\varepsilon$: the smallest relative error of $\Dop$ on the validation split is 0.013, whereas 67.1\% of its predictions are within 0.1 of their target, against a top-1 accuracy of 82.0\%.

\begin{table}[H]
\caption{Closeness of the latents predicted by the learned operators to their targets: top-1 retrieval accuracy, median relative error $\|\hat{\z} - \z\|/\|\z\|$, and shares with a relative error below 0.05 and 0.1. No prediction reaches the calibrated $\varepsilon = 0.0051$}
\label{tab:eps}
\centering
\small
\begin{tabular}{lcccc}
\toprule
Prediction & Top-1 & Rel.\ error & $< 0.05$ & $< 0.1$ \\
\midrule
$\Dop(\z_f)$ vs.\ $\z_{f'}$ & 0.820 & 0.070 & 0.317 & 0.671 \\
$\Dop(\Dop(\z_f))$ vs.\ $\z_{f''}$ & 0.458 & 0.267 & 0.075 & 0.264 \\
$\Dop(\z_{f'})$ vs.\ $\z_{f''}$ & 0.501 & 0.232 & 0.178 & 0.339 \\
Integrator $\mathcal{I}(\z_f)$ & 0.670 & 0.176 & 0.002 & 0.172 \\
\midrule
$\Mop_{\sin}$ & 0.690 & 0.246 & 0.000 & 0.042 \\
$\Mop_{\cos}$ & 0.647 & 0.275 & 0.000 & 0.008 \\
$\Mop_{\log}$ & 0.584 & 0.110 & 0.146 & 0.446 \\
$\Mop_{\sqrt{\cdot}}$ & 0.568 & 0.179 & 0.007 & 0.107 \\
$\Mop_{g^2}$ & 0.922 & 0.168 & 0.001 & 0.135 \\
$\Mop_{g^3}$ & 0.919 & 0.149 & 0.006 & 0.213 \\
$\Mop_{ag^2+bg+c}$ & 0.796 & 0.275 & 0.000 & 0.025 \\
 
\bottomrule
\end{tabular}
\end{table}

\subsection{Numerical check of the inner-product bound}
\label{app:ipcheck}
\begin{figure}[H]
\centering
\includegraphics[width=\linewidth]{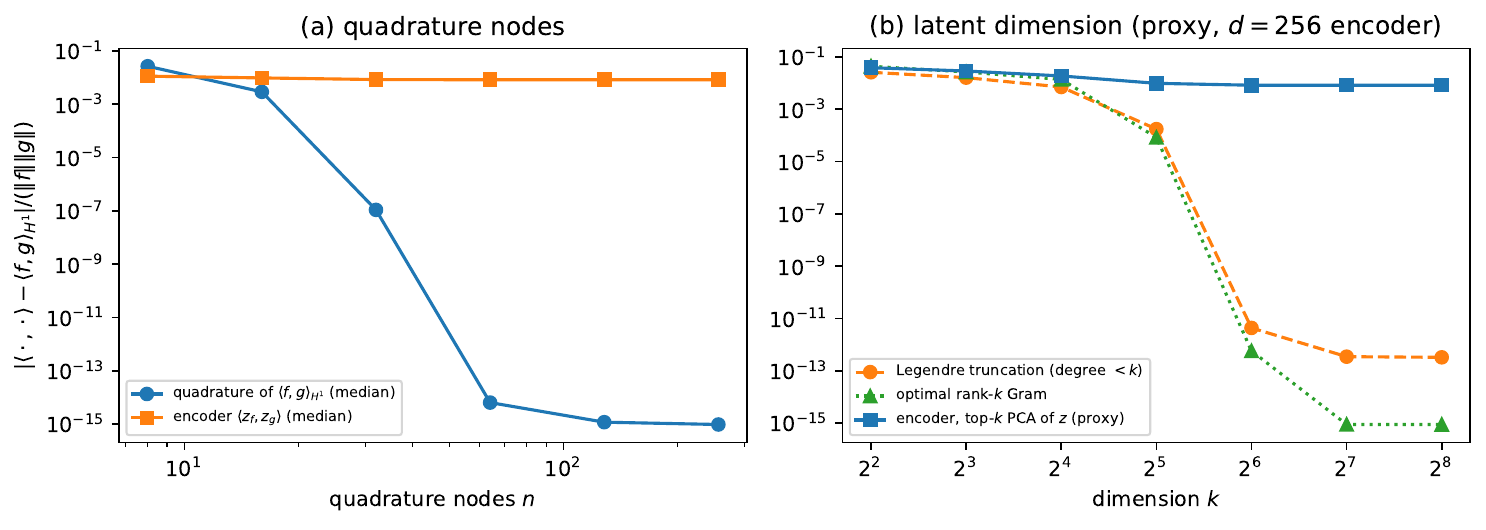}
\caption{Median relative error of the latent inner product on all pairs of 2,000 test functions: (a) against the number $N$ of quadrature nodes, and (b) against the dimension $k$ of a truncated representation}
\label{fig:ipdecay}
\end{figure}

Figure~\ref{fig:ipdecay} measures both terms of Equation~\ref{eq:ipbound} for $\Epi$ on all pairs of 2,000 test functions, against a reference inner product computed with 512 nodes.
The quadrature error decays geometrically with the number of nodes and is negligible from $N = 64$ (median $6\cdot10^{-15}$ relative to $\|f\|_{\Hone}\|g\|_{\Hone}$), whereas the error of the encoder stays at its angular floor (median 0.0083) for every $N \geq 32$, so that the error of the latent inner product is the angular error of the network.
Since no encoder with another latent dimension was trained, panel (b) uses the projection of the latents of $\Epi$ on their top $k$ principal directions as a proxy: the error reaches the floor of the encoder at $k \approx 64$, whereas a polynomial truncation of degree below $k$, as in Proposition~\ref{prop:oracle}, and the optimal rank-$k$ approximation of the Gram matrix reach machine precision at the same dimension, so the 256 dimensions of the latent space are not the limiting factor.

The propositions bound the inner-product error by $\alpha$, but they do not bound $\alpha$ itself, which depends on training and is measured empirically: its mean over held-out pairs is 0.017 (Table~\ref{tab:geometry}).

\subsection{Decoder and module predictions}
\label{app:decoder}
\begin{figure}[h]
\centering
\includegraphics[width=\linewidth]{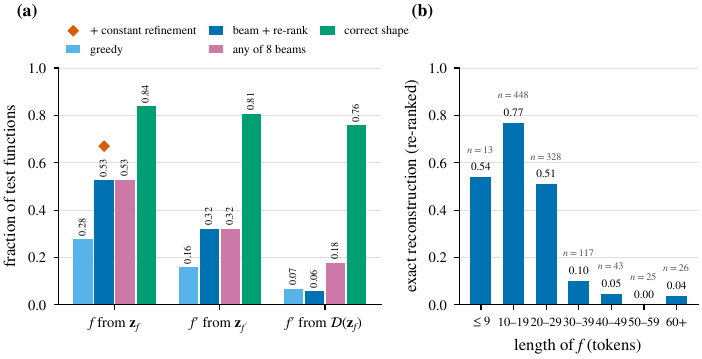}
\caption{Decoder accuracy on the test set: (a) exact reconstruction and correct shape for $f$ from $\z_f$, $f'$ from $\z_{f'}$ and $f'$ from the predicted latent $\Dop(\z_f)$, and (b) exact reconstruction of $f$ by expression length}
\label{fig:decoder}
\end{figure}

Figure~\ref{fig:decoder} and Table~\ref{tab:decexamples} detail the decoder.
Errors are dominated by constants and by long expressions, whereas the function shape is correct in most cases.
Second derivatives decoded from $\Dop(\Dop(\z_f))$ have the correct shape for 39.9\% of the functions, although they are never seen by the encoder or the decoder.

\begin{table}[H]
\caption{Examples of test functions and their decoding from $\z_f$ (beam search with round-trip re-ranking); the last column gives the relative $L^2$ error when the decoding is not exact}
\label{tab:decexamples}
\centering
\small
\resizebox{\linewidth}{!}{%
\begin{tabular}{llc}
\toprule
True function & Decoded function & Error \\
\midrule
$\frac{4 x e^{- \frac{9 x^{2}}{4}}}{3}$ & $\frac{4 x e^{- \frac{9 x^{2}}{4}}}{3}$ & \checkmark \\
$- 21 e^{- \frac{11 x^{2}}{5}} \cos{\left(9 x \right)}$ & $- 21 e^{- \frac{11 x^{2}}{5}} \cos{\left(9 x \right)}$ & \checkmark \\
$3 e^{- \frac{13 x^{2}}{10}} \sin{\left(11 x \right)}$ & $3 e^{- \frac{13 x^{2}}{10}} \sin{\left(11 x \right)}$ & \checkmark \\
$\sin^{3}{\left(x + 5 \right)}$ & $\sin{\left(\sin{\left(x + 5 \right)} \right)}$ & 4.0e-01 \\
$- 10 e^{- \frac{17 \left(x - \frac{1}{5}\right)^{2}}{2}}$ & $- 10 e^{- \frac{17 \left(x - \frac{1}{5}\right)^{2}}{2}}$ & \checkmark \\
$\frac{16 x^{2} e^{- \frac{9 x^{2}}{4}}}{5}$ & $\frac{16 x^{2} e^{- \frac{9 x^{2}}{4}}}{5}$ & \checkmark \\
$\frac{13 x}{5} + 83 \operatorname{atan}{\left(4 x \right)}$ & $3 x + 83 \operatorname{atan}{\left(4 x \right)}$ & 6.4e-03 \\
$- 70 \sin{\left(6 x \right)} + 33 \sin{\left(10 x \right)} + 6 \sin{\left(12 x \right)} + 3 \cos{\left(5 x \right)}$ & $- 66 \sin{\left(6 x \right)} + 34 \sin{\left(10 x \right)} + 3 \cos{\left(3 x \right)}$ & 1.1e-01 \\
$\left(- 7 e^{x} - 20 \sin{\left(3 x \right)}\right) e^{- x}$ & $-7 - 20 e^{- x} \sin{\left(3 x \right)} - \frac{e^{- x} \cos{\left(3 x \right)}}{2}$ & 2.8e-02 \\
$32 \cos{\left(6 x \right)} + 8$ & $32 \cos{\left(6 x \right)} + 8$ & \checkmark \\
 
\bottomrule
\end{tabular}}
\end{table}

\begin{figure}[h]
\centering
\includegraphics[width=\linewidth]{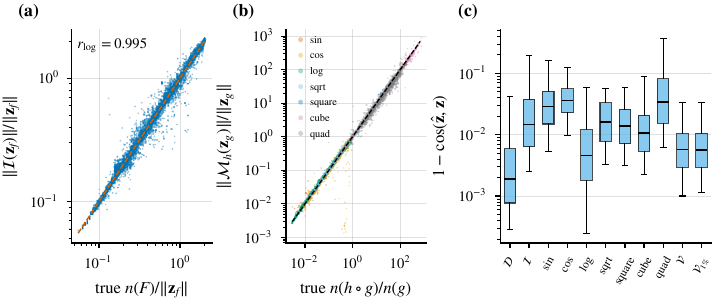}
\caption{Predicted against true quantities on the test sets: (a) gain of the integrator, (b) gain of the composition operators, and (c) distribution of $1 - \cos$ between predicted and true latents for every module}
\label{fig:scalars}
\end{figure}

Figure~\ref{fig:scalars} compares predicted and true quantities for every module.
The integrator predicts the magnitude of the antiderivative with a log-correlation of 0.995 (median absolute log error 0.021), the composition operators follow the true gains over five orders of magnitude, and the median cosine between predicted and true latents is 0.998 for $\Dop$, 0.985 for $\mathcal{I}$, between 0.964 ($\cos g$) and 0.995 ($\log g$) for the composition operators, and 0.994 for $\mathcal{V}$ with and without noise.

Table~\ref{tab:symcontrol} gives the comparison of the learned operators with the symbolic path discussed in Section~\ref{sec:exp_decoding}.

\begin{table}[H]
\caption{Accuracy comparison (top-1 / top-5 retrieval) of the learned operators with a symbolic path that decodes the input latent, applies the exact operator with SymPy, and re-encodes the result. Input not decoded: inputs that the decoder does not reconstruct exactly}
\label{tab:symcontrol}
\centering
\small
\begin{tabular}{lcccc}
\toprule
 & \multicolumn{2}{c}{All inputs} & \multicolumn{2}{c}{Input not decoded} \\
\cmidrule(lr){2-3}\cmidrule(lr){4-5}
Operator & Learned & Symbolic path & Learned & Symbolic path \\
\midrule
$d/dx$ ($\Dop$) & 0.820 / 0.960 & 0.901 / 0.963 & 0.802 / 0.953 & 0.783 / 0.917 \\
$\sin g$ & 0.675 / 0.924 & 0.850 / 0.909 & 0.585 / 0.866 & 0.543 / 0.723 \\
$\cos g$ & 0.640 / 0.889 & 0.849 / 0.905 & 0.528 / 0.792 & 0.502 / 0.686 \\
$\log g$ & 0.591 / 0.893 & 0.773 / 0.886 & 0.509 / 0.831 & 0.498 / 0.747 \\
$\sqrt{g}$ & 0.573 / 0.958 & 0.836 / 0.928 & 0.509 / 0.949 & 0.637 / 0.841 \\
$g^2$ & 0.923 / 0.990 & 0.932 / 0.971 & 0.854 / 0.971 & 0.752 / 0.894 \\
$g^3$ & 0.923 / 0.988 & 0.938 / 0.981 & 0.868 / 0.988 & 0.745 / 0.922 \\
$a g^2 + b g + c$ & 0.801 / 0.897 & 0.942 / 0.963 & 0.688 / 0.822 & 0.784 / 0.862 \\
 
\bottomrule
\end{tabular}
\end{table}

\subsection{Symbolic regression: pipeline levels and additional comparisons}
\label{app:sr}
Table~\ref{tab:symreg} reports the exact recovery of each level of Algorithm~\ref{alg:symreg} on 300 test functions.
The value encoder is robust to noise, with a median cosine similarity of 0.994 to the true latent, and under noise the additional search levels do not change the selected formula, since candidates with more atoms do not pass the description-length criterion.
Excluding the test functions that lie within a relative $\Hone$ distance of $10^{-3}$ of a training function gives 0.80, 0.63, and 0.43 instead of 0.80, 0.63, and 0.45.

\begin{table}[H]
\caption{Exact symbolic recovery (share of functions) of the cumulative levels of our pipeline from 64 samples on 300 test functions, at relative noise levels $\sigma$. All levels use the $\chi^2$ and description-length selection}
\label{tab:symreg}
\centering
\small
\begin{tabular}{lccc}
\toprule
Method & $\sigma = 0$ & $\sigma = 10^{-3}$ & $\sigma = 10^{-2}$ \\
\midrule
Nearest function in the training pool & 0.110 & 0.110 & 0.110 \\
Value encoder + decoding (beam 8) + constant refitting & 0.640 & 0.470 & 0.210 \\
+ beam 16 and refitting of every candidate & 0.730 & \textbf{0.630} & \textbf{0.447} \\
+ exact combination with polynomial atoms & 0.763 & \textbf{0.630} & \textbf{0.447} \\
+ residual peeling & \textbf{0.800} & \textbf{0.630} & \textbf{0.447} \\
\midrule
True latent $\z_f$ instead of $\mathcal{V}$ (upper bound) & 0.870 & -- & -- \\
\bottomrule
\end{tabular}
\end{table}

Table~\ref{tab:symreg_ext} is an earlier comparison on a fixed subset of 100 test functions with PySR \cite{cranmer2023pysr} and the end-to-end transformer \cite{kamienny2022e2e}, with three accuracy criteria.
PySR fits the samples accurately but recovers the exact formula only for 3\% of the clean functions, and the end-to-end transformer, designed for numerical accuracy rather than exact constants, is within 10\% of the target for 37\% to 44\% of the functions.
The exact rates of \fjepa{} differ between the subsets of 300, 100, and 50 functions (Tables~\ref{tab:symreg}, \ref{tab:symreg_ext}, and~\ref{tab:ext_sr}) because of their size; on a rerun of the noisy pipeline, the rate at $\sigma = 10^{-2}$ on the 300 functions changed from 0.447 to 0.473, which gives the order of the run-to-run variation.

\begin{table}[H]
\caption{Symbolic regression from 64 samples on a fixed subset of 100 test functions: share of formulas recovered exactly and with a relative error below $10^{-3}$ and below $10^{-1}$. The last row decodes the true latent of the formula, which does not depend on the samples}
\label{tab:symreg_ext}
\centering
\small
\resizebox{\linewidth}{!}{%
\begin{tabular}{lccccccccc}
\toprule
 & \multicolumn{3}{c}{$\sigma = 0$} & \multicolumn{3}{c}{$\sigma = 10^{-3}$} & \multicolumn{3}{c}{$\sigma = 10^{-2}$} \\
\cmidrule(lr){2-4}\cmidrule(lr){5-7}\cmidrule(lr){8-10}
Method & Exact & $< 10^{-3}$ & $< 10^{-1}$ & Exact & $< 10^{-3}$ & $< 10^{-1}$ & Exact & $< 10^{-3}$ & $< 10^{-1}$ \\
\midrule
E2E transformer \cite{kamienny2022e2e} & 0.00 & 0.00 & 0.39 & 0.00 & 0.00 & 0.44 & 0.00 & 0.00 & 0.37 \\
PySR \cite{cranmer2023pysr} & 0.03 & 0.63 & \textbf{0.98} & 0.00 & 0.66 & \textbf{0.98} & 0.00 & 0.02 & 0.94 \\
\fjepa{} & \textbf{0.77} & \textbf{0.93} & 0.97 & \textbf{0.67} & \textbf{0.86} & \textbf{0.98} & \textbf{0.55} & \textbf{0.57} & \textbf{0.99} \\
\midrule
True latent $\z_f$ instead of $\mathcal{V}$ (upper bound) & 0.84 & 0.93 & 0.97 & -- & -- & -- & -- & -- & -- \\
\bottomrule
\end{tabular}}
\end{table}

Table~\ref{tab:ood_sr} gives the out-of-distribution comparison of Section~\ref{sec:exp_transfer}, on random trees generated like the integrands of Table~\ref{tab:ood_int} but disjoint from them, without poles on $[-\pi, \pi]$.
PySR, whose genetic search is not tied to a training distribution, is the strongest method on clean samples, but at 1\% noise its formulas fit the noise.
On the 24 classic univariate benchmarks (Nguyen \cite{uy2011nguyen}, Keijzer \cite{keijzer2003improving}, and Livermore \cite{mundhenk2021seeding}, the 7 that require a positive domain being composed with an affine map to $[-\pi, \pi]$), PySR recovers 9 formulas exactly on clean samples but a single one at 1\% noise, whereas \fjepa{} recovers 8 at both noise levels.

\begin{table}[H]
\caption{Out-of-distribution symbolic regression from 64 samples on 100 random non-polynomial expression trees with 6 to 8 operators. Shares of formulas recovered exactly and with a relative error below $10^{-3}$ and below $10^{-1}$}
\label{tab:ood_sr}
\centering
\small
\begin{tabular}{lcccccc}
\toprule
 & \multicolumn{3}{c}{$\sigma = 0$} & \multicolumn{3}{c}{$\sigma = 10^{-2}$} \\
\cmidrule(lr){2-4}\cmidrule(lr){5-7}
Method & Exact & $< 10^{-3}$ & $< 10^{-1}$ & Exact & $< 10^{-3}$ & $< 10^{-1}$ \\
\midrule
E2E transformer \cite{kamienny2022e2e} & 0.000 & 0.010 & 0.450 & 0.000 & 0.000 & 0.440 \\
PySR \cite{cranmer2023pysr} & \textbf{0.170} & \textbf{0.680} & \textbf{0.890} & 0.000 & 0.010 & \textbf{0.880} \\
\fjepa{} & 0.100 & 0.250 & 0.610 & \textbf{0.070} & \textbf{0.100} & 0.640 \\
 
\bottomrule
\end{tabular}
\end{table}

\subsection{Per-equation and per-integrand results}
Table~\ref{tab:odelin} to Table~\ref{tab:nonelem} report the per-equation and per-integrand results that are summarised in Section~\ref{sec:experiments}.
For instance, the best approximation of $\int e^{\sin x}\,dx$ is $1.266\,x - 1.266\sin(\cos x)$ plus harmonics, where $1.266 = I_0(1)$ is the mean value of $e^{\sin x}$.

\begin{table}[H]
\caption{Linear ODEs on the held-out pool (11,196 functions): number of solutions in the pool, their median latent residual, precision at 20 of the latent ranking and of the re-ranking after gradient descent in latent space, and SymPy-verified solutions found by exact sparse combinations of the 16 best-ranked functions (total, and absent from the pool)}
\label{tab:odelin}
\centering
\small
\begin{tabular}{lcccccc}
\toprule
ODE & Solutions & Residual & p@20 & p@20 (GD) & Exact & New \\
\midrule
$4\,y + y'' = 0$ & 47 & 0.081 & 0.85 & 0.70 & 8 & 0 \\
$5\,y + 2\,y' + y'' = 0$ & 5 & 0.069 & 0.25 & 0.25 & 17 & 9 \\
$2\,y - 3\,y' + y'' = 0$ & 39 & 0.009 & 0.40 & 0.45 & 6 & 3 \\
$y + y'' = 0$ & 20 & 0.049 & 0.80 & 0.55 & 7 & 0 \\
$-y + y' = 0$ & 5 & 0.013 & 0.20 & 0.25 & 1 & 0 \\
 
\bottomrule
\end{tabular}
\end{table}

\begin{table}[H]
\caption{Textbook ODEs on the 2M-function pool: number of solutions in the pool, precision at 20 of the latent ranking alone, recall of the latent shortlist (top 1\%), verified solutions among the final top 20 (of the maximum possible), and verified solutions absent from the pool}
\label{tab:odec}
\centering
\small
\begin{tabular}{lccccc}
\toprule
ODE & Solutions & Latent p@20 & Recall@1\% & Verified top 20 & New \\
\midrule
$y' = y$ & 244 & 0.00 & 0.89 & 20/20 & 2 \\
$y' = -2y$ & 295 & 0.05 & 0.96 & 20/20 & 1 \\
$y'' = -4y$ & 7,522 & 1.00 & 0.91 & 20/20 & 2 \\
$y'' = -y$ & 10,567 & 1.00 & 0.99 & 20/20 & 0 \\
$y'' = y$ & 7,418 & 0.00 & 0.86 & 20/20 & 2 \\
$y'' = -2y' - 5y$ & 2,835 & 0.90 & 0.96 & 20/20 & 3 \\
$y'' = 3y' - 2y$ & 2,574 & 0.10 & 0.77 & 20/20 & 7 \\
$y'' = 4y' - 4y$ & 629 & 0.00 & 0.64 & 20/20 & 1 \\
$y''' = -y'$ & 19,278 & 1.00 & 0.97 & 20/20 & 0 \\
$y' = y^2$ & 8 & 0.00 & 1.00 & 2/2 & 2 \\
$y' = -y^2$ & 9 & 0.10 & 0.33 & 1/3 & 2 \\
$y' = y - y^2$ & 58 & 0.75 & 1.00 & 20/20 & 4 \\
$y' = 2y - 2y^2$ & 45 & 0.00 & 1.00 & 20/20 & 0 \\
$y' = y + y^2$ & 50 & 0.90 & 1.00 & 20/20 & 1 \\
$y' = y - y^3$ & 7 & 0.00 & 1.00 & 1/1 & 0 \\
$y' = y^3$ & 6 & 0.00 & 0.00 & 0/0 & 1 \\
$y' = \sqrt{y}$ & 40 & 0.60 & 1.00 & 20/20 & 4 \\
$y'' = 2y^3$ & 11 & 0.00 & 0.40 & 2/5 & 1 \\
$y'' = 6y^2$ & 7 & 0.00 & 1.00 & 1/1 & 0 \\
 
\bottomrule
\end{tabular}
\end{table}

\begin{table}[H]
\caption{Relative deviation from the numerical solution of four-atom formulas for the 12 initial-value problems without closed form: best network-free control, latent atoms only, latent and composed atoms, best learned method, and Pad\'e approximant with five coefficients}
\label{tab:odehardfull}
\centering
\small
\begin{tabular}{lccccc}
\toprule
ODE & Control & Latent atoms & + composed & Best learned & Pad\'e \\
\midrule
$y'=\sin y+x$ & 1.4e-2 & 4.0e-2 & 1.7e-2 & 9.5e-3 & 1.0e0 \\
$y'=\cos y-\sin x$ & 8.4e-3 & 2.6e-2 & 5.0e-3 & 5.0e-3 & 1.3e-1 \\
$y'=-y^2+\cos x$ & 3.3e-2 & 4.3e-2 & 4.1e-2 & 2.4e-2 & 3.5e-1 \\
$y'=\sin y+\cos x$ & 1.1e-2 & 3.4e-2 & 1.5e-2 & 7.7e-3 & 6.5e-1 \\
$y'=\sqrt{y}+x$ & 2.8e-3 & 3.8e-3 & 2.4e-3 & 2.4e-3 & 5.9e-3 \\
$y'=\cos y+x$ & 1.2e-2 & 1.7e-2 & 7.8e-3 & 5.4e-3 & 3.9e-1 \\
$y''=-\sin y$ & 5.3e-4 & 4.8e-4 & 6.4e-4 & 4.4e-4 & 4.3e-1 \\
$y''=y^2+x$ & 5.1e-2 & 5.3e-2 & 3.8e-2 & 3.8e-2 & 5.6e-1 \\
$y''=-y+y^3/6$ & 7.9e-4 & 1.3e-4 & 8.7e-5 & 8.7e-5 & 5.6e-1 \\
$y''=\sin y+\cos x$ & 3.4e-4 & 4.2e-4 & 3.6e-4 & 2.0e-4 & 4.6e-1 \\
$y''=-y^3$ & 1.9e-3 & 2.2e-3 & 2.3e-3 & 2.2e-3 & 2.8e-1 \\
$y''=-y-y^3$ & 3.9e-3 & 6.3e-3 & 2.7e-3 & 2.7e-3 & 2.8e-1 \\
 
\bottomrule
\end{tabular}
\end{table}

\begin{figure}[H]
\centering
\includegraphics[width=\linewidth]{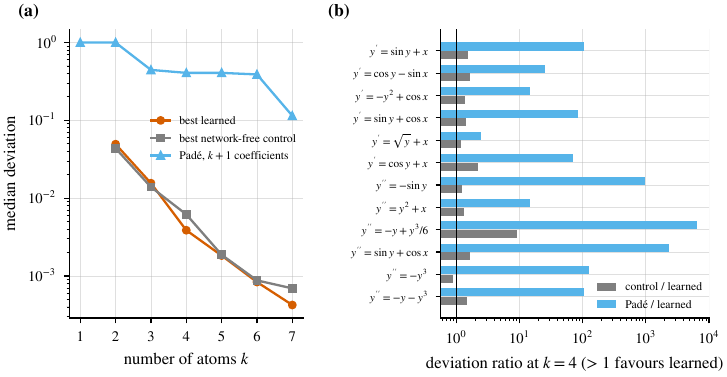}
\caption{Initial-value problems without closed form: (a) median deviation from the numerical solution against the number of atoms $k$, and (b) deviation of the best control and of the Pad\'e approximant divided by that of the best learned method for each ODE at $k = 4$}
\label{fig:odehardk}
\end{figure}

Figure~\ref{fig:odehardk} complements Table~\ref{tab:odehard}: the deviation of the learned formulas decreases steadily with the number of atoms and stays one to three orders of magnitude below that of the Pad\'e approximants for every $k \geq 2$. In median, the learned and network-free dictionaries are close for most $k$, with the clearest advantage of the learned atoms at $k = 4$ and $k = 7$; at $k = 4$ the learned formula is more accurate than the best network-free control on 11 of the 12 equations, and more accurate than Pad\'e on all of them.

As an independent check that does not rely on the numerical reference solution, Table~\ref{tab:hardcheck} reports for the same four-atom formulas (stored with four significant digits) the scale-free ODE residual $\max_x |y^{(k)} - \sum_j a_j T_j| / (\max_x |y^{(k)}| + \sum_j |a_j| \max_x |T_j|)$ over 256 points of $[-\pi, \pi]$ and the error in the initial conditions at $x_0 = 0$.
All formulas satisfy the initial conditions to about $10^{-4}$, and none of them solves the equations exactly, as expected for compact approximations; latent peeling and composed atoms have the smallest residuals (3.9\% and 4.3\%), against 7.2\% for the best network-free control.

\begin{table}[H]
\caption{Residual and initial-condition check of the four-atom formulas for the 12 initial-value problems without closed form (medians over the ODEs; formulas with four significant digits)}
\label{tab:hardcheck}
\centering
\small
\begin{tabular}{lccc}
\toprule
Method & ODEs & Median residual & Median IC error \\
\midrule
Random atoms (control) & 12 & 1.5e-01 & 1.0e-03 \\
Value-space neighbours (control) & 12 & 7.2e-02 & 2.0e-04 \\
Value-space peeling (control) & 12 & 1.0e-01 & 3.2e-04 \\
Latent neighbours & 12 & 1.1e-01 & 2.4e-04 \\
Latent ODE shortlist & 11 & 1.0e-01 & 2.9e-04 \\
Latent peeling & 12 & 3.9e-02 & 9.4e-05 \\
Latent shortlist + atoms & 11 & 1.0e-01 & 2.9e-04 \\
+ composed atoms & 12 & 4.3e-02 & 2.6e-04 \\
 
\bottomrule
\end{tabular}
\end{table}

\begin{table}[H]
\caption{Integrands without elementary antiderivative: relative error of the derivative $F'$ of the $k$-term symbolic approximation with respect to the integrand $f$}
\label{tab:nonelem}
\centering
\small
\begin{tabular}{lccc}
\toprule
Integrand & $k = 1$ & $k = 3$ & $k = 5$ \\
\midrule
$e^{- x^{2}}$ & 7.5e-2 & 1.8e-2 & 2.9e-3 \\
$\sin{\left(x^{2} \right)}$ & 6.7e-1 & 2.9e-1 & 1.1e-1 \\
$\cos{\left(x^{2} \right)}$ & 3.9e-1 & 4.9e-2 & 1.9e-2 \\
$e^{\sin{\left(x \right)}}$ & 1.4e-1 & 7.0e-3 & 1.2e-3 \\
$\frac{\sin{\left(x \right)}}{x^{2} + 1}$ & 1.1e-1 & 4.3e-2 & 1.2e-2 \\
$e^{- x^{2}} \cos{\left(x \right)}$ & 1.2e-1 & 1.7e-2 & 6.4e-3 \\
$\sqrt{x^{4} + 1}$ & 4.2e-2 & 5.8e-3 & 1.8e-3 \\
$e^{\cos{\left(x \right)}}$ & 9.0e-2 & 1.3e-2 & 1.6e-3 \\
$\log{\left(\sin{\left(x \right)} + 2 \right)}$ & 9.9e-2 & 1.7e-2 & 5.6e-3 \\
$e^{\frac{x^{2}}{5}}$ & 2.6e-2 & 1.3e-2 & 1.1e-2 \\
$x^{2} e^{- x^{2}}$ & 5.3e-1 & 1.1e-1 & 4.6e-2 \\
$\frac{\cos{\left(x \right)}}{x^{2} + 1}$ & 7.4e-2 & 8.4e-3 & 3.1e-3 \\
 
\bottomrule
\end{tabular}
\end{table}

Table~\ref{tab:oodnosympy} lists the out-of-distribution integrands of Table~\ref{tab:ood_int} for which the learned levels of our solver return an antiderivative verified by SymPy, whereas SymPy's own \texttt{integrate} returns none within 10 seconds. In every case the antiderivative is a nested composition, which the learned integrator reaches after the substitution or by-parts step has reduced the integrand to a function close to the training families.

\begin{table}[H]
\caption{Out-of-distribution integrals solved by the learned levels of \fjepa{} (verified by SymPy) for which SymPy \texttt{integrate} returns no antiderivative within 10 seconds}
\label{tab:oodnosympy}
\centering
\small
\resizebox{\linewidth}{!}{%
\begin{tabular}{ll}
\toprule
Integrand $f$ & Antiderivative $F$ found by \fjepa{} \\
\midrule
$5 e^{\frac{x}{2}} \sin{\left(5 x \right)} \sin{\left(\cos{\left(5 x \right)} \right)} + \frac{e^{\frac{x}{2}} \cos{\left(\cos{\left(5 x \right)} \right)}}{2}$ & $e^{\frac{x}{2}} \cos{\left(\cos{\left(5 x \right)} \right)}$ \\
$- 15 \left(\cos{\left(\cos{\left(\cos{\left(5 x \right)} \right)} \right)} + 5\right)^{2} \sin{\left(5 x \right)} \sin{\left(\cos{\left(5 x \right)} \right)} \sin{\left(\cos{\left(\cos{\left(5 x \right)} \right)} \right)}$ & $\left(\cos{\left(\cos{\left(\cos{\left(5 x \right)} \right)} \right)} + 5\right)^{3}$ \\
$\left(\left(\tan^{2}{\left(\cos{\left(x \right)} + 3 \right)} + 1\right) \sin{\left(x \right)} + 1\right) e^{x - \tan{\left(\cos{\left(x \right)} + 3 \right)}}$ & $e^{x} e^{- \tan{\left(\cos{\left(x \right)} + 3 \right)}}$ \\
$\frac{\left(12 + \frac{4}{x^{2}}\right) \operatorname{atan}{\left(6 x - \frac{2}{x} \right)}}{\left(6 x - \frac{2}{x}\right)^{2} + 1}$ & $\operatorname{atan}^{2}{\left(6 x - \frac{2}{x} \right)}$ \\
$\left(3 x^{2} + 15\right) \cos{\left(x^{3} + 15 x \right)} \cos{\left(\sin{\left(x^{3} + 15 x \right)} \right)}$ & $\sin{\left(\sin{\left(x \left(x^{2} + 15\right) \right)} \right)}$ \\
$- 3 x^{2} \sin{\left(x^{3} \right)} \cos{\left(x \right)} - \sin{\left(x \right)} \cos{\left(x^{3} \right)}$ & $\cos{\left(x \right)} \cos{\left(x^{3} \right)}$ \\
$\left(- 3 x^{2} - 3 x \left(2 x + 1\right) - 3 x + 12\right) e^{\cos{\left(3 x \left(x^{2} + x - 4\right) \right)}} \sin{\left(3 x \left(x^{2} + x - 4\right) \right)}$ & $e^{\cos{\left(3 x \left(x^{2} + x - 4\right) \right)}}$ \\
$\left(\frac{2 x}{5} + \frac{1}{5}\right) \sin{\left(\sin{\left(\frac{x \left(x + 1\right)}{5} \right)} \right)} \sin{\left(\cos{\left(\sin{\left(\frac{x \left(x + 1\right)}{5} \right)} \right)} \right)} \cos{\left(\frac{x \left(x + 1\right)}{5} \right)}$ & $\cos{\left(\cos{\left(\sin{\left(\frac{x \left(x + 1\right)}{5} \right)} \right)} \right)}$ \\
 
\bottomrule
\end{tabular}}
\end{table}

Table~\ref{tab:ood_int} and Table~\ref{tab:ext_odebench} give the complete results of the out-of-distribution integration and of the ODEBench discovery discussed in Section~\ref{sec:exp_transfer}. On ODEBench, the stricter threshold $R^2 > 0.99$ gives the same ranking of the methods as $R^2 > 0.9$.

\begin{table}[H]
\caption{Out-of-distribution symbolic integration on 149 integrands of random expression trees with 6 to 8 operators, with cumulative levels of our solver. Columns as in Table~\ref{tab:integration}, plus the number of solved integrands that SymPy does not solve}
\label{tab:ood_int}
\centering
\small
\resizebox{\linewidth}{!}{%
\begin{tabular}{lccc}
\toprule
Method & Solve rate & Definite integral & Not solved by SymPy \\
\midrule
Latent shortlists, decoding and constant refitting & 0.101 & 0.544 & 0 \\
+ exact combination, term splitting and residual peeling & 0.121 & 0.664 & 0 \\
+ latent tabular integration by parts & 0.174 & 0.658 & 3 \\
+ substitution and trigonometric reduction & 0.443 & \textbf{0.711} & 8 \\
\midrule
Lample and Charton, FWD+BWD+IBP, beam 10 \cite{lample2020deep} & \textbf{0.993} & -- & \textbf{42} \\
SymPy \texttt{integrate} (reference, 10 s limit) & 0.718 & -- & -- \\
 
\bottomrule
\end{tabular}}
\end{table}

\begin{table}[H]
\caption{Discovery of the 23 one-dimensional ODEBench systems \cite{dascoli2024odeformer} (two initial conditions each) from 64 noisy samples. $R^2$ is measured on the trajectory integrated from the discovered equation, and an equation is exact when it has the true terms with every coefficient within 1\%}
\label{tab:ext_odebench}
\centering
\small
\resizebox{\linewidth}{!}{%
\begin{tabular}{lccccccccccc}
\toprule
 & \multicolumn{3}{c}{$R^2 > 0.9$} & \multicolumn{3}{c}{$R^2 > 0.99$} & \multicolumn{3}{c}{Exact equation} & \multicolumn{2}{c}{Extrapolation NMSE} \\
\cmidrule(lr){2-4}\cmidrule(lr){5-7}\cmidrule(lr){8-10}\cmidrule(lr){11-12}
Method & $\sigma = 0$ & $10^{-3}$ & $10^{-2}$ & $\sigma = 0$ & $10^{-3}$ & $10^{-2}$ & $\sigma = 0$ & $10^{-3}$ & $10^{-2}$ & $\sigma = 0$ & $10^{-2}$ \\
\midrule
\fjepa{}: samples $\to$ formula $\to$ exact derivatives & 0.80 & 0.87 & \textbf{0.89} & 0.67 & 0.78 & 0.67 & \textbf{0.15} & \textbf{0.07} & \textbf{0.07} & 2.4e-03 & \textbf{1.9e-03} \\
SINDy (PySINDy) & 0.96 & 0.93 & 0.48 & 0.80 & 0.67 & 0.20 & 0.07 & 0.02 & 0.00 & 1.0e-05 & 2.0e-01 \\
Weak SINDy (PySINDy) & 0.93 & \textbf{0.96} & \textbf{0.89} & 0.83 & \textbf{0.87} & \textbf{0.70} & 0.07 & 0.04 & 0.00 & 5.8e-06 & 5.7e-03 \\
Estimated derivatives + PySR & \textbf{0.98} & 0.54 & 0.35 & \textbf{0.98} & 0.52 & 0.17 & 0.00 & 0.00 & 0.00 & \textbf{2.2e-08} & 1.1e+01 \\
ODEFormer (pretrained) & 0.85 & 0.85 & 0.83 & 0.67 & 0.67 & 0.61 & 0.00 & 0.00 & 0.00 & 2.7e-03 & 2.3e-03 \\
 
\bottomrule
\end{tabular}}
\end{table}

\subsection{Why few ODE solutions of Lample and Charton are verified}
\label{app:odediag}
Table~\ref{tab:odediag} analyses the 400 initial-value problems of Table~\ref{tab:ext_ode} from their reference solutions and the formulas returned by the samples-to-formula route of \fjepa{} (original decoder, no ODE-aware search) (`numerically exact': relative error below $10^{-6}$ with respect to the reference solution on the 64 nodes, but not proven exact by SymPy).
The reference solutions have the length of our training expressions and use only operators of our vocabulary, so the failures are not due to length or to unknown operators.
Two causes account for them.
First, a quarter to a third of the references contain irrational constants coming from the integration constants of the generator of Lample and Charton (for instance $x^4\sin 9 + 1$ or $-4x + \arctan x - 4 + \cos 4$); \fjepa{} recovers about 40\% of these formulas with a rational approximation of the constant ($\tfrac{107171}{260049}x^4 + 1$, relative error $3\cdot10^{-10}$), which no exact test can accept.
Second, for references with rational constants, about three quarters of the formulas have the wrong shape, because these solutions combine products and compositions of elementary functions that are not among the parameterised families of the pretraining data.
The last row of Table~\ref{tab:odediag} shows that the limitation lies in the pretrained encoder and decoder rather than in the value encoder: when the latent is computed from the symbolic form of the reference solution and given to the same solver, only 7.5\% and 4.5\% of the solutions are recovered exactly (34\% and 32\% within a relative error of $10^{-2}$), against 87\% for the functions of our test set.
Both causes are addressed by the two additions of Table~\ref{tab:ext_ode}: the decoder and the value encoder are fine-tuned on expressions of the Lample and Charton training sets, with the encoder and $\Dop$ frozen, and the selection of a single formula is replaced by an ODE-aware search.
The search keeps the whole candidate pool of the samples-to-formula solver, refits the coefficients of every candidate on the numerical solution, identifies them as exact constants by integer-relation detection (PSLQ) against $e^{\pm k}$, $\sin k$, $\cos k$, $\log k$, $\sqrt{k}$, and $\pi$, solves the initial conditions exactly for one or two coefficients, and ranks the resulting formulas by their residual in the equation.
Under the stricter symbolic proof, the two additions raise the verified shares from 5\% and 5.5\% to 16.5\% and 18\%.
Ranking by the latent residual under $\Dop$ instead of the numeric residual was slightly worse in a preliminary run on 40 IVPs per set, since the noise floor of $\Dop$ cannot separate the exact formula from near-exact ones. The ODE-aware search identifies the irrational constants exactly (for instance it returns $x^4\sin 9 + 1$ and $x^6 + (2 - e^{3})x + 1$), and fine-tuning the decoder on expressions of the Lample and Charton training sets raises the exact recovery from the true latent on the first 40 IVPs of each set from 15\% to 25\% (ODE1) and from 5\% to 8\% (ODE2). On ODE1, the remaining failures of the final route are formulas of the wrong shape (53.5\% of the IVPs) or close to the reference without being exact (21.5\%), and none of them solves the equation with initial conditions of another solution, so the IVP formulation itself is not a source of error.

\begin{table}[H]
\caption{Diagnosis of the IVPs of Table~\ref{tab:ext_ode} (200 per test set): reference solutions and outcome of \fjepa{}; shares within each group of references}
\label{tab:odediag}
\centering
\small
\begin{tabular}{lcc}
\toprule
 & ODE1 & ODE2 \\
\midrule
Median prefix length of the reference solution (pretraining: 18) & 19 & 19 \\
Reference beyond the 99th percentile of pretraining lengths & 0.025 & 0.000 \\
Reference with irrational constants & 0.355 & 0.245 \\
\quad \fjepa{} verified & 0.000 & 0.020 \\
\quad \fjepa{} numerically exact, not provable (rel.\ error $< 10^{-6}$) & 0.394 & 0.408 \\
Reference with rational constants only & 0.645 & 0.755 \\
\quad \fjepa{} verified & 0.078 & 0.066 \\
\quad \fjepa{} wrong shape (rel.\ error $\geq 10^{-2}$) & 0.736 & 0.748 \\
True latent $\z_f$ instead of $\mathcal{V}$: exact recovery & 0.075 & 0.045 \\
 
\bottomrule
\end{tabular}
\end{table}

\subsection{Projections of the latent space}
\label{app:proj}
\begin{figure}[t]
\centering
\begin{tabular}{@{}c@{\hspace{4pt}}c@{}}
$\Epi$ on $[-\pi, \pi]$ & $\Eunit$ on $[0, 5]$ \\
\includegraphics[width=0.49\linewidth]{figures/latent_ae_2d.pdf} & \includegraphics[width=0.49\linewidth]{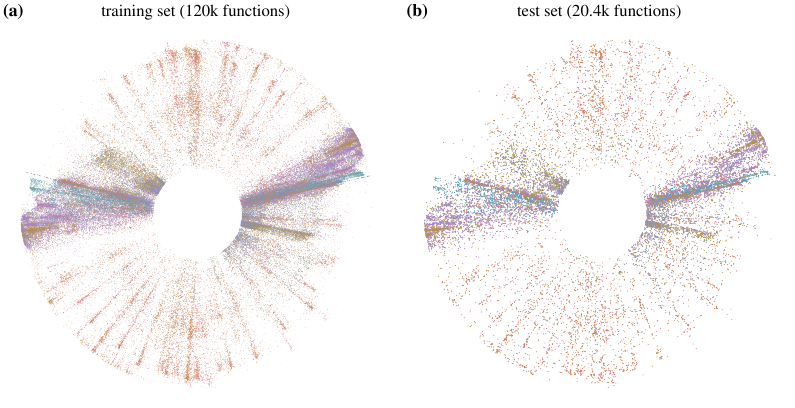} \\
\includegraphics[width=0.49\linewidth]{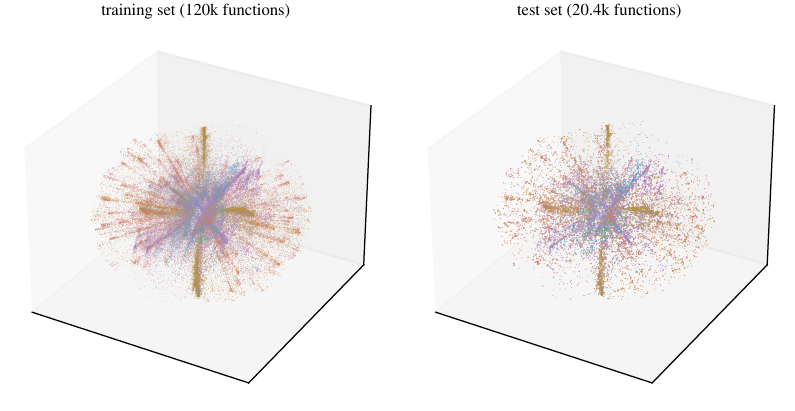} & \includegraphics[width=0.49\linewidth]{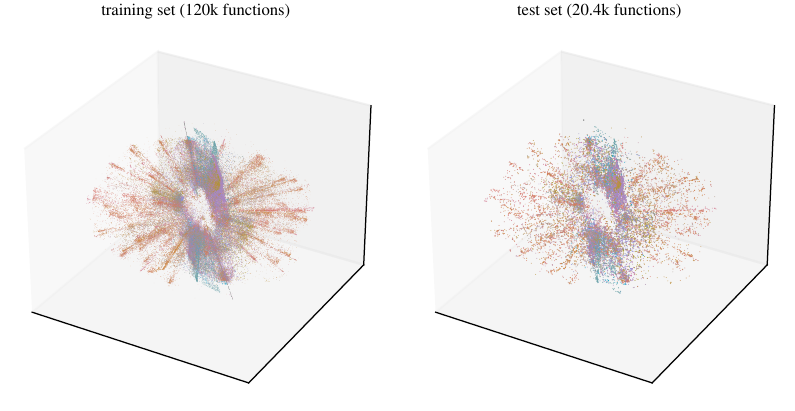} \\
\end{tabular}
\caption{Latents of $\Epi$ (left) and $\Eunit$ (right) projected in two (top) and three (bottom) dimensions, each for (a) 120k training functions and (b) 20.4k test functions of its own interval. The construction and colouring are those of Figure~\ref{fig:umap}, which is the top left panel}
\label{fig:umap3d}
\end{figure}

The projection of Figures~\ref{fig:umap} and~\ref{fig:umap3d} is built as follows.
An autoencoder with layers $256 \to 256 \to 128 \to k$ and $k \to 128 \to 256 \to 256$ maps the unit directions $\uu_f$ to codes $\mathbf{c}_f$ normalised to the unit sphere $S^{k-1}$, and is trained for 60 epochs on the 120k training directions with the loss $1 - \cos(\uu_f, \hat{\uu}_f)$.
The scale is then reintroduced as the radius, $\mathbf{c}_f\,r(n(f))$, and the display was chosen on the validation split by the 10-nearest-neighbour accuracy of the function family in the displayed coordinates, among the radii $r = \log(1 + n)$ and $r = 0.25 + 0.75\,\mathrm{rank}(\log n)$ and an additional loss term $\big(\mathbf{c}_i^\top\mathbf{c}_j - \uu_i^\top\uu_j\big)^2$ that preserves the $\Hone$ angles.
The angle term raises the correlation between code and latent cosines from 0.50 to 0.62 for $k = 2$ and from 0.61 to 0.73 for $k = 3$, but it does not improve the family accuracy, so it is not used.
On the test split, the mean of $1 - \cos(\uu_f, \hat{\uu}_f)$ is 0.283 for $k = 2$ and 0.115 for $k = 3$ (median angular errors of 33$^\circ$ and 16$^\circ$), and the validation family accuracy is 0.431 and 0.518, against 0.630 for the UMAP projection of the same functions, which preserves local neighbourhoods better but discards the norms and has no explicit map.

\section{Time and memory consumption}
\label{app:time}

Table~\ref{tab:time} reports the training cost of each component of \fjepa{}; peak GPU memory during encoder training was 2.7~GB with gradient checkpointing.
All components together have 120M parameters (encoder and $\Dop$ 35.1M, decoder 17.7M, composition bank 52.9M, integrator 7.6M and value encoder 7.4M), and every one of them is trained on a single 8~GB laptop GPU.

\begin{table}[H]
\caption{Training cost of the components of \fjepa{} on a single RTX 4070 Laptop GPU (8~GB)}
\label{tab:time}
\centering
\small
\begin{tabular}{lccc}
\toprule
Component & Parameters & Steps & Time \\
\midrule
Encoder and $\Dop$ & 35.1M & 40k & 6.6 h (0.6 s/step) \\
Decoder & 17.7M & 105k & 3.5 h (0.11 s/step) \\
Composition bank (7 operators) & 52.9M & 40k & 1 h \\
Integrator & 7.6M & 60k & 13 min (0.013 s/step) \\
Value encoder (fine-tuning) & 7.4M & 20k & 39 min \\
\bottomrule
\end{tabular}
\end{table}

Table~\ref{tab:inference} compares the inference cost and the size of \fjepa{} with those of the other methods, measured in our own runs on the same machine (16 CPU cores and the same GPU). For \fjepa{}, the parameters are those of the modules that the task uses: the value encoder and the decoder for symbolic regression and for the tasks that start from samples (ODE solving and discovery), and the encoder with $\Dop$, the integrator and the decoder for integration.
On every task, \fjepa{} has fewer parameters than every pretrained neural network it is compared with: 25.1M against 26.4M (NeSymReS) to 166.7M (SNIP) for symbolic regression, 60.4M against 185.0M for integration and for ODE solving with the models of Lample and Charton, and 25.1M against 60.6M for ODE discovery with ODEFormer. Even all the modules of \fjepa{} together (120M) are smaller than a single integrator of Lample and Charton.
For symbolic regression, \fjepa{} is also the fastest method: a forward pass of the value encoder, a beam search of the decoder and a few least-squares refits take 0.44 seconds per function, whereas the search-based methods use their whole budget (PySR and Operon, 20 seconds), SNIP refines its latent by an evolutionary search (23 seconds) and MMSR needs 93 seconds; the pretrained transformers E2E and NeSymReS take 2.0 and 1.6 seconds.
For integration and ODE solving, in contrast, \fjepa{} is slower than the sequence-to-sequence models, since it explores several routes and verifies every candidate with SymPy: 3.4 seconds per integrand against 0.15 to 0.91 seconds for Lample and Charton, and 1.4 seconds per IVP against 0.29 seconds. In terms of memory, the integration solver peaks at 4.1~GB of GPU memory, most of which holds the latents of the 2M-function pool used for the shortlists.

\begin{table}[H]
\caption{Inference time per item (median wall-clock seconds on the same machine) and number of parameters of the networks used by each method. For \fjepa{} on symbolic regression and integration, the time is the mean over the 300 test items}
\label{tab:inference}
\centering
\small
\begin{tabular}{llcc}
\toprule
Task & Method & Seconds per item & Parameters \\
\midrule
\multirow{7}{*}{Symbolic regression} & \fjepa{} (value encoder and decoder) & \textbf{0.44} & \textbf{25.1M} \\
 & NeSymReS \cite{biggio2021nesymres} & 1.6 & 26.4M \\
 & E2E transformer \cite{kamienny2022e2e} & 2.0 & 93.5M \\
 & Operon \cite{burlacu2020operon} (20 s budget) & 20.0 & -- \\
 & PySR \cite{cranmer2023pysr} (20 s budget) & 20.2 & -- \\
 & SNIP \cite{meidani2024snip} & 23.1 & 166.7M \\
 & MMSR \cite{li2025mmsr} & 92.7 & 90.0M \\
\midrule
\multirow{3}{*}{Integration} & \fjepa{} (encoder, $\Dop$, integrator, decoder) & 3.4 & \textbf{60.4M} \\
 & Lample and Charton \cite{lample2020deep} & 0.15--0.91 & 185.0M \\
 & SymPy \texttt{integrate} & \textbf{0.03--0.43} & -- \\
\midrule
\multirow{3}{*}{ODE solving (IVPs)} & \fjepa{} (value encoder and decoder) & 1.43 & \textbf{25.1M} \\
 & Lample and Charton \cite{lample2020deep} & \textbf{0.29} & 185.0M \\
 & SymPy \texttt{dsolve} & 4.34 & -- \\
\midrule
\multirow{5}{*}{ODE discovery} & \fjepa{} (value encoder and decoder) & 3.21 & \textbf{25.1M} \\
 & ODEFormer \cite{dascoli2024odeformer} & 0.54 & 60.6M \\
 & SINDy (PySINDy) & \textbf{0.03} & -- \\
 & Weak SINDy (PySINDy) & 0.35 & -- \\
 & Estimated derivatives + PySR & 20.3 & -- \\
\bottomrule
\end{tabular}
\end{table}

\section{Galleries of predictions}
\label{app:gallery}
\begin{figure}[p]
\centering
\includegraphics[width=\linewidth,height=0.82\textheight,keepaspectratio]{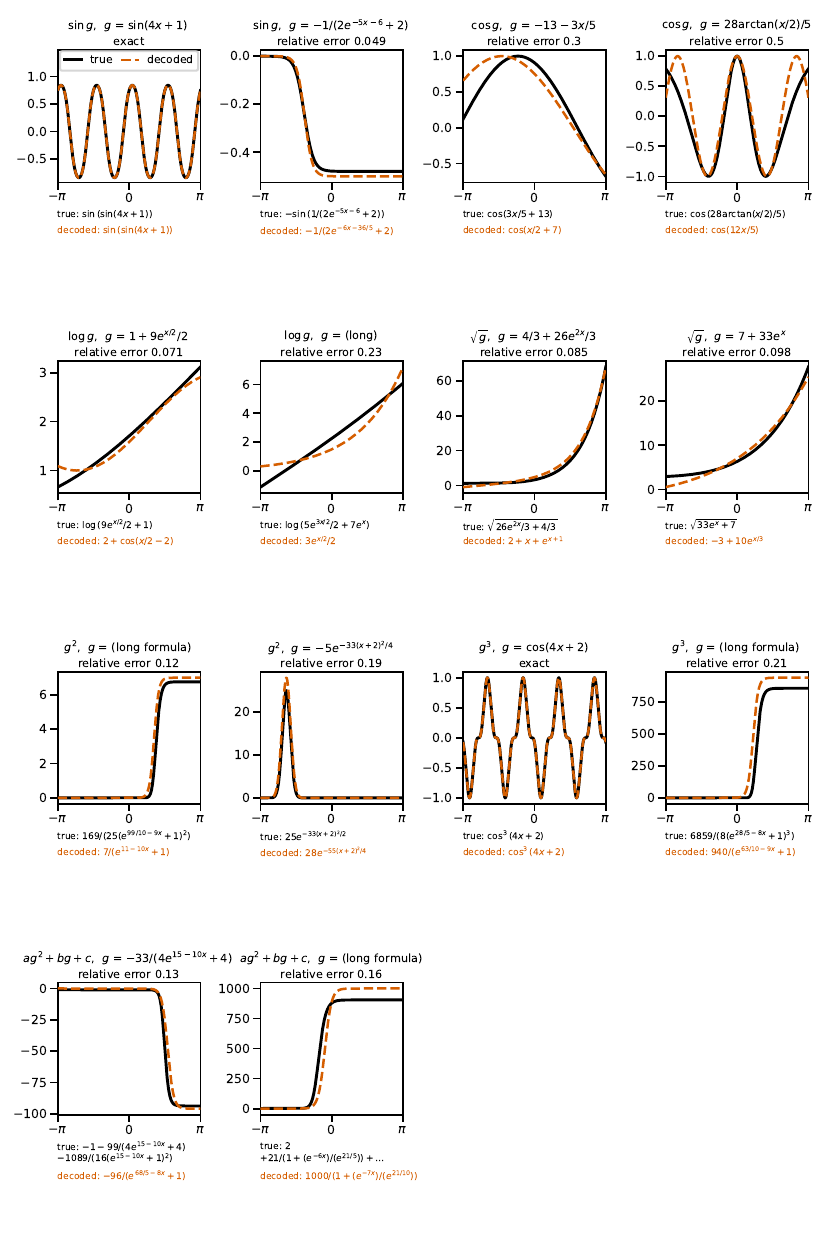}
\caption{Composition outputs $\Mop_h(\z_g)$ decoded to formulas, two per operator in the order $\sin$, $\cos$, $\log$, $\sqrt{\cdot}$, square, cube, and quadratic. These are the decodings with the smallest error among the first 150 random test pairs, and are not representative of the average decoding}
\label{fig:galcomp}
\end{figure}

The following figures show unselected examples (the first test functions in a fixed random order, or every problem of a benchmark), except Figure~\ref{fig:galcomp}, whose examples are selected as described in its caption; the examples already shown in Figure~\ref{fig:predops} and Figure~\ref{fig:predapp} are not repeated. In Figure~\ref{fig:predops}, the panels show the first exact (a) and the first inexact (b) reconstruction of $f$, the first exact (c) and an inexact (d) decoding of $\Dop(\z_f)$ with the correct shape, the first decoding of $\Dop(\Dop(\z_f))$ with the correct shape (e), and the first inexact decoding of $\mathcal{I}(\z_f)$ (f), with antiderivatives shown with zero mean; in Figure~\ref{fig:predapp}, panels (a) and (b) are the first exact and the first inexact recovery.
The decoded formula is the top candidate of beam search without re-ranking or constant refitting, so these examples include the failure modes of the latent operators and of the decoder. Below each panel, the true formula is written in black and the produced formula in orange (floating-point coefficients rounded to three significant digits; very long sums are truncated).

\begin{figure}[p]
\centering
\includegraphics[width=\linewidth,height=0.82\textheight,keepaspectratio]{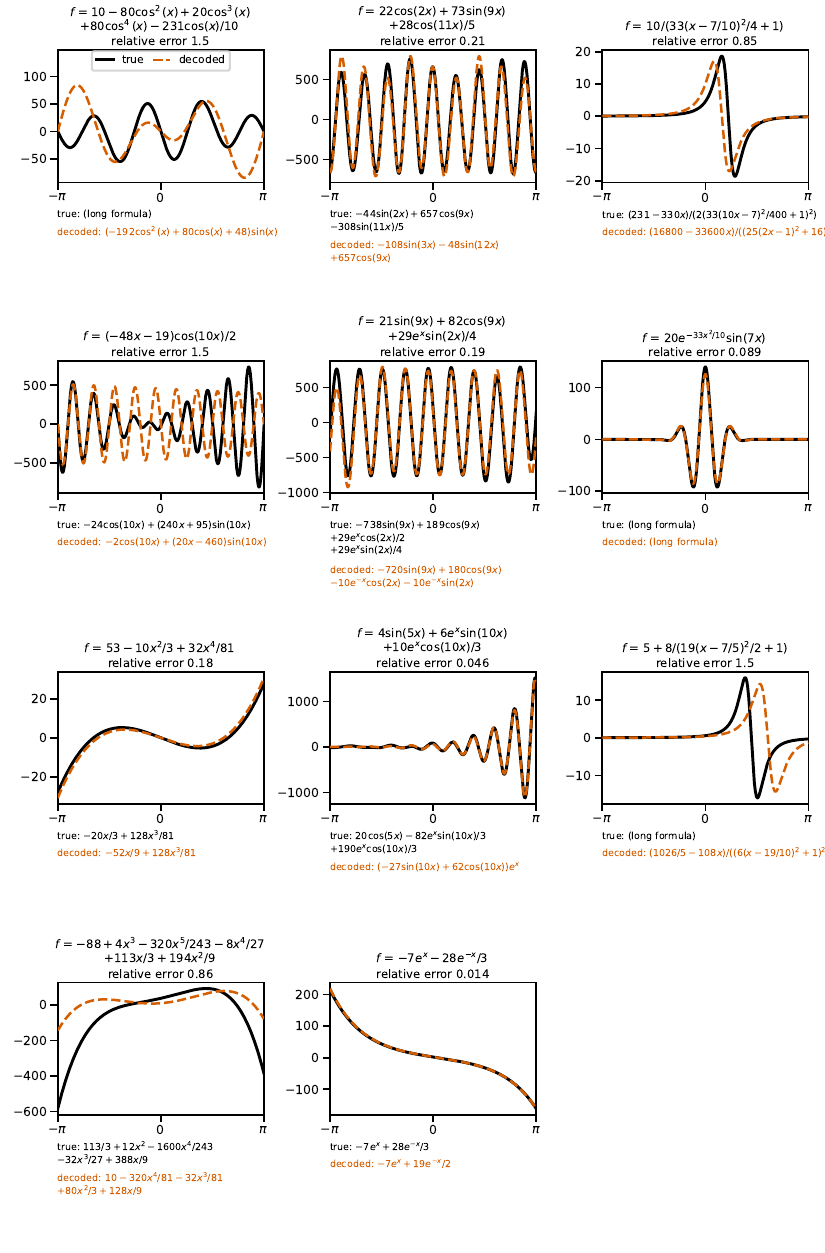}
\caption{Derivatives predicted in latent space and decoded, $f'$ from $\Dop(\z_f)$, for the first test functions in a fixed random order, without the one shown in Figure~\ref{fig:predops} (titles give $f$)}
\label{fig:galderiv}
\end{figure}

In this unselected sample (Figure~\ref{fig:galderiv}) no derivative is decoded exactly, in agreement with the 6.8\% exact rate of Section~\ref{sec:exp_decoding}: 6 of the 11 decoded derivatives have the correct family and frequencies with relative errors below 0.25, whereas the other 5 miss a phase, a term or a scale, mostly for oscillatory functions with large coefficients.

Decoding the outputs of the composition operators is much harder than retrieving them, since the decoder is not trained on composed latents and the composed expressions are long; nevertheless, Figure~\ref{fig:galcomp} shows that for every operator some outputs are decoded into formulas that follow the true function closely, two of them exactly ($\sin(\sin(4x+1))$ and $\cos^3(4x+2)$), and that the typical residual error is a constant or a frequency slightly off, whereas the same latents retrieve the correct function for 57\% to 92\% of the pairs (Table~\ref{tab:composition}).

\begin{figure}[p]
\centering
\includegraphics[width=\linewidth,height=0.82\textheight,keepaspectratio]{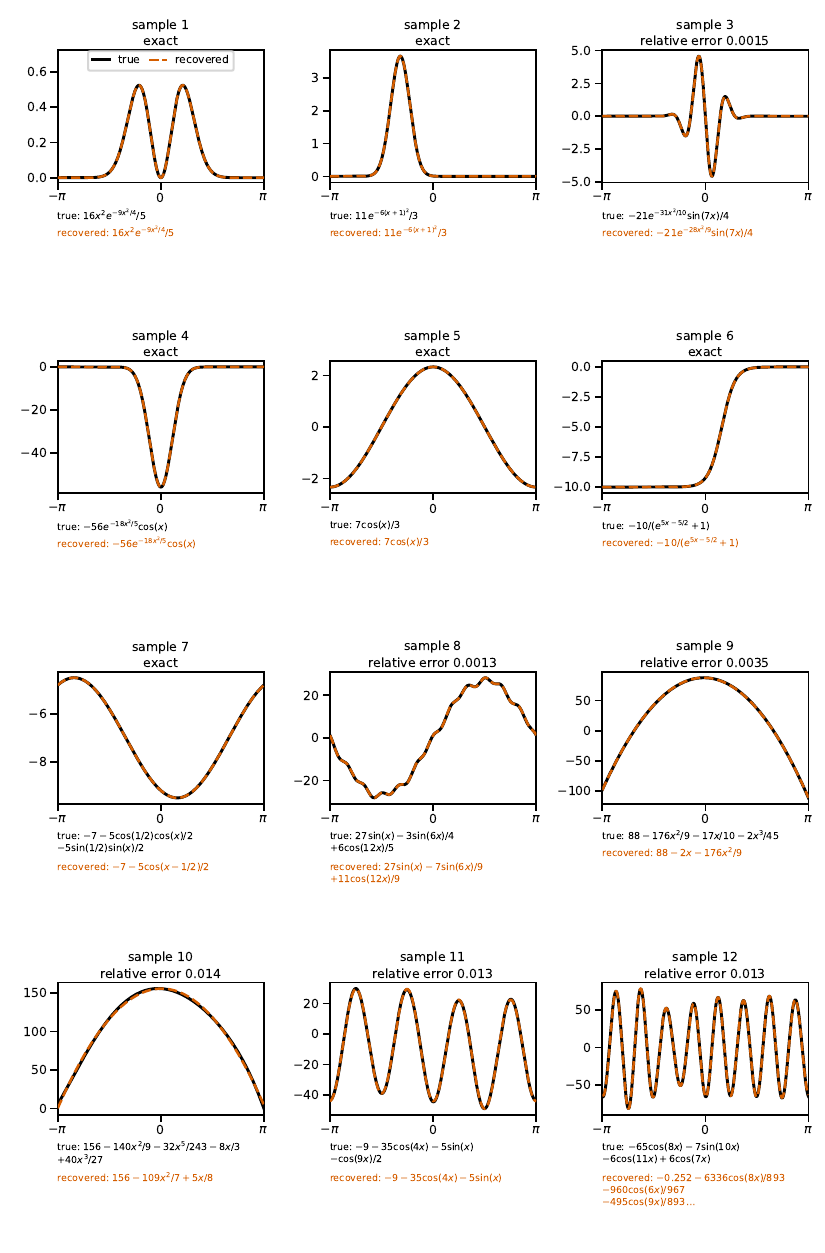}
\caption{Symbolic regression from 64 samples at $\sigma = 10^{-2}$ for the first 12 test functions in a fixed random order that are not shown in Figure~\ref{fig:predapp} (titles give the true function)}
\label{fig:galsr}
\end{figure}

\begin{figure}[p]
\centering
\includegraphics[width=\linewidth,height=0.82\textheight,keepaspectratio]{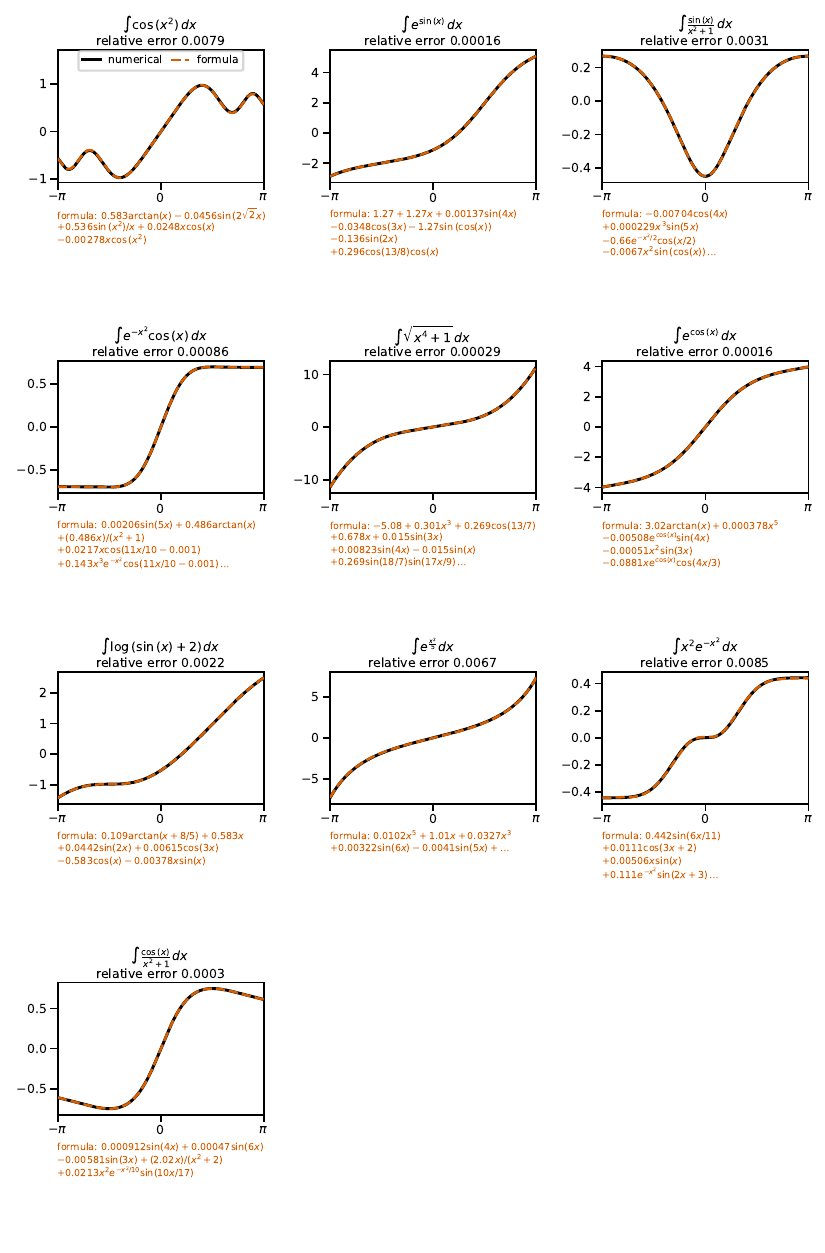}
\caption{The integrands without elementary antiderivative of Table~\ref{tab:nonelem} that are not shown in Figure~\ref{fig:predapp}: five-term symbolic approximation against cumulative quadrature (relative $L^2$ error of the antiderivative, both curves with zero mean)}
\label{fig:galnonelem}
\end{figure}

\begin{figure}[p]
\centering
\includegraphics[width=\linewidth,height=0.82\textheight,keepaspectratio]{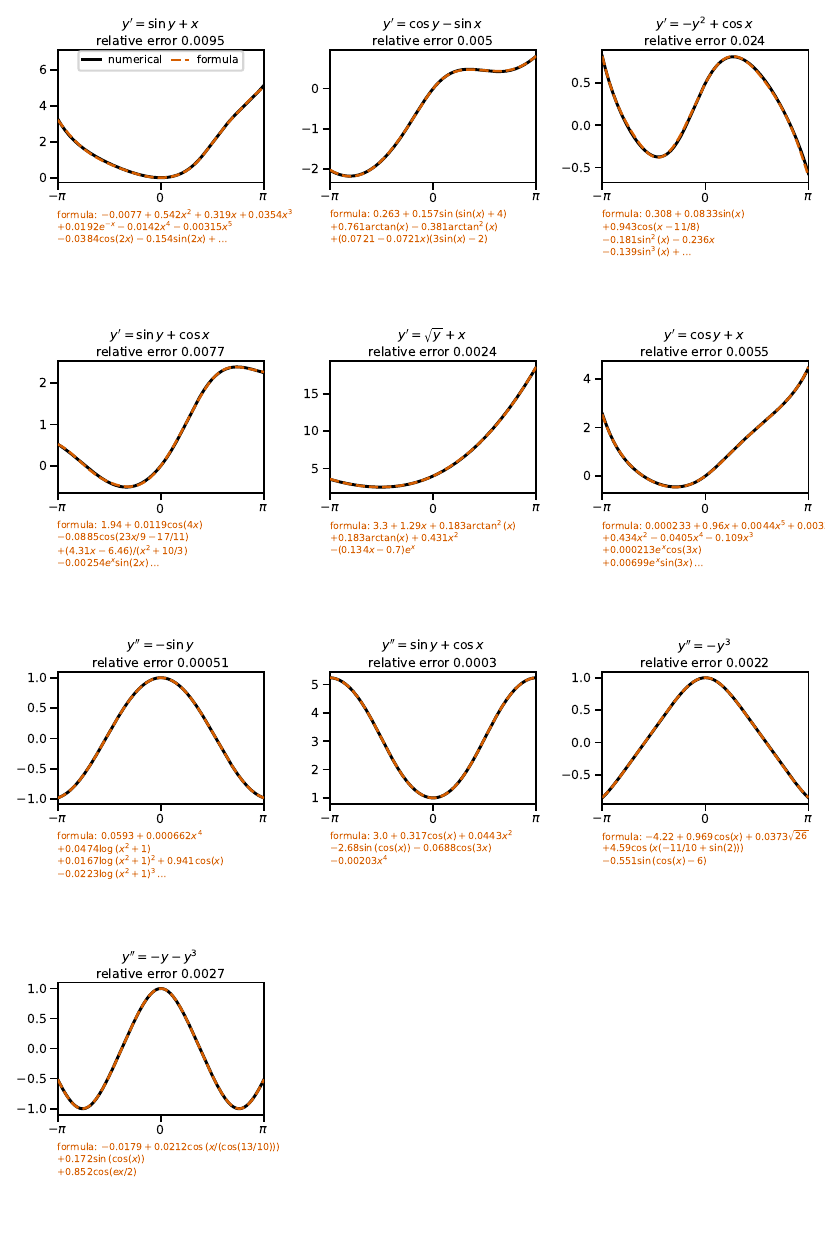}
\caption{The ten initial-value problems without closed-form solution that are not shown in Figure~\ref{fig:predapp}: four-atom formula of the best learned method against the numerical solution}
\label{fig:galivp}
\end{figure}

Figure~\ref{fig:galsr} to Figure~\ref{fig:galivp} complete the picture for the applications: at 1\% noise, symbolic regression errors are small constant errors, the five-term approximations of non-elementary antiderivatives deviate by at most 3.1\% from the antiderivative obtained by quadrature (Table~\ref{tab:nonelem} reports the larger error of their derivative with respect to the integrand), and four-atom formulas track the numerical solutions of all twelve initial-value problems, with the Painlev\'e-I type equation of Figure~\ref{fig:predapp} as the only case with a visible deviation.

\end{document}